\documentclass[onefignum,onetabnum]{siamonline220329}

\makeatletter
\newcommand*{\addFileDependency}[1]{
  \typeout{(#1)}
  \@addtofilelist{#1}
  \IfFileExists{#1}{}{\typeout{No file #1.}}
}
\makeatother

\usepackage{lipsum}
\usepackage{amsfonts}
\usepackage{graphicx}
\usepackage{epstopdf}
\usepackage{algorithmic}
\ifpdf
  \DeclareGraphicsExtensions{.eps,.pdf,.png,.jpg}
\else
  \DeclareGraphicsExtensions{.eps}
\fi

\usepackage{enumitem}
\setlist[enumerate]{leftmargin=.5in}
\setlist[itemize]{leftmargin=.5in}

\newsiamremark{remark}{Remark}
\newsiamremark{hypothesis}{Hypothesis}
\crefname{hypothesis}{Hypothesis}{Hypotheses}
\newsiamthm{claim}{Claim}

\usepackage{amssymb}
\usepackage{amsmath}
\usepackage{mathtools}
\usepackage{subfig}

\renewcommand{\vec}[1]{\mathbf{#1}}
\newcommand{\vecgreek}[1]{\boldsymbol{#1}}
\newcommand{\mtx}[1]{\mathbf{#1}}

\DeclareMathOperator*{\argmin}{arg\,min}

\DeclareMathOperator{\Tr}{Tr}
\newcommand{\mtxtrace}[1]{\Tr\left\lbrace{#1}\right\rbrace}
\newcommand{\expectation}[1]{\mathbb{E}\left[{{#1}}\right]} 
\newcommand{\expectationwrt}[2]{\mathbb{E}_{#2}\left[{{#1}}\right]}

\newcommand{\Ltwonorm}[1]{\left\Vert{{#1}}\right\Vert_2^2} 
\newcommand{\Frobnorm}[1]{\left\Vert{{#1}}\right\Vert_F^2} 
\newcommand{\Frobnormnonsquared}[1]{\left\Vert{{#1}}\right\Vert_F} 

\newcommand{\eigenvalue}[2]{\lambda_{#2}\Big\lbrace{{#1}}\Big\rbrace}

\newtheorem{assumption}{Assumption}

\usepackage{color}
\newcommand {\richb}[1]{}
\newcommand {\ydar}[1]{}
\newcommand {\daniel}[1]{}
\newcommand {\danielR}[1]{}

\newcommand {\ydarnew}[1]{}

\newcommand {\ydarR}[1]{}

\newcommand {\ydarRminor}[1]{}

\headers{The Common Intuition to Transfer Learning Can Win or Lose}{Yehuda Dar, Daniel LeJeune, Richard G.~Baraniuk}

\title{The Common Intuition to Transfer Learning Can Win or Lose:\\Case Studies for Linear Regression\thanks{Submitted to the editors DATE.
\funding{This work was supported by NSF grants CCF-1911094, IIS-1838177, and IIS-1730574; ONR grants N00014-18-12571, N00014-20-1-2534, N00014-23-1-2714, and MURI N00014-20-1-2787; AFOSR grant FA9550-22-1-0060; and a Vannevar Bush Faculty Fellowship, ONR grant N00014-18-1-2047.}}}

\author{Yehuda Dar\thanks{Department of Computer Science, Ben-Gurion University
  (\email{ydar@bgu.ac.il}).}
  \and Daniel LeJeune\thanks{Department of Statistics, Stanford University
  (\email{daniel@dlej.net}).}
\and Richard G.~Baraniuk\thanks{Department of Electrical and Computer Engineering, Rice University
  (\email{richb@rice.edu}).}}

\usepackage{amsopn}

\definecolor{editone}{HTML}{0000FF}
\definecolor{edittwo}{HTML}{FF0099}

\ifpdf
\hypersetup{
  pdftitle={The Common Intuition to Transfer Learning Can Win or Lose: Case Studies for Linear Regression},
  pdfauthor={Y. Dar, D. LeJeune, and R. G. Baraniuk}
}
\fi

\begin{document}

\maketitle

\begin{abstract}%
 We study a fundamental transfer learning process from source to target linear regression tasks, including overparameterized settings where there are more learned parameters than data samples. 
The target task learning is addressed by using its training data together with the parameters previously computed for the source task. 
We define a transfer learning approach to the target task as a linear regression optimization with a regularization on the distance between the to-be-learned target parameters and the already-learned source parameters. 
We analytically characterize the generalization performance of our transfer learning approach and demonstrate its ability to resolve the peak in generalization errors in double descent phenomena of the minimum $\ell_2$-norm solution to linear regression. 
Moreover, we show that for sufficiently related tasks, the optimally tuned transfer learning approach can outperform the optimally tuned ridge regression method, even when the true parameter vector conforms to an isotropic Gaussian prior distribution.  
Namely, we demonstrate that transfer learning can beat the minimum mean square error (MMSE) solution of the independent target task. 
Our results emphasize the ability of transfer learning to extend the solution space to the target task and, by that, to have an improved MMSE solution. We formulate the linear MMSE solution to our transfer learning setting and point out its key differences from the common design philosophy to transfer learning.
\end{abstract}



\section{Introduction}
Contemporary machine learning models are often {\em overparameterized}, meaning that they are more complex (e.g., have more parameters to be learned) than the amount of data available for their training. 
Deep neural networks are a very successful example for highly overparameterized models that are often trained without explicit regularization.
The challenge in such overparameterized learning is to be able to generalize well beyond the given dataset, despite the tendency of overparameterized models to perfectly fit their (possibly noisy) training data \cite{zhang2017understanding}. 


The empirical studies by  \cite{belkin2019reconciling,geiger2019scaling,spigler2018jamming} show that generalization errors follow a \textit{double descent} shape when examined with respect to the complexity of the learned model. In the double descent shape, the generalization error peaks when the learned model becomes sufficiently complex and begins to perfectly fit (i.e., interpolate) the training data. This peak in generalization error reflects poor generalization performance, but when the learned model complexity increases further then the generalization error starts to decrease again and eventually may achieve excellent generalization ability for highly overparameterized models---even despite perfectly fitting noisy training data! 
The prevalence of double descent phenomena in deep learning motivated a corresponding field of theoretical research where learning of overparameterized models is analytically studied mainly for linear regression problems \cite{hastie2019surprises,belkin2020two,bartlett2020benign,xu2019number,mei2019generalization,muthukumar2020harmless,nakkiran2020optimal,dascoli2020double}, as well as for other statistical learning problems such as classification (e.g., \cite{gerace2020generalisation,kini2020analytic,mignacco2020role,muthukumar2021classification,wang2021benign,deng2021model}) and linear subspace learning \cite{dar2020subspace}.
The existing literature show that double descent phenomena occur in minimum-norm solutions to overparameterized least squares regression; i.e., when there is no explicit regularization in the learning process. Moreover, \cite{hastie2019surprises,nakkiran2020optimal} show that the explicit regularization in ridge regression is able to resolve the generalization error peak of the double descent behavior in the minimum $\ell_2$-norm solution to linear regression (similar findings were provided in \cite{gerace2020generalisation,mei2019generalization} for models other than ridge regression).

{\em Transfer learning} \cite{pan2009survey} is a key approach in the practical training of deep neural networks (DNNs) where learning is conducted not only using a dataset that is relatively small compared to the complexity of the DNN, but also using layers of parameters taken from a ready-to-use DNN that was properly trained for a related task \cite{bengio2012deep,shin2016deep,long2017deep}. Transfer learning between DNNs can be done by transferring network layers from the source to target models and setting them fixed (while other layers are learned), fine tuning them (i.e., moderately adjusting to the target task data), or using them as initialization for a comprehensive learning process. 
Clearly, the source task should be sufficiently related to the target task in order to have a useful transfer learning  \cite{rosenstein2005transfer,zamir2018taskonomy,kornblith2019better}. Nevertheless, finding successful transfer learning settings is still a fragile task \cite{raghu2019transfusion} that requires a further understanding---also from theoretical perspectives. 

Surprisingly, there are only a few \textit{analytical} theories for transfer learning, with limited insight regarding double descent. Lampinen and Ganguli \cite{lampinen2018analytic} analyze the optimization dynamics of transfer learning for multi-layer linear networks. 
Dhifallah and Lu \cite{dhifallah2021phase} analyze transfer learning of perceptron models for classification and regression where the target model is trained with a fixed subset of source parameters or trained with regularization on a weighted Euclidean distance from the source model; in \cite{dhifallah2021phase}, the training includes explicit $\ell_2$-norm regularization on the learned parameters such that training data interpolation and the double descent phenomenon do not seem to appear in both the source model and transfer learning of the target model.
Gerace et al. \cite{gerace2021probing} examine transfer learning of two-layer nonlinear models for binary classification where the source and target data generating models are related via a correlated hidden manifold model. In \cite{gerace2021probing}, the first layer of the target model is set fixed as the first source model layer and only the second layer of the target model is learned (this approach has an interesting interpretation as the transfer learning analog to the random feature model); they also numerically examine fine tuning of the entire model. The target model training in \cite{gerace2021probing} includes explicit $\ell_2$-norm regularization that prevents training data interpolation although attenuated double descent phenomena are still observed. 
Obst et al. \cite{obst2021transfer} analyze fine tuning of underparameterized linear regression via gradient descent. 
Clearly, there are still more aspects and settings of transfer learning that should be analytically understood, even for linear architectures.

In this paper, we study transfer learning between two linear regression tasks where the transferred parameters from the source task are utilized for the learning of the target task's parameters. Specifically, we formulate the target task as a linear regression problem that includes regularization on the distance between the transferred source parameters (that were already learned)  and the to-be-learned parameters of the target task. 
One can also perceive the transfer learning approach in this paper as transferring parameters from the source task and adjusting them to the training data of the target task; for relatively similar source and target tasks the adjustment of parameters is modest and can be interpreted as a \textit{fine tuning} mechanism. 
The settings and analytical techniques in this paper significantly differ from those in previous studies on transfer learning.  \daniel{perhaps we should contrast with specific works?}\ydarnew{can't the contrast be implied from the text of the previous paragraph and this one? if you suggest to simply add the references \cite{lampinen2018analytic,dar2020double,dhifallah2021phase,gerace2021probing,obst2021transfer} then it's possible, although it can be also implied from the text}

We examine target and source tasks that have a noisy linear relation between their true parameter vectors. Namely, the true parameter vector of the source task is a linear transformation of the true parameter vector of the target task plus additive white Gaussian noise. 
We study our transfer learning approach under two different assumptions: 
\begin{itemize}
    \item For a \textit{partial} knowledge of the statistical relation between the tasks: (i) we consider the utilization of the linear transformation in a well-specified form, but also in a misspecified form where only part of the linear transformation is known; (ii) the second-order statistics of the task-relation noise is known but not the specific realization of the noise vector.
    \item For an \textit{unknown} relation between the tasks. Specifically, we consider a setting where one does not know the linear transformation in the task relation, and therefore assumes it is the identity operator. 
\end{itemize}

Our transfer learning approach includes a regularization coefficient that determines the importance of the source parameters in the learning of the target parameters. We consider the optimally tuned version of our transfer learning approach and study its generalization performance from analytical and empirical perspectives. 

In various cases in the overparameterized regime, our transfer learning approach can be improved by ignoring the true task relation operator and assuming it is the identity matrix. Namely, our approach is highly suitable for various cases of unknown task relation. We explain this by noting that the true task relation can induce small eigenvalues in a matrix inversion (that also involves the rank deficient feature matrix) in the overparameterized learned model and, thus, can increase the test error; on the other hand, using an identity matrix instead of the true task relation operator regularizes the required matrix inversion and can achieve lower test error despite ignoring the true task relation. 

We show that our optimally tuned transfer learning can outperform the optimally tuned ridge regression solution of the independent target task. Remarkably, we prove this result also for the case where optimally tuned ridge regression provides the minimum mean square error (MMSE) estimate for the parameters of the individual target task (namely, the case where the true parameters of the target task follow an isotropic Gaussian distribution and the source task solution is not utilized). 
We show that transfer learning outperforms ridge regression if the target and source tasks are sufficiently related and the source task solution generalizes sufficiently well at the source task itself (that is, the source task solution is sufficiently accurate). 
%

The main contribution of optimally tuned ridge regression is to resolve the generalization error peak of the minimum $\ell_2$-norm (ML2N) solution of the individual target task (see, e.g., red curves in Fig.~\ref{fig:error_curves_diagrams_isotropic_H_is_orthonormal} where their peaks are located at the point where the \textit{target} task model shifts from being underparameterized to being overparameterized). 
Optimally tuned transfer learning from a sufficiently related and accurate source task has the ability to not only resolve the peak of the ML2N solution, but also to achieve significantly lower generalization errors than the ridge solution over a wide range of parameterization levels (see, e.g., blue curves in Fig.~\ref{fig:error_curves_diagrams_isotropic_H_is_orthonormal}). 
Importantly, the usefulness of transfer learning based on the ML2N solution of the source task may have a by-product in the form of another generalization error peak, which is located at the point where the \textit{source} task model shifts from being underparameterized to being overparameterized (see, e.g., blue curves in Fig.~\ref{fig:error_curves_diagrams_isotropic_H_is_orthonormal}). 

While our transfer learning approach relies on the common intuition for how to utilize the source task solution for the learning of the target model, we demonstrate that it is not always beneficial compared to ridge regression. This motivates us to examine the linear MMSE (LMMSE) solution to the transfer learning problem. By this, we exemplify that the common design philosophy to transfer learning can sometimes be far from the best utilization of the pre-trained source model. 

\subsection{Summary of the Principal Concepts of the Proposed Theory}
The following further emphasizes the main concepts which our work contributes to the understanding of. 
\begin{enumerate}
    \item \textbf{Negative transfer.} Our theory enables us to analyze negative transfer cases where our transfer learning makes the target model to generalize worse than the optimally tuned ridge regression solution (and the minimum $\ell_2$-norm solution to least squares) for the target model (without any transfer from the source model). Negative transfer is an important aspect which is reflected in various transfer learning theories (e.g., \cite{dar2020double,gerace2021probing}), and indeed each of these previous works analyzes the negative transfer topic from a different perspective that stems from the particular transfer learning method under study. 
    For example, negative transfer in \cite{dar2020double} is defined as cases where transferring more parameters degrades the generalization performance of the target model, and the learning from scratch benchmark is the minimum $\ell_2$-norm solution to least squares. In \cite{gerace2021probing}, the two-layer model and transfer of the first layer of feature maps leads to the definition of negative transfer as when such transfer learning generalizes worse than learning the target model from scratch via the random feature model. 
    In our work, negative transfer is reflected by Corollary \ref{corollary:well specified - transfer learning is better than ridge - isotropic beta assumption} that formulates (for a known task relation and isotropic input feature) when the optimally tuned version of the proposed transfer learning generalizes worse than ridge regression. Moreover, our results in Figures \ref{fig:error_curves_diagrams_isotropic_H_is_orthonormal}-\ref{fig:error_curves_diagrams_isotropic_general_H__wellspec_dct_domain} demonstrate cases of negative transfer as parameterization levels at which the ridge regression (and sometimes also the minimum $\ell_2$-norm solution to least squares) has a lower test error than the examined transfer learning models. 
    
    \item \textbf{Improved generalization by ignoring the true task relation.} We demonstrate foundational study cases where not using the true task relation is beneficial. While most of the transfer learning theories \textit{implicitly} ignore some or all of the task relation, in our case we \textit{explicitly} study the implications of the task relation on our transfer learning method -- including a direct comparison of the generalization performance with and without utilizing the task relation (see Section \ref{subsec:improvements due to Htilde=I}). As task relations are usually unknown (at least not sufficiently accurate) in practice, our results suggest that this does not necessarily reduce the generalization performance.
    
    \item \textbf{Transfer learning from an interpolating source model.} To the best of our knowledge, only \cite{dar2020double} previously considered transfer from interpolating source models but with the transferred parameters being fixed in the target model. In this paper, our theory considers transfer that regularizes the distance of the target model from a source model that interpolates its own dataset (if the number of parameters is larger than the number of source train samples). Hence, our theory further adds to the understanding of transfer learning from interpolating source models. 
    
    Specifically, a \textbf{double descent} phenomenon in the source model can induce a corresponding test error peak in transfer learning of the target model (this error peak is around a target parameterization level that corresponds to the interpolation threshold of the source model); this behavior was first analyzed in \cite{dar2020double} and we further demonstrate it here for our new transfer learning approach (for examples, see the blue error curves in Figures \ref{fig:error_curves_diagrams_isotropic_H_is_orthonormal}, \ref{fig:error_curves_diagrams_isotropic_general_H__misspecified}, \ref{fig:error_curves_diagrams_isotropic_general_H__wellspec_dct_domain}).

    \item \textbf{The optimal transfer learning for linear models.} The main transfer learning approach in this work has settings at which it generalizes worse than ridge regression (i.e., these are negative transfer cases). This leads us to define and examine the linear minimum MSE (LMMSE) solution for our transfer learning setting (Section \ref{subsec:The Linear MMSE Solution to Transfer Learning}). Among the insights that this LMMSE solution provides, our results in Figures \ref{fig:error_curves_diagrams_isotropic_general_H__with_LMMSE_TL}, \ref{fig:error_curves_diagrams_isotropic_general_H__with_LMMSE_TL__HtildeI} clearly demonstrate that the LMMSE transfer learning approach does not have negative transfer cases with respect to other linear models. \ydarRminor{does this correspond to something from previous transfer learning theories?}
\end{enumerate}
It should be noted that our theory considers a linear transfer learning model and therefore it cannot shed lights on the effects of model depth (as in \cite{lampinen2018analytic}), transfer of feature maps from the source model (as in \cite{gerace2021probing}), or nonlinear activation functions (e.g., as in \cite{dhifallah2021phase,gerace2021probing}).

\subsection{Paper Organization}
This paper is organized as follows. In Section~\ref{sec:Problem Settings: Transfer Learning between Linear Regression Tasks}, we outline the settings of the source and target tasks, as well as the model for their relation. In Section~\ref{sec:Well-specified Feature Selection}, we present an intuitive design to transfer learning and study its generalization performance in Sections~\ref{sec:Analysis for H of an Orthonormal Matrix Form}--\ref{sec:Analysis for H of a General Form}. 
Specifically, in Section \ref{sec:Analysis for H of an Orthonormal Matrix Form} we focus on a noisy task relation model where the linear transformation is based on a \textbf{known} orthonormal matrix and the target features are isotropic; this yields formulations that clearly show the effect of the proximity between the two tasks (Section \ref{subsec:Analysis for H of an Orthonormal Matrix Form}), explain the target test error peak around the source interpolation threshold (Section \ref{sec:The Test Error Peak in the Intuitive Transfer Learning}), and provide a clear comparison to ridge regression (Section~\ref{sec:Transfer Learning versus Ridge Regression}). In Section~\ref{sec:misspecification} we analyze the effect of misspecification, which in our transfer learning case also implies a partial knowledge of the task relation. 
In Section \ref{sec:Analysis for H of a General Form} we extend our analysis by considering an \textbf{unknown} task relation with any linear transformation and anisotropic target features; for this we formulate the generalization error of the intuitive transfer learning method (Section \ref{subsec:Analysis for H of a General Form - Analysis of the Intuitive Transfer Learning Approach}), explore why the intuitive transfer learning can be improved by ignoring the true task relation (Section \ref{subsec:improvements due to Htilde=I}), and also examine the linear MMSE solution to transfer learning (Section~\ref{subsec:The Linear MMSE Solution to Transfer Learning}).
Section \ref{sec:Conclusions} concludes this paper. Additional details and mathematical proofs are provided in Appendices \ref{appendix:sec:Additional Details for Section of The Test Error of the Source Task}-\ref{appendix:sec:Details and Proofs for Theorem 5}

\section{Problem Settings: Two Related Linear Regression Tasks}
\label{sec:Problem Settings: Transfer Learning between Linear Regression Tasks}

\subsection{Source Task: Data Model and Solution Form}
The \textit{source task} is a linear regression problem with a $d$-dimensional Gaussian input ${\vec{z}\sim \mathcal{N}\left(\vec{0},\mtx{I}_d\right)}$ and a response value $v\in\mathbb{R}$ that is induced by ${v = \vec{z}^T \vecgreek{\theta} + \xi}$, 
where $\xi\sim\mathcal{N}\left(0,\sigma_{\xi}^2\right)$ is a noise variable independent of $\vec{z}$, and $\vecgreek{\theta}\in\mathbb{R}^d$ is an unknown parameter vector. Motivation for linear models with random features can be found, e.g., in \cite{hastie2019surprises,chizat2018note}.
While not knowing the true distribution of $\left(\vec{z},v\right)$, the source learning task is carried out using a dataset ${\widetilde{\mathcal{D}}\triangleq\Big\{ { \left({\vec{z}^{(i)},v^{(i)}}\right) }\Big\}_{i=1}^{\widetilde{n}}}$ that includes  $\widetilde{n}$ independent and identically distributed (i.i.d.)~samples of $\left(\vec{z},v\right)$.
We also denote the $\widetilde{n}$ data samples in ${\widetilde{\mathcal{D}}}$ using an $\widetilde{n}\times d$ input matrix ${\mtx{Z}\triangleq \lbrack {\vec{z}^{(1)}, \dots, \vec{z}^{(\widetilde{n})}} \rbrack^{T}}$ and a $\widetilde{n}\times 1$ response vector ${\vec{v}\triangleq \lbrack{ v^{(1)}, \dots, v^{(\widetilde{n})} } \rbrack^{T}}$. Therefore, ${\vec{v} = \mtx{Z} \vecgreek{\theta} + \vecgreek{\xi}}$ where ${\vecgreek{\xi}\triangleq \lbrack{ {\xi}^{(1)}, \dots, {\xi}^{(\widetilde{n})} } \rbrack^{T}}$ is an unknown noise vector whose $i^{\rm th}$ component ${\xi}^{(i)}$ originates in the $i^{\rm th}$ data sample relation ${v^{(i)} = \vec{z}^{(i),T} \vecgreek{\theta} + \xi^{(i)}}$.

The source task is addressed via the minimum $\ell_2$-norm solution to linear regression, namely,   
\begin{align} 
\label{eq:linear regression - source data class}
\widehat{\vecgreek{\theta}} = \argmin_{\vec{r}\in\mathbb{R}^{d}} \left \Vert  \vec{v} - \mtx{Z}\vec{r} \right \Vert _2^2 = \mtx{Z}^{+} \vec{v},
\end{align}
where $\mtx{Z}^{+}$ is the Moore-Penrose pseudoinverse of $\mtx{Z}$. 
The source test error of the solution (\ref{eq:linear regression - source data class}) can be formulated according to the existing literature on linear regression (without transfer learning aspects); see details in Appendix \ref{appendix:subsec:The Test Error of the Source Task}.
The focus of this paper is on transfer learning and, therefore, we do not analyze the source test error. Yet, it is important to note the peak that occurs in the generalization error ${\mathcal{E}_{\rm src}}$ of the source task around $d=\widetilde{n}$ (see (\ref{eq:out of sample error - source task})), namely, at the threshold between the under and over parameterized regimes of the source model\footnote{Note that for the models in this work the number of learned parameters is equal to the dimension of the input data.}.

\subsection{Target Task: Data Model and Relation to Source Task}
\label{subsec: Target Task}

Our interest is in a target task with data
$\left(\vec{x},y\right)\in\mathbb{R}^{d}\times\mathbb{R}$ that follow the model 
\begin{equation}
\label{eq:target data model}
y = \vec{x}^T \vecgreek{\beta} + \epsilon,
\end{equation}
where ${\vec{x}\sim \mathcal{N}\left(\vec{0},\mtx{\Sigma}_{\vec{x}}\right)}$ is a $d$-dimensional Gaussian input vector, ${\epsilon\sim\mathcal{N}\left(0,\sigma_{\epsilon}^2\right)}$ is a Gaussian noise independent of $\vec{x}$, and ${\vecgreek{\beta}\in\mathbb{R}^d}$ is an unknown parameter vector.

The unknown parameter vector of the source task, $\vecgreek{\theta}$, is related to the unknown parameter of the target task, $\vecgreek{\beta}$, by the relation 
\begin{equation}
\label{eq:theta-beta relation}
\vecgreek{\theta} = \mtx{H}\vecgreek{\beta} + \vecgreek{\eta},
\end{equation}
where ${\mtx{H}\in\mathbb{R}^{d\times d}}$ is a fixed (non-random) matrix and ${\vecgreek{\eta}\sim\mathcal{N}\left(\vec{0},\frac{\sigma_{\eta}^2}{d}\mtx{I}_d\right)}$ is a vector of i.i.d. Gaussian noise components with zero mean and variance ${\frac{\sigma_{\eta}^2}{d}}$. The random elements $\vecgreek{\eta}$, $\vec{x}$, $\epsilon$, $\vec{z}$ and $\xi$ are independent. 
The relation in (\ref{eq:theta-beta relation}) recalls a common data model in inverse problems, which in our case relates to the recovery of the true $\vecgreek{\beta}$ from the true $\vecgreek{\theta}$. However, in our setting, we do not have the true $\vecgreek{\theta}$ but only its estimate $\widehat{\vecgreek{\theta}}$ that was learned for the source task purposes. 
Moreover, in this paper we examine learning settings where $\mtx{H}$ can be known or unknown.

While not knowing the true distribution of $\left(\vec{x},y\right)$, the target learning task is performed based on a dataset ${\mathcal{D}\triangleq\Big\lbrace { \left(\vec{x}^{(i)},y^{(i)}\right) }\Big\rbrace_{i=1}^{n}}$ that contains $n$ i.i.d.\ draws of ${\left(\vec{x},y\right)}$ pairs. We denote the $n$ data samples in ${\mathcal{D}}$ using an $n\times d$ matrix of input variables ${\mtx{X}\triangleq \lbrack {\vec{x}^{(1)}, \dots, \vec{x}^{(n)}} \rbrack^{T}}$ and an $n\times 1$ vector of responses ${\vec{y}\triangleq \lbrack{ y^{(1)}, \dots, y^{(n)} } \rbrack^{T}}$. The training data satisfy ${\vec{y} = \mtx{X} \vecgreek{\beta} + \vecgreek{\epsilon}}$ where ${\vecgreek{\epsilon}\triangleq \lbrack{ {\epsilon}^{(1)}, \dots, {\epsilon}^{(n)} } \rbrack^{T}}$ is an unknown noise vector whose $i^{\rm th}$ component ${\epsilon}^{(i)}$ originates in the $i^{\rm th}$ data sample relation ${y^{(i)} = \vec{x}^{{(i)},T} \vecgreek{\beta} + \epsilon^{(i)}}$.

Consider a test input-response pair $\left( { \vec{x}^{(\rm test)}, y^{(\rm test)} } \right)$ that is independently drawn from the $\left(\vec{x},y\right)$ distribution defined above. 
Given the input $\vec{x}^{(\rm test)}$, the target task aims to estimate the response value $y^{(\rm test)}$ by the value ${\widehat{y} \triangleq \vec{x}^{({\rm test}),T}\widehat{\vecgreek{\beta}}}$, where $\widehat{\vecgreek{\beta}}$ is formed using  ${\mathcal{D}}$ in a transfer learning process that also utilizes the source estimate $\widehat{\vecgreek{\theta}}$.  We evaluate the generalization performance of the target task using the test squared error 
\begin{equation}
\label{eq:out of sample error - target data class - beta form}
\mathcal{E} \triangleq \expectation{ \left( \widehat{y} - y^{(\rm test)} \right)^2 } = \sigma_{\epsilon}^2 + \expectation{ \left \Vert { \widehat{\vecgreek{\beta}} - \vecgreek{\beta} } \right \Vert _{\mtx{\Sigma}_{\vec{x}}}^2 }
\end{equation}
where $\left \Vert { \vec{a} } \right \Vert _{\mtx{\Sigma}_{\vec{x}}}^2 = \vec{a}^{T}\mtx{\Sigma}_{\vec{x}}\vec{a}$ for $\vec{a}\in\mathbb{R}^{d}$, and the expectation in the definition of ${\mathcal{E}}$ is with respect to the test data $\left( { \vec{x}^{(\rm test)}, y^{(\rm test)} } \right)$ of the target task and the training data ${{\mathcal{D}}}$, ${\widetilde{\mathcal{D}}}$ of both the target and source tasks. Note that, in a transfer learning process, $\widehat{\vecgreek{\beta}}$ is a function of the training data of both the target and source tasks. A lower value of ${\mathcal{E}}$ reflects better generalization performance of the target task. 

In this paper we study the generalization performance of the target task based on ${n}$ data samples, using $d$ features of the data in the learning process. Then, we analyze the generalization performance with respect to the parameterization level that is determined by the number of samples $n$ and the number of learned parameters $d$. 
In Appendix \ref{appendix:subsec:The Operator H at Different Resolutions} we explain how (in a well-specified setting) the dimension $d$ can be considered as the \textit{resolution} at which we examine the transfer learning problem; and provide corresponding examples for orthonormal and circulant forms of $\mtx{H}$. 

Now we can proceed to the definition of a transfer learning procedure and the analysis of its generalization performance.


~

~

\section{An Intuitive Design to Transfer Learning}
\label{sec:Well-specified Feature Selection}

Let us start by considering a well-specified model that enables the learning of all the $d$ parameters of $\widehat{\vecgreek{\beta}}\in \mathbb{R}^{d}$.
Since the target task is related to the source task by the model (\ref{eq:theta-beta relation}), the optimization of the target task estimate $\widehat{\vecgreek{\beta}}$ can utilize the source task estimate $\widehat{\vecgreek{\theta}}$ that was already computed. 
This means that we consider a learning setting where parameters of the source model are \textit{transferred and adjusted} for the target model learning, and in many cases this can be conceptually perceived as a \textit{fine tuning} strategy. 
Accordingly, we suggest to optimize the parameters of $\widehat{\vecgreek{\beta}}$ via 
\begin{equation} 
\label{eq:well specified - constrained linear regression - target task}
\widehat{\vecgreek{\beta}} = \argmin_{\vec{b}\in\mathbb{R}^{d}} \left \Vert  \vec{y} - \mtx{X}\vec{b} \right \Vert _2^2 + n{\alpha_{\rm TL}} \left \Vert \widetilde{\mtx{H}}\vec{b} - \widehat{\vecgreek{\theta}} \right \Vert _2^2 
\end{equation}
where $\widetilde{\mtx{H}}\in\mathbb{R}^{d\times d}$ takes the role of $\mtx{H}$, which connects ${\vecgreek{\beta}}$ to ${\vecgreek{\theta}}$ in the true task relation (\ref{eq:theta-beta relation}). If $\mtx{H}$ is known, one is likely to set $\widetilde{\mtx{H}}=\mtx{H}$; otherwise (i.e., if $\mtx{H}$ is unknown), one can typically set $\widetilde{\mtx{H}}=\mtx{I}_d$ despite its potential differences from the unknown $\mtx{H}$.  Note that $\widetilde{\mtx{H}}\in\mathbb{R}^{d\times d}$ and $\widehat{\vecgreek{\theta}}$ are fixed in the optimization in (\ref{eq:well specified - constrained linear regression - target task}). The second term in the optimization cost in (\ref{eq:well specified - constrained linear regression - target task}) evaluates the proximity of the target parameters (after processing by $\widetilde{\mtx{H}}$) to the source parameters. 

Setting ${\alpha_{\rm TL}}=0$ in (\ref{eq:well specified - constrained linear regression - target task}) disables the transfer learning aspects and provides a least squares problem whose minimum $\ell_2$-norm solution is ${ \widehat{\vecgreek{\beta}}_{\rm ML2N}=\mtx{X}^{+} \vec{y} }$ and its corresponding test error is 
\begin{equation}
\label{eq:well specified - out of sample error - target task - no transfer learning - alpha is zero - OLS}
\mathcal{E}_{\rm ML2N} = 
\begin{cases}
\mathmakebox[15em][l]{\left({ 1 + \frac{d}{n-d-1} }\right)  \sigma_{\epsilon}^2 }     \text{for } d \le n-2,  
\\
\mathmakebox[15em][l]{\infty} \text{for } n-1 \le d \le n+1,
\\
\mathmakebox[15em][l]{\left({ 1 + \frac{n}{d-n-1} }\right) \sigma_{\epsilon}^2  + \left({ 1 - \frac{n}{d} }\right) \Ltwonorm{\vecgreek{\beta}} }  \text{for } d \ge n+2, 
\end{cases}
\end{equation}		
which is a special case of previous results, e.g., \cite{belkin2020two,dar2020double}.
Yet, the main focus in this paper is on settings where ${\alpha_{\rm TL}} > 0$ and the transfer learning aspects in the optimization (\ref{eq:well specified - constrained linear regression - target task}) are applied.

Let us consider the following assumption. 
\begin{assumption}
	\label{assumption: H is full rank}
	$\widetilde{\mtx{H}}$, $\mtx{H}$ and $\mtx{\Sigma}_{\vec{x}}$ are full rank $d\times d$ real matrices. 
\end{assumption}
Then, based on the full rank of $\widetilde{\mtx{H}}$, the closed-form solution of the target task in (\ref{eq:well specified - constrained linear regression - target task}) for ${\alpha_{\rm TL}}>0$ is 
\begin{equation}
\label{eq:well specified - constrained linear regression - solution - target task - positive alpha}
{\widehat{\vecgreek{\beta}}_{\rm TL}= \left({ \mtx{X}^{T} \mtx{X} + n{\alpha_{\rm TL}}\widetilde{\mtx{H}}^{T} \widetilde{\mtx{H}} }\right)^{-1} \left( { \mtx{X}^{T} \vec{y} + n{\alpha_{\rm TL}}\widetilde{\mtx{H}}^{T} \widehat{\vecgreek{\theta}} }\right).
}
\end{equation}
We study the generalization performance of the \textit{target} task solution with respect to the parameterization level between $d$ and $n$. Accordingly, the learning process is underparameterized when $d<n$, and overparameterized when $d>n$. 

\begin{assumption}[Isotropic prior distribution]
	\label{assumption:isotropic prior on beta - well specified}
	The target parameter vector $\vecgreek{\beta}$ is random and has isotropic Gaussian distribution with zero mean and covariance matrix $\mtx{B}_{d}=\frac{b}{d}\mtx{I}_{d}$ for some constant $b>0$. 
\end{assumption}
Under Assumption \ref{assumption:isotropic prior on beta - well specified}, we will evaluate generalization performance using the test error from (\ref{eq:out of sample error - target data class - beta form}) with an additional expectation over $\vecgreek{\beta}$, i.e., ${\bar{\mathcal{E}}_{\rm TL} \triangleq \expectationwrt{\mathcal{E}_{\rm TL}}{\vecgreek{\beta}}=\sigma_{\epsilon}^2 + \expectation{ \left \Vert { \widehat{\vecgreek{\beta}}_{\rm TL} - \vecgreek{\beta} } \right \Vert _{\Sigma_{\vec{x}}}^2 }}$.

\section{Known $\mtx{H}$: Analysis for Orthonormal $\mtx{H}$ and Isotropic $\mtx{\Sigma}_{\vec{x}}$}
\label{sec:Analysis for H of an Orthonormal Matrix Form}

In this section we mathematically analyze the transfer learning approach for a relatively simple setting where the task relation operator $\mtx{H}$ is known and has an orthonormal structure. This section serves as a starting point towards the more general settings in the following sections. Specifically, in Section \ref{sec:misspecification} we will examine transfer learning in a misspecified setting with a partly known task relation. Eventually, in Section \ref{sec:Analysis for H of a General Form} we will analyze transfer learning with an unknown $\mtx{H}$. 

\subsection{Analysis of the Transfer Learning Approach}
\label{subsec:Analysis for H of an Orthonormal Matrix Form}
We first examine the case of $\widetilde{\mtx{H}}=\mtx{H}=\mtx{\Psi}^{T}$ 
where $\mtx{\Psi}$ is a $d\times d$ real orthonormal matrix; namely, it is a multi-dimensional rotation operator. Let us also consider isotropic feature covariance $\mtx{\Sigma}_{\vec{x}}=\mtx{I}_{d}$. This will let us to obtain a relatively simple analytical characterization of the generalization performance. Later on, in Section \ref{sec:Analysis for H of a General Form}, we will proceed to the more intricate case where $\mtx{H}$ and $\mtx{\Sigma}_{\vec{x}}$ have general forms and $\widetilde{\mtx{H}}$ differs from an unknown $\mtx{H}$. 

Let us denote $\mtx{X}_{\mtx{\Psi}}\triangleq \mtx{X}\mtx{\Psi}^{T}$. Because $\mtx{\Psi}$ is an orthonormal matrix and the rows of $\mtx{X}$ are i.i.d.~from $\mathcal{N}\left({\vec{0},\mtx{I}_{d}}\right)$, $\mtx{X}_{\mtx{\Psi}}$ is a $n\times d$ random matrix with the same distribution as $\mtx{X}$. 

\begin{lemma}
	\label{lemma:well specified - out of sample error - target task -  nonzero alpha - expectation of eigenvalues - orthonormal H}
	Under Assumptions \ref{assumption: H is full rank}-\ref{assumption:isotropic prior on beta - well specified}, and for $\widetilde{\mtx{H}}=\mtx{H}=\mtx{\Psi}^{T}$ where $\mtx{\Psi}$ is a $d\times d$ orthonormal matrix, the expected test error of the solution from (\ref{eq:well specified - constrained linear regression - solution - target task - positive alpha}) for ${\alpha_{\rm TL}}>0$ can be written as 
	\begin{equation}
	\label{eq:well specified - out of sample error - target task - nonzero alpha - expectation of eigenvalues - orthonormal H}
	\bar{\mathcal{E}}_{\rm TL} = 
	\sigma_{\epsilon}^2  + \mathbb{E}\Biggl\{{ \sum_{k=1}^{d} {\frac{ n^2 {{\alpha_{\rm TL}^2}} C_{\rm TL} + \sigma_{\epsilon}^2 \cdot\eigenvalue{\mtx{X}_{\mtx{\Psi}}^{T} \mtx{X}_{\mtx{\Psi}}}{k} }{\left({\eigenvalue{\mtx{X}_{\mtx{\Psi}}^{T} \mtx{X}_{\mtx{\Psi}}}{k} + n{\alpha_{\rm TL}}}\right)^{2} } } }\Biggr\}
	\end{equation}
	where $\eigenvalue{{\mtx{X}}_{\mtx{\Psi}}^{T} \mtx{X}_{\mtx{\Psi}}}{k}$ is the $k^{\rm th}$ eigenvalue of the $d\times d$ matrix $\mtx{X}_{\mtx{\Psi}}^{T} \mtx{X}_{\mtx{\Psi}}$, and the transfer learning aspects are included in
	\begin{equation}
	\label{eq:well specified - out of sample error - target task - nonzero alpha - expectation of eigenvalues - source noise coefficient}
	C_{\rm TL} \triangleq
	\begin{cases}
	\mathmakebox[16em][l] {\frac{\sigma_{\eta}^2}{d} + \frac{\sigma_{\xi}^2}{\widetilde{n}-d-1} }   \text{for } d \le \widetilde{n}-2,  
	\\
	\mathmakebox[16em][l]{\infty} \text{for } \widetilde{n}-1 \le d \le \widetilde{n}+1,
	\\
	\mathmakebox[16em][l] { \left({1-\frac{\widetilde{n}}{d}}\right) \frac{b}{d} + \frac{\widetilde{n}}{d}\left({\frac{\sigma_{\eta}^2}{d} + \frac{\sigma_{\xi}^2}{d-\widetilde{n}-1}}\right) } \text{for } d \ge \widetilde{n}+2.
	\end{cases} 
	\end{equation}	
\end{lemma}
This lemma is proved in Appendix \ref{appendix:subsec:proof of lemma 3.1}. Note that the expectation over the sum in (\ref{eq:well specified - out of sample error - target task - nonzero alpha - expectation of eigenvalues - orthonormal H}) is only with respect to the eigenvalues of $\mtx{X}_{\mtx{\Psi}}^{T} \mtx{X}_{\mtx{\Psi}}$. 
Importantly, note that $C_{\rm TL}$ reflects two aspects that determine the success of the transfer learning process: The first is the distance between the tasks as induced by the noise level $\sigma_{\eta}^2$ in the task relation. The second is the accuracy of the source task solution which is associated with the source data noise level $\sigma_{\xi}^2$ and the source parameterization level corresponding to $({\widetilde{n}},{d})$.

\begin{figure*}[t]
	\centering
	\subfloat[Well specified]{\includegraphics[width=0.4\textwidth]{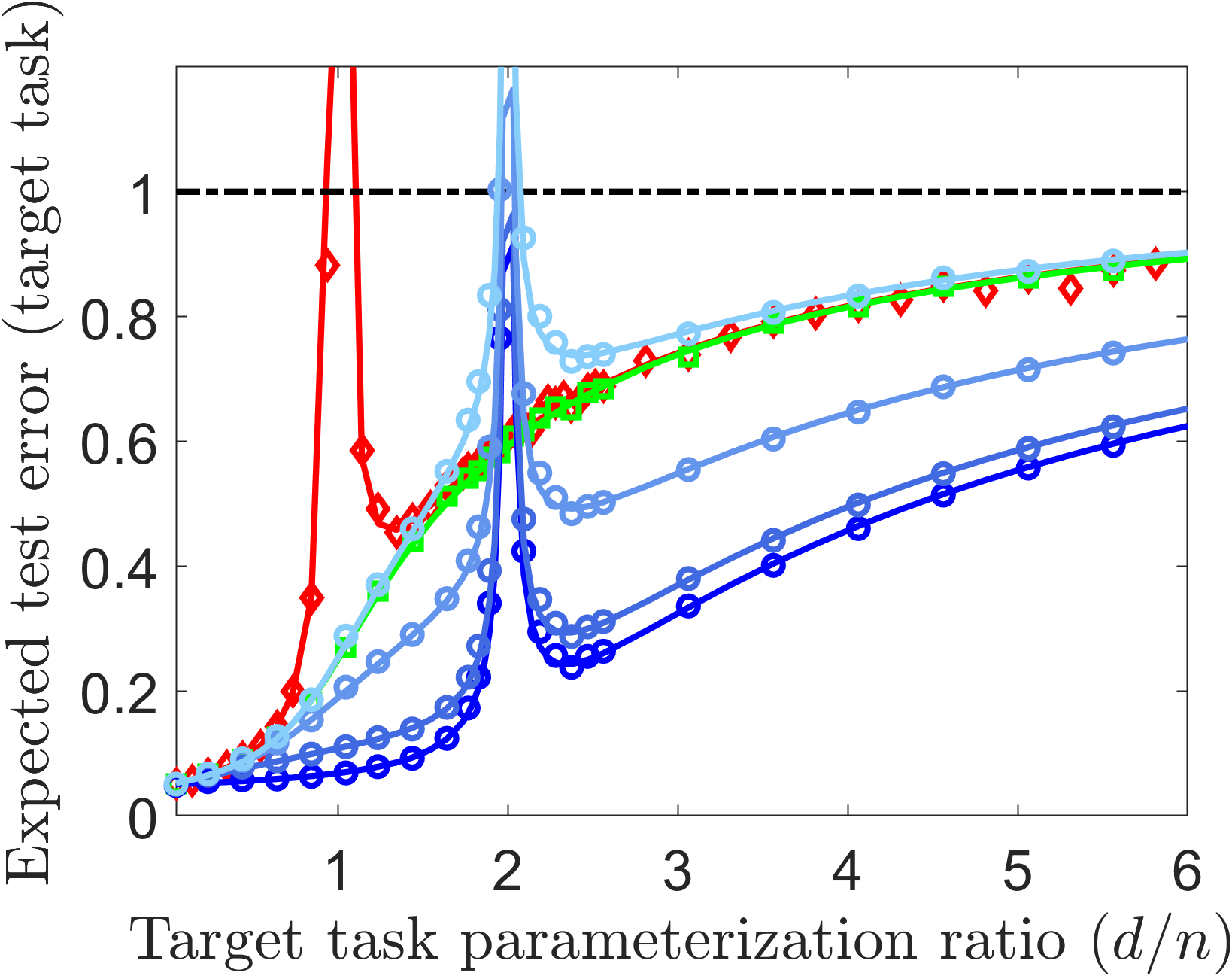}\label{fig:orthonormal_H_error_curves_wellspec}}
	\subfloat[Misspecified]{\includegraphics[width=0.4\textwidth]{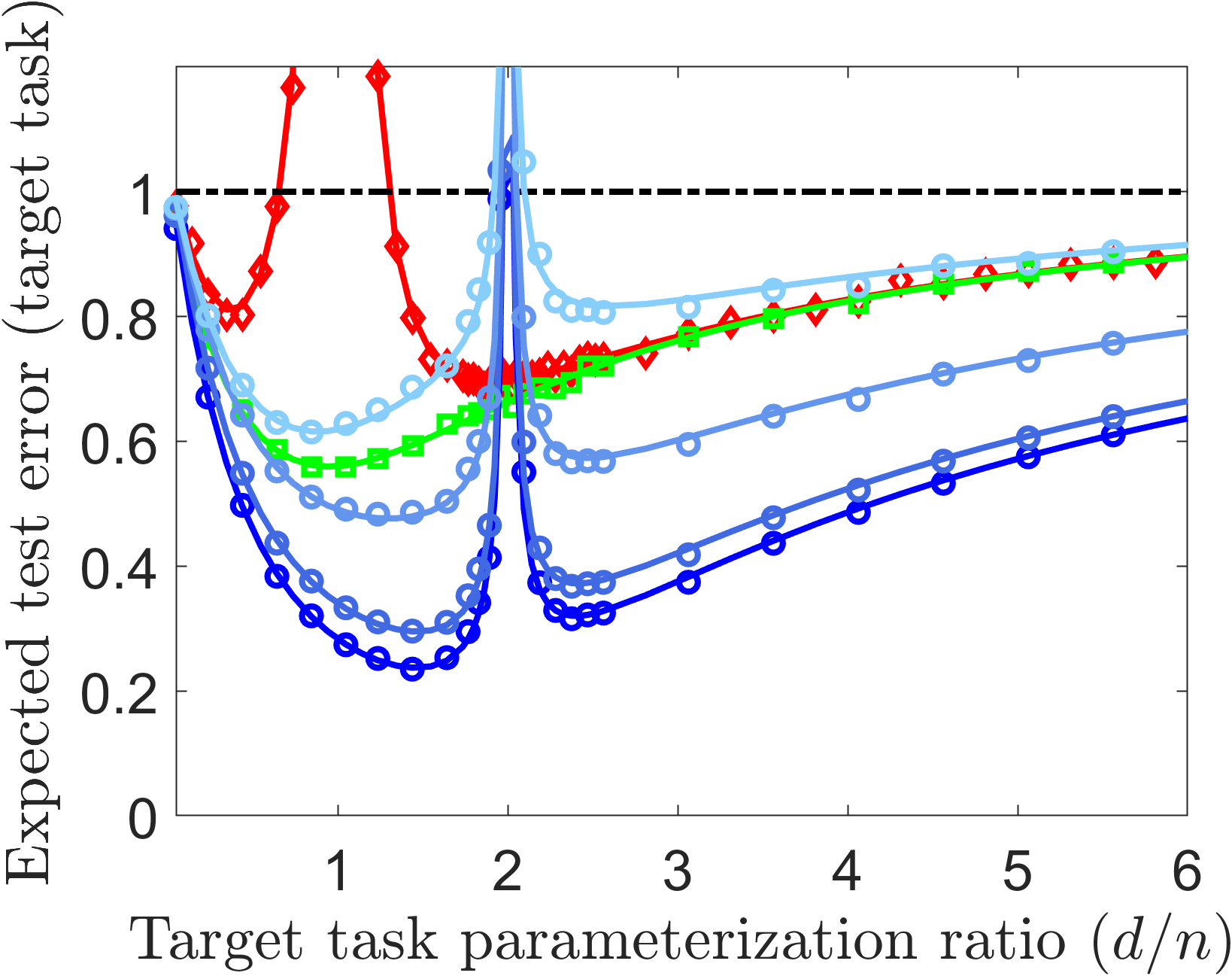}\label{fig:orthonormal_H_error_curves_missspec}}
	\subfloat{\includegraphics[width=0.16\textwidth]{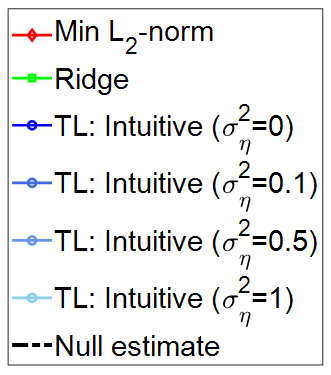}}
	\caption{The test error of the target task under isotropic Gaussian assumption on $\vecgreek{\beta}$ and isotropic target features. Here $\widetilde{\mtx{H}}=\mtx{H}=\mtx{\Psi}^T$ where $\mtx{\Psi}$ is the orthonormal DCT matrix.  Analytical results are presented in solid lines: red curves correspond to minimum $\ell_2$-norm (ML2N) solutions of the target task, green curves correspond to optimally tuned ridge regression, blue curves correspond to optimally tuned transfer learning (TL) in its intuitive form from Section \ref{sec:Well-specified Feature Selection}. The corresponding empirical results (errors averaged over 150 experiments) are denoted by markers in the relevant colors. The number of data samples for the target task is $n=64$ and for the source task is $\widetilde{n}=128$. The misspecified models in \textit{(b)} correspond to Assumptions \ref{assumption:misspecification}-\ref{assumption:Independent misspecification with isotropic features} and polynomial reduction with $a=2.5$, $q=500$, $\rho=2$. }
	\label{fig:error_curves_diagrams_isotropic_H_is_orthonormal}
\end{figure*}

\begin{theorem}
	\label{theorem:well specified - out of sample error - target task -  nonzero alpha - expectation of eigenvalues - optimal - orthonormal H}
	Consider Assumptions \ref{assumption: H is full rank}-\ref{assumption:isotropic prior on beta - well specified} and for $\widetilde{\mtx{H}}=\mtx{H}=\mtx{\Psi}^{T}$ where $\mtx{\Psi}$ is a $d\times d$ orthonormal matrix. 
	The optimal tuning of the transfer learning solution (i.e., ${\alpha_{\rm TL}}>0$) from (\ref{eq:well specified - constrained linear regression - solution - target task - positive alpha}) is achieved for $d\notin\{\widetilde{n}-1,\widetilde{n},\widetilde{n}+1\}$ by setting ${\alpha_{\rm TL}}$ to 
	\begin{equation}
	\label{eq:well specified - out of sample error - target task -  nonzero alpha - optimal alpha - orthonormal H}
	\alpha_{\rm TL}^{\rm opt} = 
	\frac{ \sigma_{\epsilon}^2 }{ nC_{\rm TL} } 
	\end{equation}
	and the corresponding minimal test error is 
	\begin{equation}
	\label{eq:well specified - out of sample error - target task - nonzero alpha - expectation of eigenvalues - optimal - orthonormal H}
	\bar{\mathcal{E}}_{\rm TL}^{\rm opt}= 
	\sigma_{\epsilon}^2
	\left( { 1 + \expectationwrt{  \mtxtrace{ \left( {\mtx{X}_{\mtx{\Psi}}^{T} \mtx{X}_{\mtx{\Psi}} + n\alpha_{\rm TL}^{\rm opt}\mtx{I}_{d}  }\right)^{-1} } }{\mtx{X}_{\mtx{\Psi}} } } \right).
	\end{equation}
	For $d\in\{\widetilde{n}-1,\widetilde{n},\widetilde{n}+1\}$, $\bar{\mathcal{E}}_{\rm TL}=\infty$ for any ${\alpha_{\rm TL}}> 0$. 
\end{theorem}
Theorem \ref{theorem:well specified - out of sample error - target task -  nonzero alpha - expectation of eigenvalues - optimal - orthonormal H} is proved in Appendix \ref{appendix:subsec:proof of theorem 3.2}.
Note that the discontinuity of $C_{\rm TL}$ around $d=\widetilde{n}$ in (\ref{eq:well specified - out of sample error - target task - nonzero alpha - expectation of eigenvalues - source noise coefficient}) is a consequence of the infinite test error of the source model at $d\in\{\widetilde{n}-1,  \widetilde{n}, \widetilde{n}+1\}$ (see the error formulation in (\ref{eq:out of sample error - source task})). Consequently, plugging infinite-valued $C_{\rm TL}$ in the target test error in (\ref{eq:well specified - out of sample error - target task - nonzero alpha - expectation of eigenvalues - orthonormal H}) leads to infinite test error $\bar{\mathcal{E}}_{\rm TL}$ for $d\in\{\widetilde{n}-1,  \widetilde{n}, \widetilde{n}+1\}$. 

To develop the optimal test error from (\ref{eq:well specified - out of sample error - target task - nonzero alpha - expectation of eigenvalues - optimal - orthonormal H}) into a more explicit analytical form, we will make use of an asymptotic setting, which is described next. 
\begin{assumption}[Asymptotic setting]
	\label{assumption:Asymptotic settings - well specified}
	The quantities $d,n,\widetilde{n}\rightarrow\infty$ such that the target task parameterization level ${\frac{d}{n} \rightarrow \gamma_{\rm tgt} \in (0,\infty)}$, and the source task parameterization level $\frac{d}{\widetilde{n}}\rightarrow \gamma_{\rm src}\in (0,\infty)$. 
	The task relation model ${\vecgreek{\theta} = \mtx{H}\vecgreek{\beta} + \vecgreek{\eta}}$ includes an operator $\mtx{H}$ that satisfies ${\frac{1}{d}\Frobnorm{\mtx{H}} \rightarrow \kappa_{\mtx{H}}}$.  
 Moreover, the operator $\widetilde{\mtx{H}}$ satisfies ${\frac{1}{d}\Frobnorm{\widetilde{\mtx{H}}} \rightarrow \kappa_{\widetilde{\mtx{H}}}}$.
\end{assumption}
Our current case of $\mtx{H}$ being an orthonormal matrix, and $\widetilde{\mtx{H}}=\mtx{H}$, implies that $\kappa_{\mtx{H}}=\kappa_{\widetilde{\mtx{H}}}=1$. 

\begin{theorem}
	\label{theorem:well specified - out of sample error - target task -  nonzero alpha - expectation of eigenvalues - optimal - asymptotic isotropic - orthonormal H}
	Consider Assumptions \ref{assumption: H is full rank}-\ref{assumption:Asymptotic settings - well specified}, ${d\notin\{{\widetilde{n}-1},\widetilde{n},{\widetilde{n}+1}\}}$, and $\widetilde{\mtx{H}}=\mtx{H}=\mtx{\Psi}^{T}$ where $\mtx{\Psi}$ is a $d\times d$ orthonormal matrix. Then, the transfer learning form of (\ref{eq:well specified - constrained linear regression - solution - target task - positive alpha}) whose ${\alpha_{\rm TL}}>0$ is optimally tuned to minimize the expected test error ${\bar{\mathcal{E}}_{\rm TL}}$ (i.e., with expectation w.r.t.~the isotropic prior on $\vecgreek{\beta}$) almost surely satisfies 
	\begin{equation}
	\label{eq:well specified - out of sample error - target task - nonzero alpha - expectation of eigenvalues - optimal - asymptotic isotropic - orthonormal H}
	{\bar{\mathcal{E}}_{\rm TL} ^{\rm opt}} \rightarrow 
	\sigma_{\epsilon}^2 
	\left( { 1 + \gamma_{\rm tgt}\cdot m\left( -\alpha_{{\rm TL}, \infty}^{\rm opt}; \gamma_{\rm tgt} \right) }\right)
	\end{equation}
	where 
	\begin{equation}
	\label{eq:well specified - out of sample error - target task -  nonzero alpha - optimal alpha - asymptotic isotropic - orthonormal H}
	\alpha_{{\rm TL}, \infty}^{\rm opt} = 
	\sigma_{\epsilon}^2 \gamma_{\rm tgt} \times 
	\begin{cases}
	\mathmakebox[16em][l]{ \left({\sigma_{\eta}^2 + \frac{\gamma_{\rm src}\cdot \sigma_{\xi}^2 }{1-\gamma_{\rm src}}}\right)^{-1} }   \text{for } \gamma_{\rm src} < 1 
	\\
	\mathmakebox[16em][l]{ \left({ \frac{\gamma_{\rm src}-1}{\gamma_{\rm src}} b + \frac{1}{\gamma_{\rm src}}\left({ \sigma_{\eta}^2 + \frac{\gamma_{\rm src}\cdot \sigma_{\xi}^2 }{\gamma_{\rm src}-1}}\right) }\right)^{-1} } \text{for } \gamma_{\rm src} > 1 
	\end{cases} 
	\end{equation}
	is the limiting value of the optimal ${\alpha_{\rm TL}}>0$,  
	and 
	\begin{align}
	\label{eq:well specified - out of sample error - target task - nonzero alpha - expectation of eigenvalues - optimal - asymptotic isotropic - definition of m function - orthonormal H}
	m\left( -\alpha_{{\rm TL}, \infty}^{\rm opt}; \gamma_{\rm tgt} \right) = \frac{ -\left({ 1 - \gamma_{\rm tgt} + \alpha_{{\rm TL}, \infty}^{\rm opt} }\right) + \sqrt{ \left({ 1 - \gamma_{\rm tgt} + \alpha_{{\rm TL}, \infty}^{\rm opt} }\right)^2 + 4\gamma_{\rm tgt} \alpha_{{\rm TL}, \infty}^{\rm opt} } }{2\gamma_{\rm tgt} \alpha_{{\rm TL}, \infty}^{\rm opt} }
	\end{align}
	is the Stieltjes transform of the Marchenko-Pastur distribution, which is the limiting spectral distribution of the sample covariance associated with $n$ samples that are drawn from a Gaussian distribution $\mathcal{N}\left(\vec{0},\mtx{I}_{d}\right)$. 
\end{theorem}
The proof outline of the last theorem is provided in Appendix \ref{appendix:subsec:proof of theorem 3.3}.
The blue curves in Fig.~\ref{fig:orthonormal_H_error_curves_wellspec} show the analytical formula for the test error ${\bar{\mathcal{E}}_{\rm TL} ^{\rm opt}} $ of the optimally tuned transfer learning under Assumptions \ref{assumption: H is full rank}-\ref{assumption:Asymptotic settings - well specified} and $\widetilde{\mtx{H}}=\mtx{H}$, for instances of the task relation model where $\mtx{H}$ is an orthonormal matrix. Specifically,  Fig.~\ref{fig:orthonormal_H_error_curves_wellspec} shows transfer learning results for several noise variances $\sigma_{\eta}^2$. 

Let us compare the test errors of the examined transfer learning approach with the corresponding errors of the minimum $\ell_2$-norm solution (recall the definition of $\widehat{\vecgreek{\beta}}_{\rm ML2N}$ before (\ref{eq:well specified - out of sample error - target task - no transfer learning - alpha is zero - OLS})) that appear in red curves in Fig.~\ref{fig:orthonormal_H_error_curves_wellspec} (see error formulation in Appendix \ref{appendix:subsec:Generalization Error of OLS Regression Under Isotropic Prior Assumption}). 
One can observe that if the source and target tasks are sufficiently related (in terms of a sufficiently low task relation noise level $\sigma_\eta$), then the intuitive transfer learning approach indeed succeeds to resolve the peak that is induced by the ML2N solution and also to lower the test errors for the majority of parameterization levels. The only exception is that the examined transfer learning solution induces  another peak in the generalization errors of the target task, and the location of this peak is determined by the point where the source task shifts from under to over parameterization. This is a side effect of transferring parameters from the source task, which by itself is a ML2N solution and therefore suffers from a peak of double descent in its own test error curves (see, e.g., (\ref{eq:out of sample error - source task})). 

The test error peak in the examined transfer learning is an example for \textit{negative transfer}, namely, a case where the target model learning can degrade due to using the source model. Specifically, in this work, we can identify a case as negative transfer if the test error for a transfer learning setting is higher than for the minimum $\ell_2$-norm solution and for the optimally tuned ridge regression (the latter will be characterized in Corollary \ref{corollary:well specified - transfer learning is better than ridge - isotropic beta assumption}).
Other mathematical analyses of negative transfer, in other transfer learning methods, are provided in \cite{dar2020double,gerace2021probing}.

\begin{figure*}[t]
	\centering
	\subfloat[Well specified]{\includegraphics[width=0.4\textwidth]{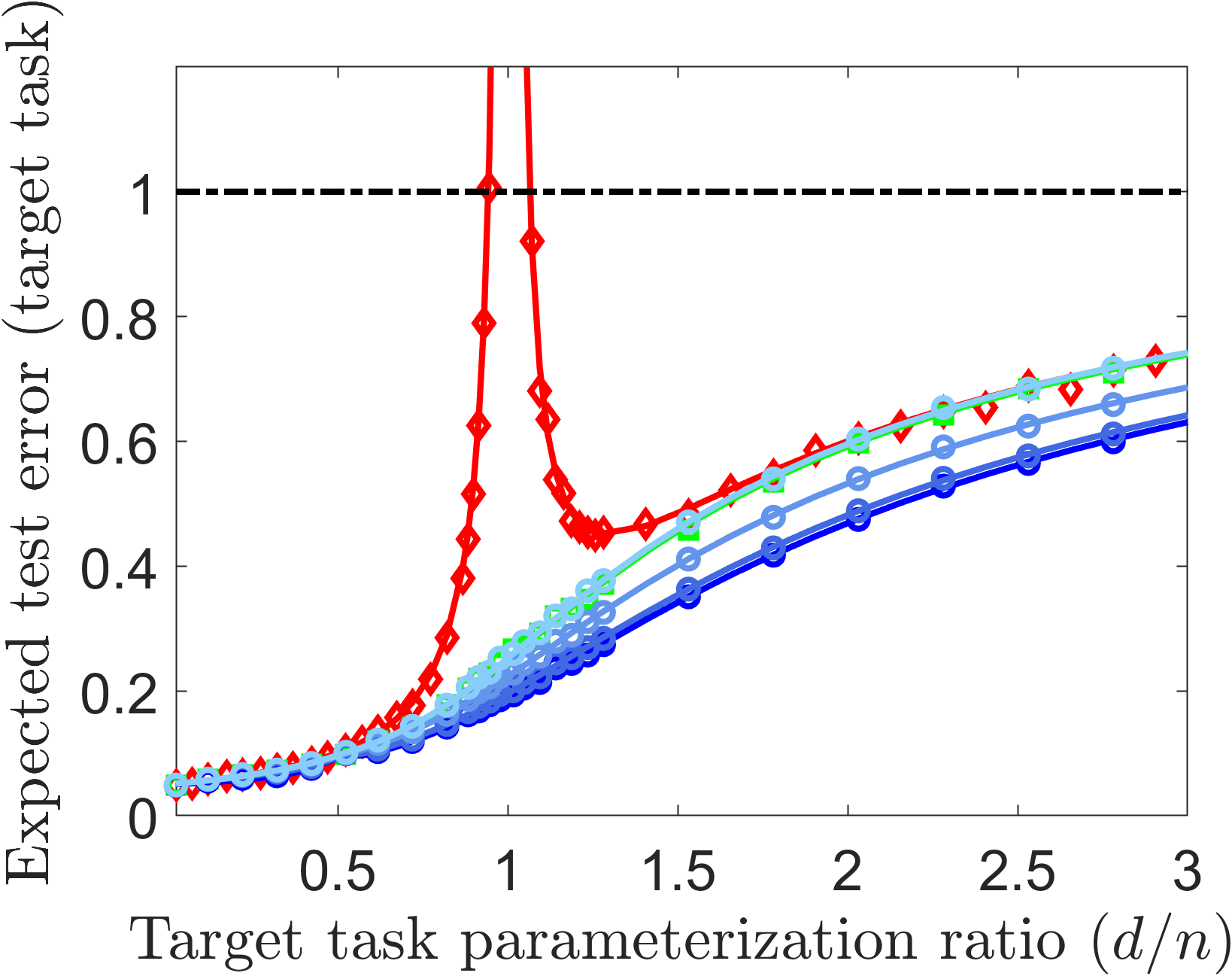}\label{fig:Hdct_HtildeDCT_error_curves_wellspec__ntildeless}}
	\subfloat[Misspecified]{\includegraphics[width=0.4\textwidth]{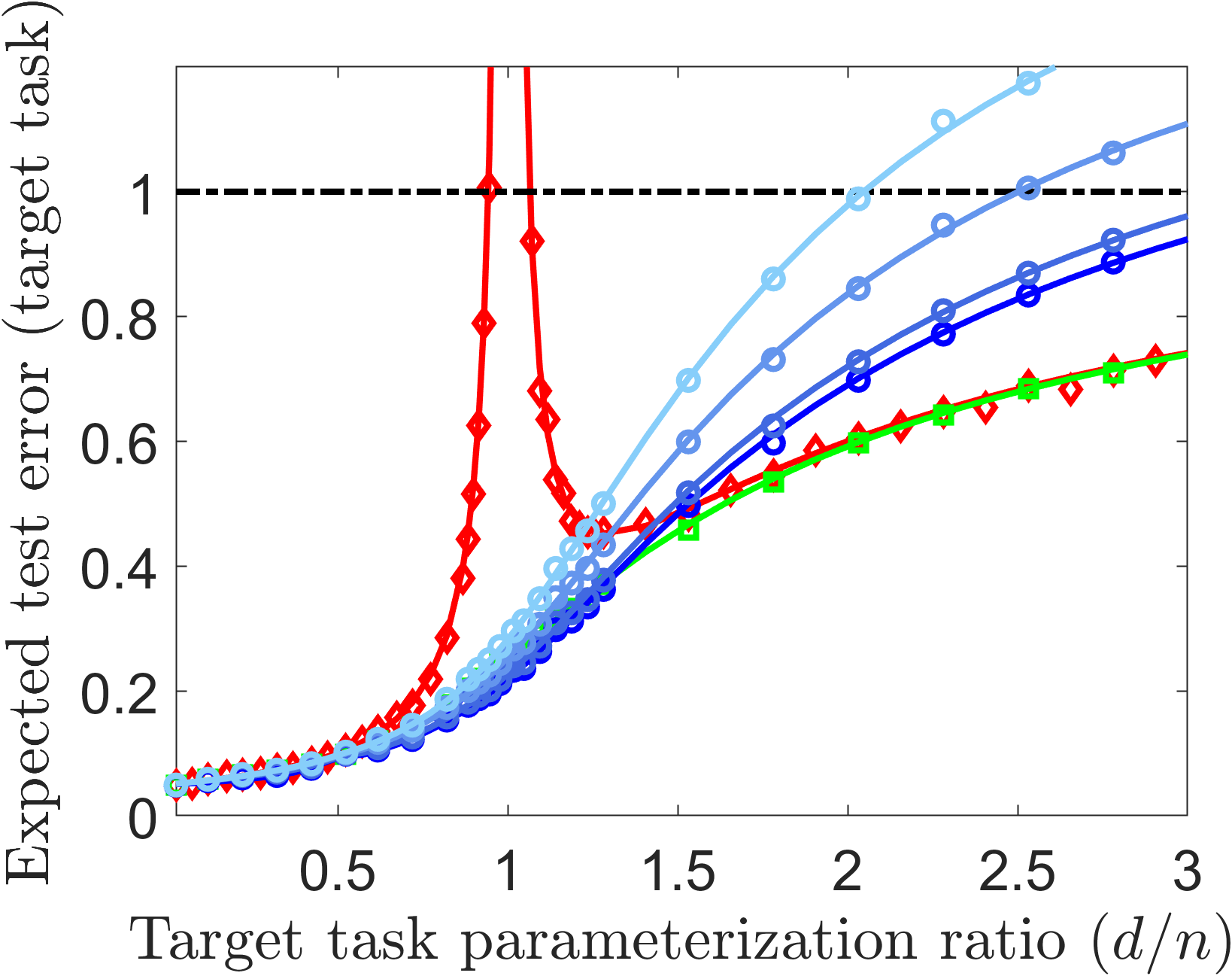}\label{fig:Hgaussian75delta1_HtildeGaussian75delta1_error_curves_wellspec__ntildeless}}
	\subfloat{\includegraphics[width=0.16\textwidth]{figures/error_curve_legend_without_LMMSE.png}}
	\caption{The effect of less source training samples than target training samples, i.e., $n>\widetilde{n}$. Number of data samples for the target task is $n=128$ and for the source task is $\widetilde{n}=64$. The test error of the target task under isotropic Gaussian assumption on $\vecgreek{\beta}$ and isotropic target features. Both subfigures refer to well specified settings with $\widetilde{\mtx{H}}=\mtx{H}$. In (a), $\mtx{H}=\mtx{\Psi}^T$ where $\mtx{\Psi}$ is the orthonormal DCT matrix. In (b), $\mtx{H}$ is a $d\times d$ circulant matrix corresponding to the discrete version of the continuous-domain convolution kernel ${h_{\rm ker}(\tau)=\delta(\tau)+ e^{-\frac{\lvert\tau-0.5\rvert}{w_{\rm ker}}}}$ with ~${w_{\rm ker}=2/75}$. The non-orthonormal $\mtx{H}$ is defined in more detail in Section \ref{subsec:Analysis for H of a General Form - Analysis of the Intuitive Transfer Learning Approach}.}
	\label{fig:error_curves_diagrams_isotropic_H_is_orthonormal__ntilde_less_than_n}
\end{figure*}

\subsection{The Test Error Peak in the Intuitive Transfer Learning: A Closer Look}
\label{sec:The Test Error Peak in the Intuitive Transfer Learning}

It is natural to expect that the transferred source parameters would induce a peak in the target test error around the interpolation threshold of the source task, where the source model itself performs very poorly. Yet, one might also think that the optimal tuning of ${\alpha_{\rm TL}}>0$ in the intuitive transfer learning formula (\ref{eq:well specified - constrained linear regression - solution - target task - positive alpha}) would not perform much worse than the minimum $\ell_2$-norm solution to least squares (recall that setting ${\alpha_{\rm TL}}=0$ in the initial optimization form in (\ref{eq:well specified - constrained linear regression - target task}) degenerates the transfer learning into a least squares problem). We now turn to explain this behavior in more detail.

The formulation in (\ref{eq:well specified - out of sample error - target task -  nonzero alpha - optimal alpha - asymptotic isotropic - orthonormal H}) for the optimal transfer learning parameter shows that as the setting approaches the source task interpolation threshold (namely, $\gamma_{\rm src}\rightarrow 1$ from the left or the right side of the limit), the optimal tuning approaches to zero ($\alpha_{{\rm TL}, \infty}^{\rm opt}\rightarrow 0^+$). 
In the underparameterized regime of the target task, the matrix $\mtx{X}$ is full rank and the limit of the transfer learning  (\ref{eq:well specified - constrained linear regression - solution - target task - positive alpha}) for $\alpha_{{\rm TL}, \infty}^{\rm opt}\rightarrow 0^+$ is the ordinary least squares regression:   
\begin{equation}
\label{eq:limit of intuitive TL - underparameterized target}
\text{For } \gamma_{\rm tgt} < 1:~~~ \lim_{\alpha_{{\rm TL}, \infty}^{\rm opt}\rightarrow 0^+} \widehat{\vecgreek{\beta}}_{\rm TL} =  \left({ \mtx{X}^{T} \mtx{X} }\right)^{-1}  \mtx{X}^{T} \vec{y}.
\end{equation}
However, in the overparameterized regime of the target task, the matrix $\mtx{X}$ is rank deficient. Accordingly, $\mtx{X}^{T} \mtx{X}$ has a $(d-n)$-dimensional null space, whose orthogonal projection matrix is $\mtx{P}_{N(\mtx{X})}\triangleq \mtx{I}_d -  \mtx{X}^+ \mtx{X}$. Then, the transfer learning  (\ref{eq:well specified - constrained linear regression - solution - target task - positive alpha}) for $\alpha_{{\rm TL}, \infty}^{\rm opt}\rightarrow 0^+$ can be written as 
\begin{align}
\label{eq:limit of intuitive TL - overparameterized target}
\text{For } \gamma_{\rm tgt} > 1:~ \lim_{\alpha_{{\rm TL}, \infty}^{\rm opt}\rightarrow 0^+} \widehat{\vecgreek{\beta}}_{\rm TL} & =  \left({ \mtx{X}^{T} \mtx{X} }\right)^{+}  \mtx{X}^{T} \vec{y} + \mtx{P}_{N(\mtx{X})}\left({ \widetilde{\mtx{H}}^{T} \widetilde{\mtx{H}} }\right)^{+}\widetilde{\mtx{H}}^T \widehat{\vecgreek{\theta}} \\
\nonumber
& =   \widehat{\vecgreek{\beta}}_{\rm ML2N} + \mtx{P}_{N(\mtx{X})}\widetilde{\mtx{H}}^{+} \widehat{\vecgreek{\theta}}.
\end{align}
Eq.~(\ref{eq:limit of intuitive TL - overparameterized target}) shows that if the source task interpolation threshold occurs in the overparameterized regime of the target task, the corresponding optimal tuning  (${\alpha_{{\rm TL}, \infty}^{\rm opt}\rightarrow 0^+}$) leads to transfer learning that can significantly deviate from the minimum $\ell_2$-norm solution in the null space of the target data. Moreover, due to the involvement of $\widehat{\vecgreek{\theta}}$, this deviation can increase the transfer learning error and cause a peak in the target test error around the interpolation threshold of the source task (where $\widehat{\vecgreek{\theta}}$ has a poor performance by itself).

Equations (\ref{eq:limit of intuitive TL - underparameterized target})-(\ref{eq:limit of intuitive TL - overparameterized target}) show that transferring the significant test error peak from the source task is possible only in the overparameterized regime of the optimally tuned transfer learning. Accordingly, we show in Fig.~\ref{fig:error_curves_diagrams_isotropic_H_is_orthonormal__ntilde_less_than_n} the error curves for a setting where the target task has more training samples than the source task, namely, $n > \widetilde{n}$. In this setting, the source interpolation threshold resides in the underparameterized regime of the target task (for example, in Fig.~\ref{fig:error_curves_diagrams_isotropic_H_is_orthonormal__ntilde_less_than_n} the source interpolation threshold is at $d/n = 0.5$ whereas the target interpolation threshold is at  $d/n = 1$). Indeed, Fig.~\ref{fig:error_curves_diagrams_isotropic_H_is_orthonormal__ntilde_less_than_n} exemplifies that, for $n > \widetilde{n}$, the optimally tuned intuitive transfer learning does not have a significant error peak around the source interpolation threshold.

Transfer learning is often motivated by insufficient training data for the target task, which is accommodated by using a source model that was trained on a large dataset. Therefore, the case of $n<\widetilde{n}$ is of main interest, and we will focus on it in the rest of this paper.

\subsection{Transfer Learning versus Ridge Regression}
\label{sec:Transfer Learning versus Ridge Regression}

Let us compare the transfer learning of (\ref{eq:well specified - constrained linear regression - target task}) with the standard ridge regression approach, which is independent of the source task and does not involve any transfer learning aspect. 
We start in a non-asymptotic setting; i.e., without Assumption \ref{assumption:Asymptotic settings - well specified}. 
The standard ridge regression approach can be formulated as 
\begin{align} 
\label{eq:well specified - standard ridge resgression}
\widehat{\vecgreek{\beta}}_{\rm ridge} &= \argmin_{\vec{b}\in\mathbb{R}^{d}} \left \Vert  \vec{y} - \mtx{X}\vec{b} \right \Vert _2^2 + n\alpha_{\rm ridge} \left \Vert \vec{b} \right \Vert _2^2 = \left({ \mtx{X}^{T} \mtx{X} + n\alpha_{\rm ridge} \mtx{I}_{d} }\right)^{-1}  { \mtx{X}^{T} \vec{y} }. 
\end{align}
Here $\alpha_{\rm ridge}>0$ is a parameter whose optimal value, which minimizes the expected test error of the target task, is $\alpha_{\rm ridge}^{\rm opt}=\frac{d\sigma_{\epsilon}^2}{n b}$, and the respective test error is 
\begin{equation}
\label{eq:well specified - standard ridge regression - optimal - test error}
\bar{\mathcal{E}}_{\rm ridge}^{\rm opt}= 
\sigma_{\epsilon}^2
\left( { 1 + \expectationwrt{  \mtxtrace{ \left( {\mtx{X}^{T} \mtx{X} + n\alpha_{\rm ridge}^{\rm opt}\mtx{I}_{d}  }\right)^{-1} } }{\mtx{X} } } \right).
\end{equation}
Related results for optimally tuned ridge regression were provided, e.g., in \cite{nakkiran2020optimal,dobriban2018high}. See Appendix \ref{appendix:subsec:Generalization Error of Ridge Regression at Non-Asymptotic Settings} for the proof outline in our notations. 

Note that the test error formulations for the optimally tuned transfer learning for the case of $\widetilde{\mtx{H}}=\mtx{H}$ and an orthonormal $\mtx{H}$ in (\ref{eq:well specified - out of sample error - target task - nonzero alpha - expectation of eigenvalues - optimal - orthonormal H}) and for the optimally tuned ridge regression in (\ref{eq:well specified - standard ridge regression - optimal - test error}) are the same except for the optimal regularization parameters $\alpha_{\rm TL}^{\rm opt}$ and $\alpha_{\rm ridge}^{\rm opt}$, respectively. Accordingly, the cases where transfer learning is better than ridge regression are characterized for non-asymptotic settings as follows (see proof in Appendix \ref{appendix:subsec:proof of corollary 4.1}). 
\begin{corollary}
	\label{corollary:well specified - transfer learning is better than ridge - isotropic beta assumption}
	Consider $\widetilde{\mtx{H}}=\mtx{H}=\mtx{\Psi}^{T}$ where $\mtx{\Psi}$ is a $d\times d$ orthonormal matrix. Then, the test error of optimally tuned transfer learning, $\bar{\mathcal{E}}_{\rm TL}^{\rm opt}$, is lower than the test error of optimally tuned standard ridge regression, $\bar{\mathcal{E}}_{\rm ridge}^{\rm opt}$, if $\alpha_{\rm TL}^{\rm opt} > \alpha_{\rm ridge}^{\rm opt}$, which is satisfied if  	${\sigma_{\eta}^2 + \frac{d\cdot\sigma_{\xi}^2}{|d-\widetilde{n}|-1} <  b }$  for $d \notin \{{\widetilde{n}-1,\widetilde{n},\widetilde{n}+1}\}$,	and never for ${d \in \{{\widetilde{n}-1,\widetilde{n},\widetilde{n}+1}\}}$.
\end{corollary}

The formula for the test error of the optimally tuned ridge regression in asymptotic settings (that is, under Assumptions \ref{assumption:isotropic prior on beta - well specified}-\ref{assumption:Asymptotic settings - well specified}, without our transfer learning and source task aspects) was already provided in \cite{dobriban2018high,hastie2019surprises}. For completeness of presentation we provide this formulation in our notations in Appendix \ref{appendix:subsec:Generalization Error of Ridge Regression at Asymptotic Settings}. 
In Fig.~\ref{fig:orthonormal_H_error_curves_wellspec} we provide the analytical and empirical evaluations of the test error of the optimally tuned ridge regression solution (see green curves and markers). The results exhibit that if the source and target tasks are sufficiently related (e.g., the blue curves that correspond to $\sigma_{\eta}^2=0,0.1,0.5$), then the optimally tuned transfer learning approach outperforms the optimally tuned ridge regression solution for all the parameterization levels besides those in the proximity of the threshold between the under and over parameterized regimes of the source model. 
Remarkably, whereas ridge regression indeed resolves the peak of the double descent of the target task, the examined transfer learning approach can reduce the test errors much further and for a wide range of parameterization levels.


Corollary \ref{corollary:well specified - transfer learning is better than ridge - isotropic beta assumption} characterizes the cases where using a sufficiently related source task is more useful than using the true prior of the desired $\vecgreek{\beta}$. 
Moreover, we consider $\vecgreek{\beta}$ to originate from an isotropic Gaussian distribution, hence, the optimally tuned ridge regression solution is the minimum MSE estimate of the target task parameters; i.e., the solution that minimizes the test error of the target task when only the sample data $\mtx{X},\vec{y}$ are given.
This means that we demonstrated a case where even though optimally tuned ridge regression is the best approach for solving the independent target task, it is not necessarily the best approach when there is an option to utilize transferred parameters from a sufficiently-related source task (e.g., low $\sigma_{\eta}^2$) that was already solved in a sufficiently accurate manner w.r.t.~the source task goal (i.e., low $\sigma_{\xi}^2$ and high $|d-\widetilde{n}|$). 


\section{Misspecified Models: An Example for Beneficial Overparameterization}
\label{sec:misspecification}
Our discussion so far has focused on the learning of well-specified models, namely, where the number of parameters $d$ corresponds to the dimension of both the learned $\widehat{\vecgreek{\beta}}$ and true $\vecgreek{\beta}$ vectors. In learning well-specified models using isotropic features, the test errors in the overparameterized regime are usually higher than in the underparameterized regime (see the detailed discussion in \cite{hastie2019surprises} for ML2N and ridge regression). Indeed, we see this behavior in all the well-specified methods that we examine in Fig.~\ref{fig:orthonormal_H_error_curves_wellspec}. Nevertheless, it is also shown in \cite{hastie2019surprises} that ML2N regression can become highly beneficial in the overparameterized regime when the learned model is misspecified; namely, when the number of learned parameters in $\widehat{\vecgreek{\beta}}$ is lower than the number of parameters in the true $\vecgreek{\beta}$, and that this gap decreases as the learned model becomes more parameterized. 

In our transfer learning setting  we interpret the misspecification aspect as follows. 
\begin{assumption}[Misspecification of the target task]
    \label{assumption:misspecification}
    The data model in (\ref{eq:target data model}) is extended into 
        \begin{equation}
    \label{eq:misspecified data model}
    y = \vec{x}^T \vecgreek{\beta} + {\vec{x}_{\rm ms}}^T \vecgreek{\beta}_{\rm ms} + \epsilon
    \end{equation}
    where the additional features $\vec{x}_{\rm ms}\in\mathbb{R}^q$ and true parameters $\vecgreek{\beta}_{\rm ms}\in\mathbb{R}^q$  are ignored in the learning process that only estimates the $d$-dimensional $\vecgreek{\beta}$ using its corresponding features $\vec{x}$. Moreover, the task relation model in (\ref{eq:theta-beta relation}) is extended into 
    \begin{equation}
    \label{eq:misspecified task relation}
     \vecgreek{\theta} = \mtx{H}\vecgreek{\beta} + {\mtx{H}_{\rm ms}} \vecgreek{\beta}_{\rm ms} + \vecgreek{\eta}   
    \end{equation}
     where $\mtx{H}_{\rm ms}\in\mathbb{R}^{d\times q}$ corresponds to the misspecified parameters $\vecgreek{\beta}_{\rm ms}$. Also, $\vecgreek{\beta}_{\rm ms}$ and $\vec{x}_{\rm ms}$ are independent of $\epsilon$ and $\vecgreek{\eta}$.
\end{assumption}
Importantly, the transfer learning utilizes the operator $\mtx{H}$ but not the additional operator $\mtx{H}_{\rm ms}$ from  (\ref{eq:misspecified task relation}), implying that the misspecified transfer learning uses only a part of the task relation. Specifically, although in this section we still assume a known $\mtx{H}$ and set $\widetilde{\mtx{H}}=\mtx{H}$ in (\ref{eq:well specified - constrained linear regression - solution - target task - positive alpha}), the operator $\mtx{H}_{\rm ms}$ is unknown and not used in the learning. 
\begin{assumption}[Independent misspecification with isotropic features]
    \label{assumption:Independent misspecification with isotropic features}
    Consider a random $\vecgreek{\beta}_{\rm ms}$ which is zero-mean, isotropic, and independent of (the possibly anisotropic) $\vecgreek{\beta}$. The misspecified features $\vec{x}_{\rm ms}\sim\mathcal{N}\left(\vec{0},\mtx{I}_q\right)$ are independent of the other $d$ features in $\vec{x}$. Also, $\mtx{H}_{\rm ms}\mtx{H}_{\rm ms}^T = \rho\mtx{I}_d$ for $\rho\ge0$, which implies that $q\ge d$ and $\mtx{H}_{\rm ms}$ has orthogonal rows. 
\end{assumption}
We assume that $\expectation{\Ltwonorm{\vecgreek{\beta}}+\Ltwonorm{\vecgreek{\beta}_{\rm ms}}}=\omega_{\vecgreek{\beta}_{\rm all}}$ for the same constant $\omega_{\vecgreek{\beta}_{\rm all}}$ for all $d,q$. Then, similarly to \cite{hastie2019surprises}, we assume that the relative misspecification energy reduces polynomially as the parameterization level increases: $\expectation{\Ltwonorm{\vecgreek{\beta}_{\rm ms}}}/\omega_{\vecgreek{\beta}_{\rm all}}=\left(1+\frac{d}{n}\right)^{-a}$ for $a>0$. 
In Appendix \ref{appendix:sec: misspecification details and proofs} we show that under Assumptions \ref{assumption:misspecification}-\ref{assumption:Independent misspecification with isotropic features}, the misspecification effect is equivalent to learning of a well-specified model where the noise levels of $\epsilon$ and $\vecgreek{\eta}$ are effectively higher (but this effective increase reduces with $d$).
Such misspecification signficantly changes the behavior of the test error as a function of the parameterization level. Specifically, overparameterized models (i.e., where $d/n>1$) can achieve the best generalization performance --- for example, see the results for an orthonormal $\mtx{H}$ in Fig.~\ref{fig:orthonormal_H_error_curves_missspec}, the results for a circulant (non-orthonormal) $\mtx{H}$ in Fig.~\ref{fig:error_curves_diagrams_isotropic_general_H__misspecified}, and additional results for stronger misspecification in Fig.~\ref{fig:error_curves_misspec_rho_25} in Appendix \ref{appendix:subsec:Additional Examples for Generalization Performance in Misspecified Settings}. 

\section{Unknown $\mtx{H}$: Analysis for $\mtx{H}$ and $\mtx{\Sigma}_{\vec{x}}$ of General Forms}
\label{sec:Analysis for H of a General Form}

Having provided a detailed analytical characterization for the case where $\mtx{H}$ is a known orthonormal matrix, we now proceed to the more intricate case where the matrix $\mtx{H}$ is unknown and has a general form. 

In this section, we examine the intuitive transfer learning approach from (\ref{eq:well specified - constrained linear regression - solution - target task - positive alpha}) with $\widetilde{\mtx{H}}$ that might differ from the unknown $\mtx{H}$. Then, the main question is how to choose $\widetilde{\mtx{H}}$. Surprisingly, we will show that even the simple choice of $\widetilde{\mtx{H}}=\mtx{I}_d$ can perform well and, in some cases, can also outperform the seemingly better option of $\widetilde{\mtx{H}}=\mtx{H}$. 

\subsection{Analysis of the Intuitive Transfer Learning Approach}
\label{subsec:Analysis for H of a General Form - Analysis of the Intuitive Transfer Learning Approach}

Recall that in our transfer learning approach (\ref{eq:well specified - constrained linear regression - target task}) we do not have the true $\vecgreek{\theta}$ but its estimate $\widehat{\vecgreek{\theta}}$, which is the ML2N solution to the source task. 
The benefits from utilizing $\widehat{\vecgreek{\theta}}$ in our transfer learning process (\ref{eq:well specified - constrained linear regression - target task}) are affected by the distribution of the transferred source parameters $\widehat{\vecgreek{\theta}}$. The second-order statistics of the distribution of $\widehat{\vecgreek{\theta}}$ given the true target parameters $\vecgreek{\beta}$ are formulated as follows (the proof is provided in Appendix \ref{appendix:subsec:proof of proposition 3.4}).
\begin{proposition}
	\label{proposition:moments of theta given beta}	
	The expected value of $\widehat{\vecgreek{\theta}}$ given $\vecgreek{\beta}$ is 
	\begin{equation}
	\expectation{\widehat{\vecgreek{\theta}} \big| \vecgreek{\beta}} = 
	\begin{cases}
	\mathmakebox[6em][l] {\mtx{H}\vecgreek{\beta} }   \text{for } d \le \widetilde{n},  
	\\
	\mathmakebox[6em][l] {\frac{\widetilde{n}}{d}\mtx{H}\vecgreek{\beta} }   \text{for } d > \widetilde{n}.
	\end{cases} 
	\end{equation}
	The covariance matrix of $\widehat{\vecgreek{\theta}}$ given $\vecgreek{\beta}$, namely, $\mtx{C}_{\widehat{\vecgreek{\theta}} | \vecgreek{\beta}} \triangleq \expectation{ \left( \widehat{\vecgreek{\theta}} - \expectation{\widehat{\vecgreek{\theta}} \big| \vecgreek{\beta}}\right)\left( \widehat{\vecgreek{\theta}} - \expectation{\widehat{\vecgreek{\theta}} \big| \vecgreek{\beta}}\right)^{T} \Big| \vecgreek{\beta}}$, is 
	\begin{equation}
	\label{eq:covariance of theta give beta - underparameterized source task}
	 \mtx{C}_{\widehat{\vecgreek{\theta}} | \vecgreek{\beta}} = \left(\frac{\sigma_{\eta}^2}{d} + \frac{\sigma_{\xi}^{2}}{\widetilde{n}-d-1} \right)\mtx{I}_{d}
	\end{equation} 
	for $d\le \widetilde{n}-2$, and 
	\begin{align}
	\label{eq:covariance of theta give beta - overparameterized source task}
	& \resizebox{.93\hsize}{!}{$\mtx{C}_{\widehat{\vecgreek{\theta}} | \vecgreek{\beta}} = \frac{\widetilde{n}}{d}\left({ \frac{d-\widetilde{n}}{d(d+1)}\mtx{H}\vecgreek{\beta}\vecgreek{\beta}^{T}\mtx{H}^{T} +  \frac{d-\widetilde{n}}{d^2 -1} {\rm diag}\left({ \{ \Ltwonorm{\mtx{H}\vecgreek{\beta}} - \left(\{\mtx{H}\vecgreek{\beta}\}_{j}\right)^2 \}_{j=1,\dots,d} }\right) + \left(\frac{\sigma_{\eta}^2}{d}  + \frac{\sigma_{\xi}^{2}}{d-\widetilde{n}-1} \right)\mtx{I}_{d}}\right)$}
	\end{align} 
	for $d \ge \widetilde{n}+2$. 
	For $d\in\{\widetilde{n}-1,\widetilde{n},\widetilde{n}+1\}$ the covariance matrix is infinite valued. 

 In (\ref{eq:covariance of theta give beta - overparameterized source task}), ${\{\mtx{H}\vecgreek{\beta}\}_{j}}$ is the $j^{\rm th}$ component of the vector $\mtx{H}\vecgreek{\beta}$. The notation ${\rm diag}\left(\cdot\right)$ refers to the $d\times d$ diagonal matrix whose main diagonal values are specified as the arguments of ${\rm diag}\left(\cdot\right)$.
\end{proposition}
Proposition \ref{proposition:moments of theta given beta} demonstrates two very different forms of the covariance of $\widehat{\vecgreek{\theta}}$: For an underparameterized source task with $d\le\widetilde{n}-2$, the covariance is isotropic with a simple diagonal form that does not reflect $\vecgreek{\beta}$ nor $\mtx{H}$. However, for an overparameterized source task with $d\ge\widetilde{n}+2$, the covariance can be anisotropic with a form that depends on $\vecgreek{\beta}$ and $\mtx{H}$. This is a  consequence of the ML2N solution to the source task, which by itself has two different error forms in its under- and over-parameterized cases. 


Next, we turn to formulate the asymptotic generalization error for any asymptotic parameterization level $\frac{d}{n}\rightarrow\gamma_{\rm tgt} \in (0, \infty)$. The proof is provided in Appendix \ref{appendix:subsec:proof of theorem 5}.
\begin{theorem}
	\label{theorem:well specified - out of sample error - target task -  asymptotic - anisotropic target features - general H}
	Consider Assumptions \ref{assumption: H is full rank}-\ref{assumption:isotropic prior on beta - well specified}, and general forms of  $\widetilde{\mtx{H}}$, $\mtx{H}$ and $\mtx{\Sigma}_{\vec{x}}$.  Then, the transfer learning form of (\ref{eq:well specified - constrained linear regression - solution - target task - positive alpha}) with a (not necessarily optimal) parameter ${\alpha_{\rm TL}}>0$ almost surely satisfies 
	\begin{align}
	\label{eq:well specified - out of sample error - target task -  asymptotic - anisotropic target features - general H}
	&{\bar{\mathcal{E}}_{\rm TL} ^{\rm opt}} \rightarrow 
	\sigma_{\epsilon}^2 
	\left(  1 + \gamma_{\rm tgt}\cdot \mtxtrace{\frac{1}{d}\mtx{W}\left( c(\alpha_{\rm TL})\mtx{W}+\alpha_{\rm TL}\mtx{I}_d\right)^{-1} } \right.
	\\ \nonumber
	& \left. +\gamma_{\rm tgt}\cdot \mtxtrace{\left({\frac{\alpha_{\rm TL}^2}{\gamma_{\rm tgt}\sigma_{\epsilon}^2} \mtx{\Gamma}_{{\rm TL},\infty} - \frac{\alpha_{\rm TL}}{d}\mtx{I}_d }\right) \left( c(\alpha_{\rm TL})\mtx{W}+\alpha_{\rm TL}\mtx{I}_d\right)^{-1} s \mtx{W}\left( c(\alpha_{\rm TL})\mtx{W}+\alpha_{\rm TL}\mtx{I}_d\right)^{-1}   }	\right)
	\end{align}
	where $\mtx{W} \triangleq \left({\widetilde{\mtx{H}}^{-1}}\right)^{T}\mtx{\Sigma}_{\vec{x}}\widetilde{\mtx{H}}^{-1}$, $s\triangleq c'(\alpha_{\rm TL}) + 1$, 
  \begin{align}
	    \label{eq:well specified - out of sample error - target task -  asymptotic - anisotropic target features - general H - definition of Gamma_TLinf}
	    &\mtx{\Gamma}_{{\rm TL},\infty} \triangleq \begin{cases}
	\mathmakebox[24em][l]{ \frac{1}{d}\left({\sigma_{\eta}^2 + \frac{\gamma_{\rm src}\cdot \sigma_{\xi}^2 }{1-\gamma_{\rm src}}}\right)\mtx{I}_{d} + \frac{b}{d}\left({\mtx{H}-\widetilde{\mtx{H}}}\right)\left({\mtx{H}-\widetilde{\mtx{H}}}\right)^T }   \text{for } d \le \widetilde{n}-2,  
	\\
	\mathmakebox[24em][l]{\infty}\text{for } \widetilde{n}-1 \le d \le \widetilde{n}+1,
	\\
	\mathmakebox[24em][l]{ \frac{b(\gamma_{\rm src}-1)}{d \gamma_{\rm src}^2 }\left({ \gamma_{\rm src}\widetilde{\mtx{H}}\widetilde{\mtx{H}}^T - \mtx{H}\mtx{H}^T + \kappa_{\mtx{H}}\mtx{I}_{d}-\frac{1}{d}{\rm diag}\left({ \{ \left[\mtx{H}\mtx{H}^{T}\right]_{jj} \}_{j=1,\dots,d} }\right)}\right) }
        \\
	\mathmakebox[24em][l]{+ \frac{b}{d\gamma_{\rm src}}\left({\mtx{H}-\widetilde{\mtx{H}}}\right)\left({\mtx{H}-\widetilde{\mtx{H}}}\right)^T + \frac{1}{d\gamma_{\rm src}}\left({\sigma_{\eta}^2 + \frac{\gamma_{\rm src}\cdot \sigma_{\xi}^2 }{\gamma_{\rm src}-1}}\right)\mtx{I}_{d}} \text{for } d \ge \widetilde{n}+2, 
	\end{cases} 
	\end{align}
	$c(\alpha_{\rm TL})$ is obtained as the solution of $ \frac{1}{c(\alpha_{\rm TL})} - 1 = \frac{\gamma_{\rm tgt}}{d}\mtxtrace{\mtx{W}\left({ c(\alpha_{\rm TL})\mtx{W}+\alpha_{\rm TL}\mtx{I}_d}\right)^{-1}}$,
    and then
    \begin{equation}
    \label{eq:c_tag definition}
    c'(\alpha_{\rm TL}) = \frac{\frac{\gamma_{\rm tgt}}{d} \Frobnorm{\mtx{W}\left({ c(\alpha_{\rm TL})\mtx{W}+\alpha_{\rm TL}\mtx{I}_d}\right)^{-1}} }{(c(\alpha_{\rm TL}))^{-2} - \frac{\gamma_{\rm tgt}}{d} \Frobnorm{\mtx{W}\left({ c(\alpha_{\rm TL})\mtx{W}+\alpha_{\rm TL}\mtx{I}_d}\right)^{-1}} } 
    \end{equation}
where $\Frobnormnonsquared{\cdot}$ is the Frobenius norm.
\end{theorem}

\begin{figure*}[t]
		\subfloat[Known $\mtx{H}$ (${w_{\rm ker}=2/75}$): $\widetilde{\mtx{H}}=\mtx{H}$]{\includegraphics[width=0.4\textwidth]{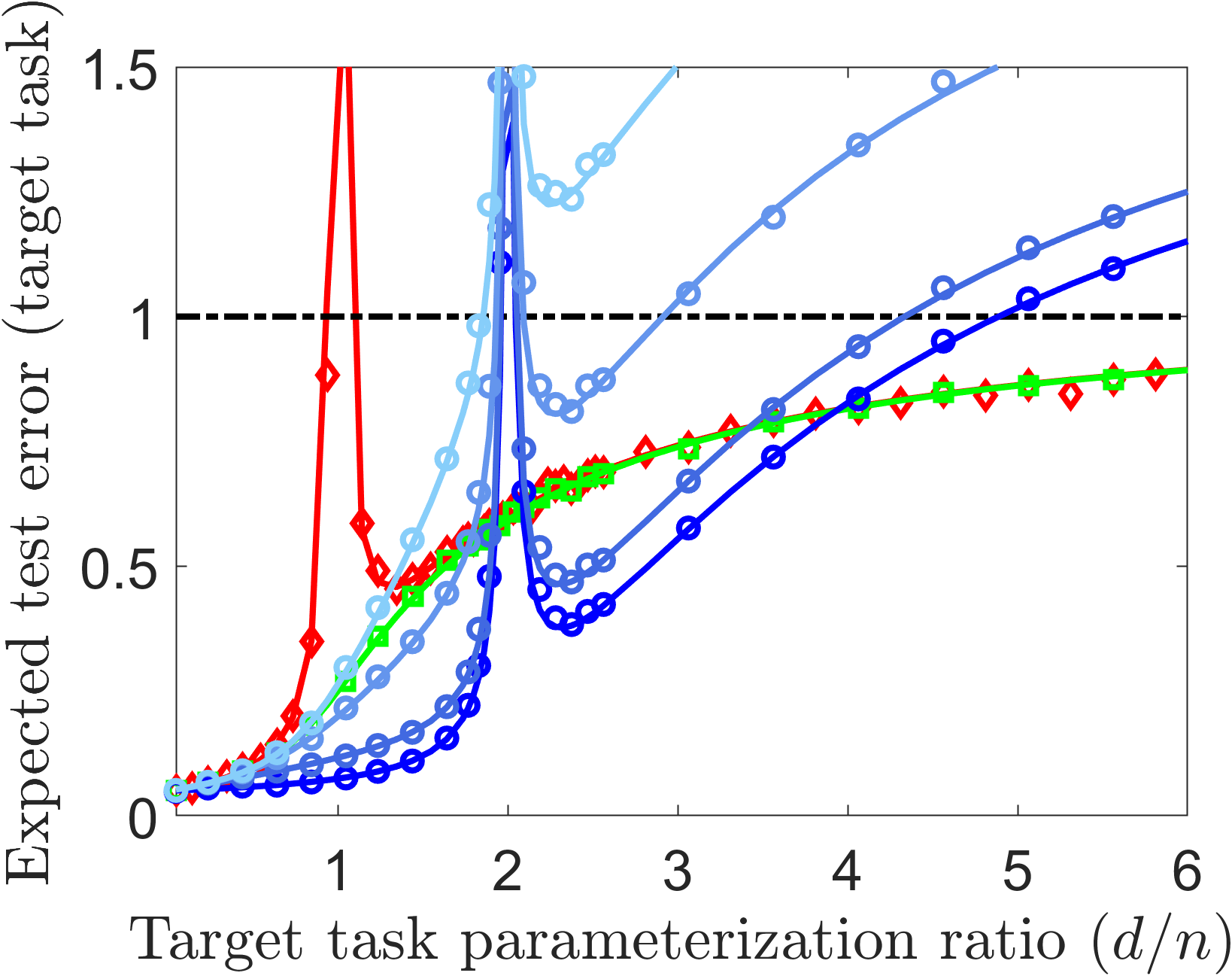}\label{fig:Hgaussian75delta1_HtildeGaussian75delta1_error_curves_wellspec}}
	\subfloat[Unknown $\mtx{H}$ (${w_{\rm ker}=2/75}$): $\widetilde{\mtx{H}}=\mtx{I}_d$]{\includegraphics[width=0.4\textwidth]{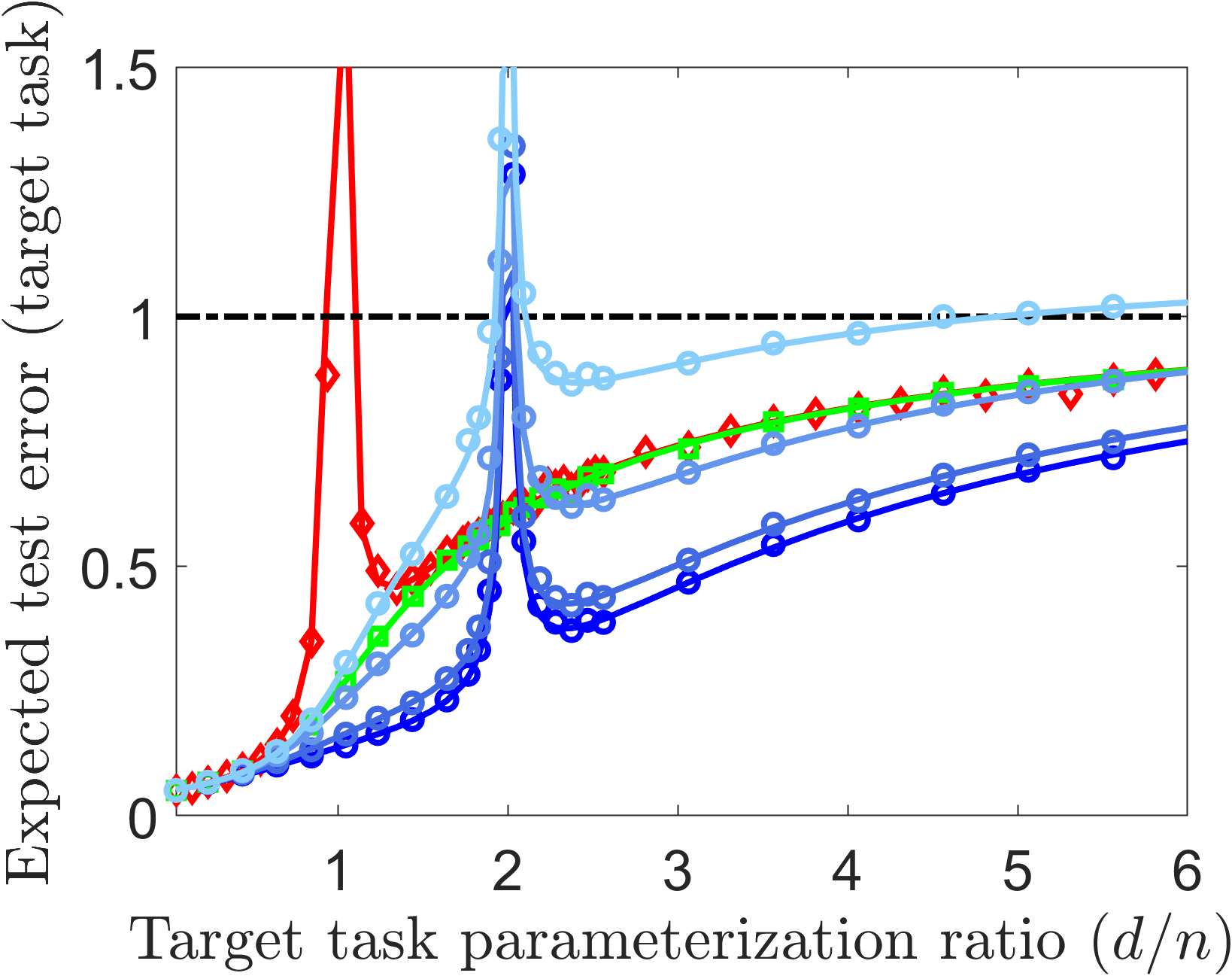}\label{fig:Hgaussian75delta1_HtildeI_error_curves_wellspec}}	
 \\
 	\subfloat[Known $\mtx{H}$ (${w_{\rm ker}=2/25}$): $\widetilde{\mtx{H}}=\mtx{H}$]{\includegraphics[width=0.4\textwidth]{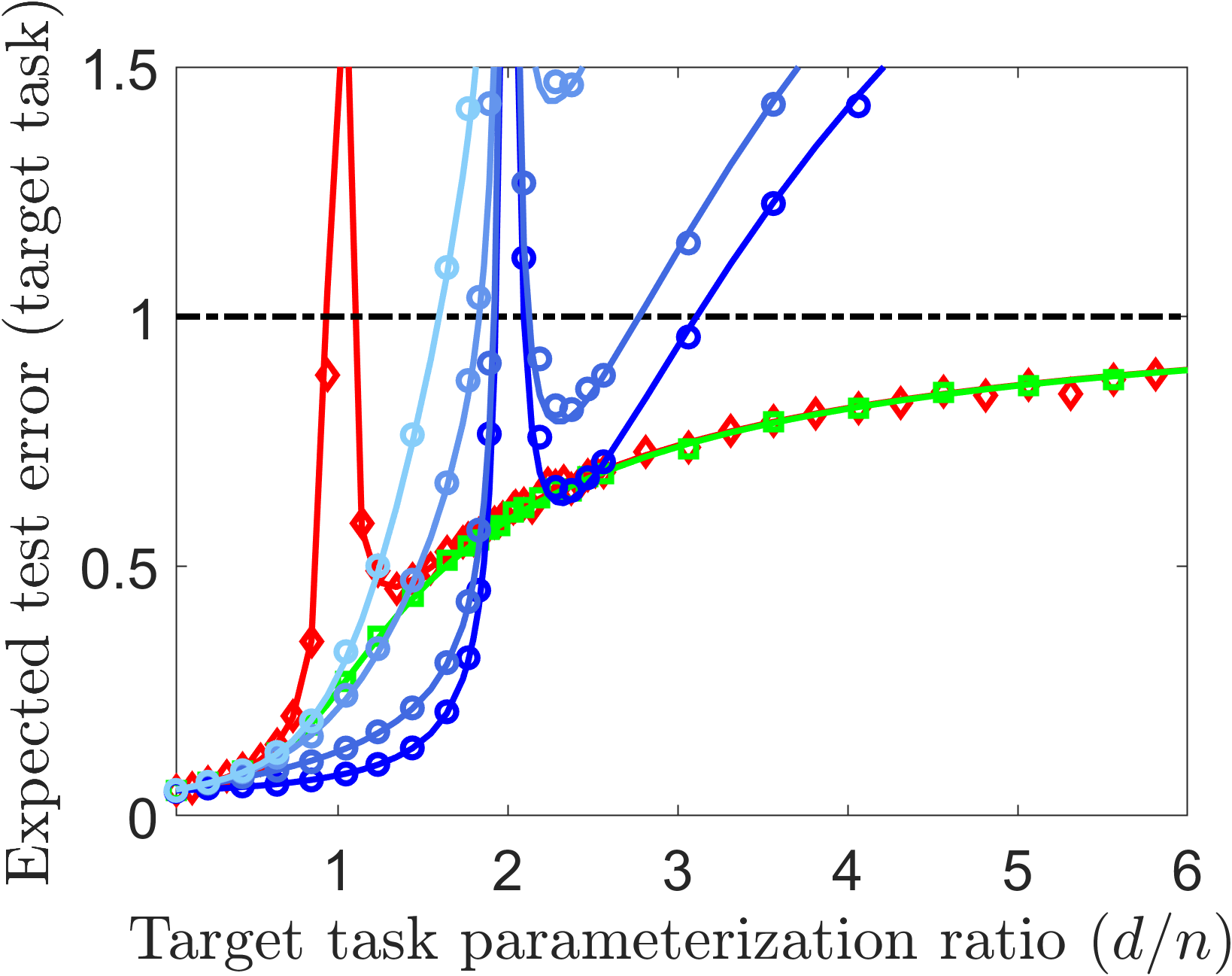}\label{fig:Hgaussian25delta1_HtildeGaussian25delta1_error_curves_wellspec}}
	\subfloat[Unknown $\mtx{H}$ (${w_{\rm ker}=2/25}$): $\widetilde{\mtx{H}}=\mtx{I}_d$]{\includegraphics[width=0.4\textwidth]{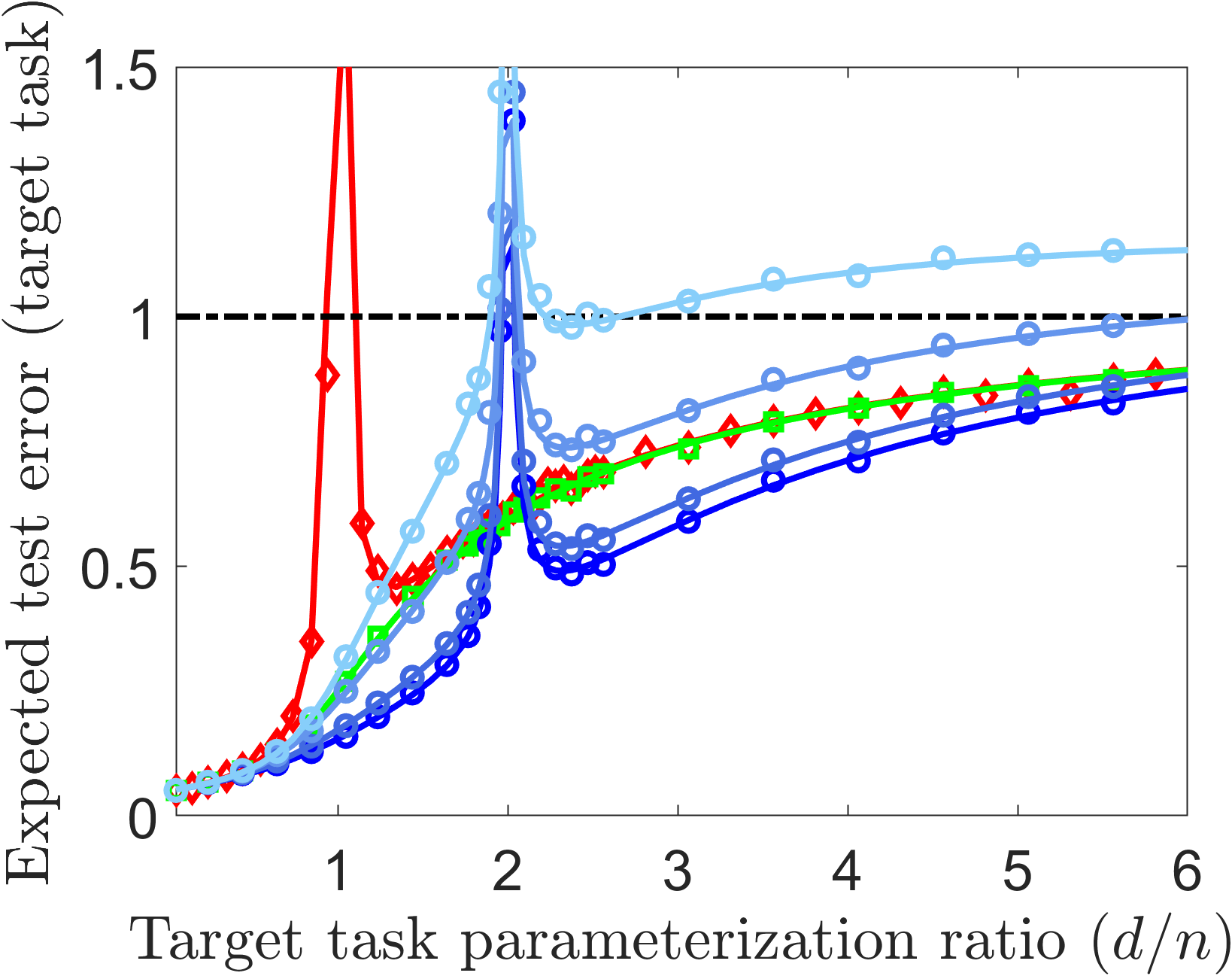}\label{fig:Hgaussian25delta1_HtildeI_error_curves_wellspec}}
	\subfloat{\includegraphics[width=0.16\textwidth]{figures/error_curve_legend_without_LMMSE.png}}
	\caption{The test error of the target task under isotropic Gaussian assumption on $\vecgreek{\beta}$ and isotropic target features in a \textbf{well specified} setting. The matrix $\mtx{H}$ is a $d\times d$ circulant matrix corresponding to the discrete version of the continuous-domain convolution kernel ${h_{\rm ker}(\tau)=\delta(\tau)+ e^{-\frac{\lvert\tau-0.5\rvert}{w_{\rm ker}}}}$, here the kernel width is ~${w_{\rm ker}=2/75}$ in (a)-(b) and ${w_{\rm ker}=2/25}$ in (c)-(d). Curve colors and markers are as in Fig.~\ref{fig:error_curves_diagrams_isotropic_H_is_orthonormal}. The number of data samples for the target task is $n=64$ and for the source task is $\widetilde{n}=128$.}
	\label{fig:error_curves_diagrams_isotropic_general_H__well_specified}
\end{figure*}

\begin{figure*}[t]
		\subfloat[Known $\mtx{H}$ (${w_{\rm ker}=2/75}$): $\widetilde{\mtx{H}}=\mtx{H}$]{\includegraphics[width=0.4\textwidth]{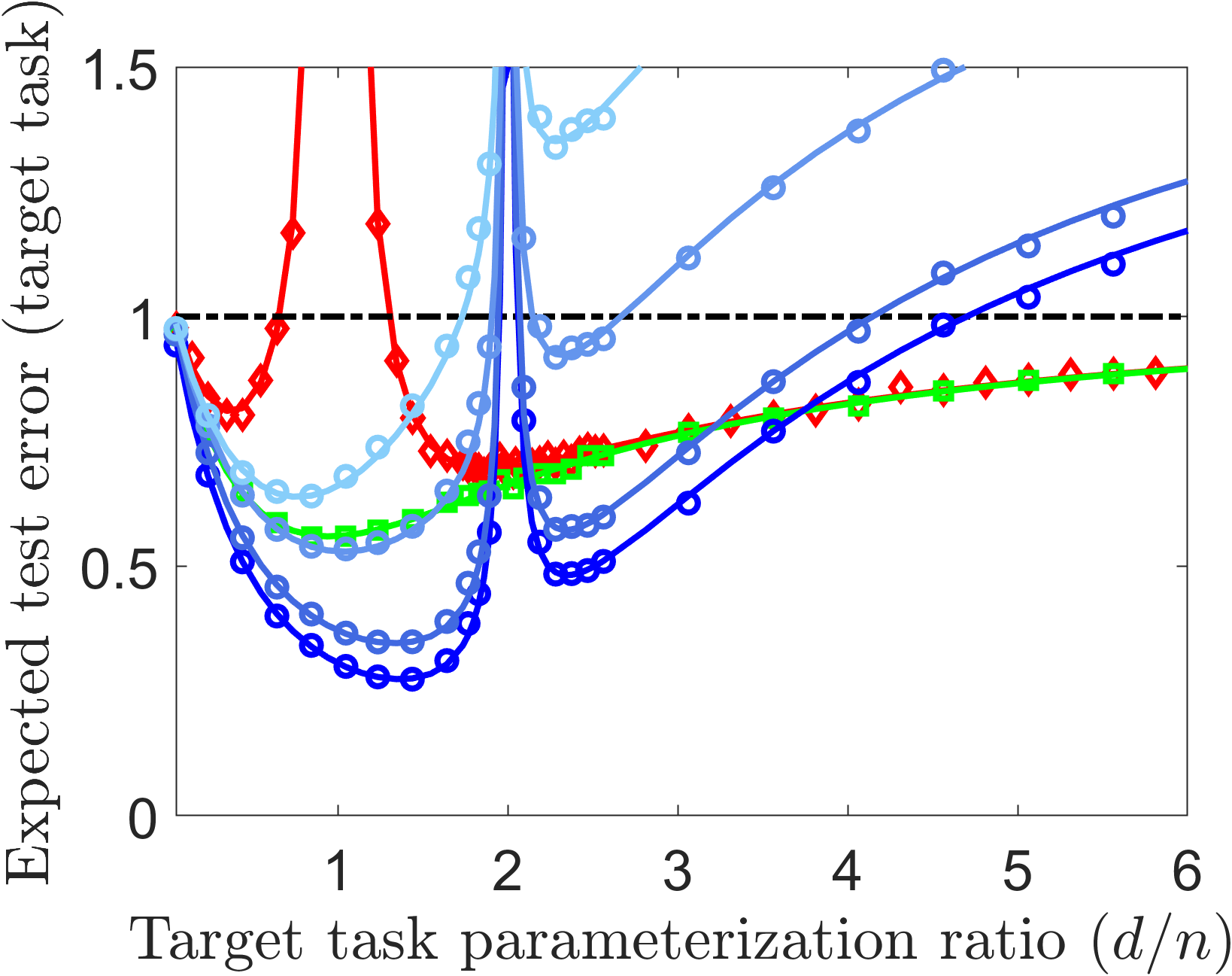}\label{fig:Hgaussian75delta1_HtildeGaussian75delta1_error_curves_misspec_Hrho2}}
	\subfloat[Unknown $\mtx{H}$ (${w_{\rm ker}=2/75}$): $\widetilde{\mtx{H}}=\mtx{I}_d$]{\includegraphics[width=0.4\textwidth]{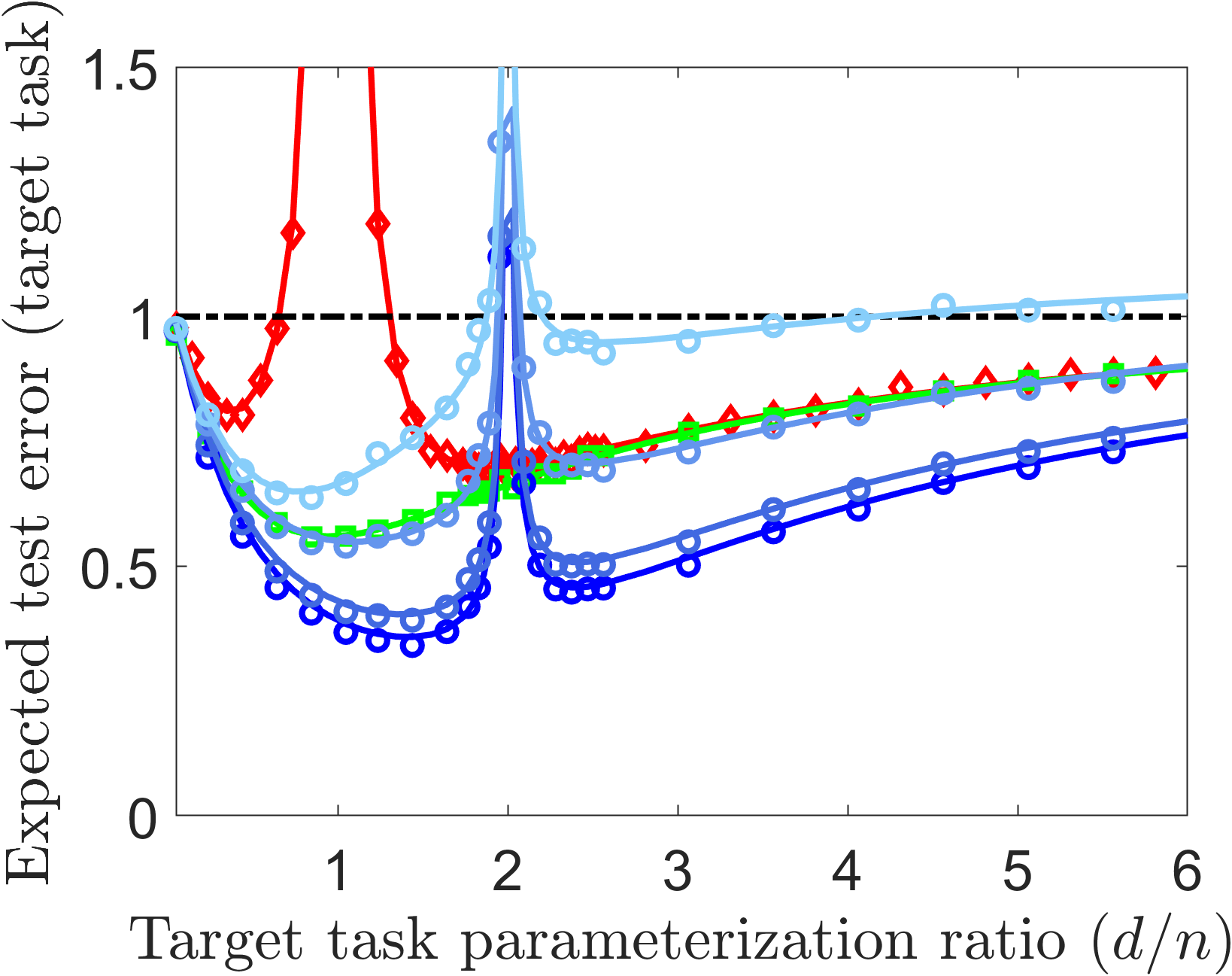}\label{fig:Hgaussian75delta1_HtildeI_error_curves_misspec_Hrho2}}	
 \\
 	\subfloat[Known $\mtx{H}$ (${w_{\rm ker}=2/25}$): $\widetilde{\mtx{H}}=\mtx{H}$]{\includegraphics[width=0.4\textwidth]{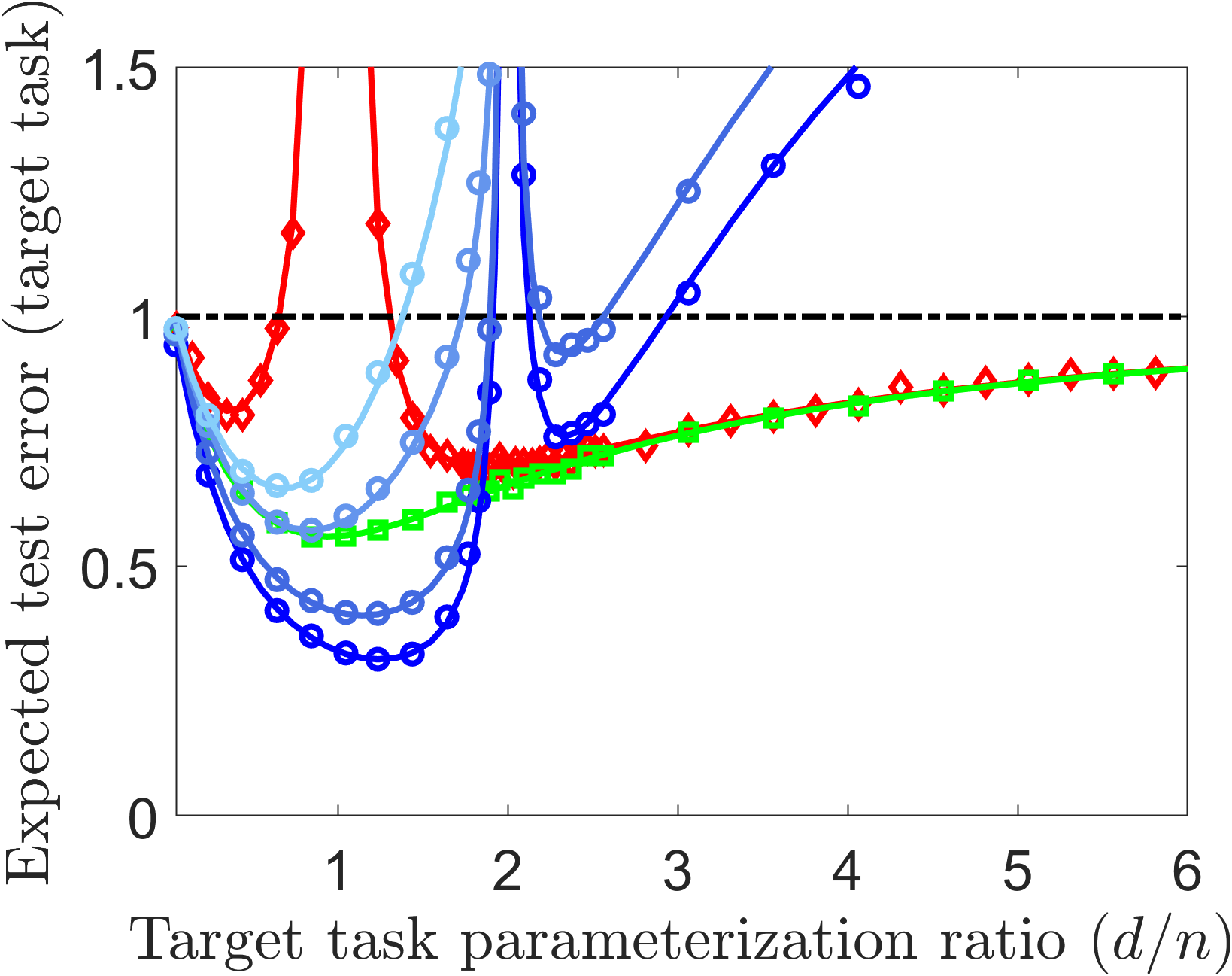}\label{fig:Hgaussian25delta1_HtildeGaussian25delta1_error_curves_misspec_Hrho2}}
	\subfloat[Unknown $\mtx{H}$ (${w_{\rm ker}=2/25}$): $\widetilde{\mtx{H}}=\mtx{I}_d$]{\includegraphics[width=0.4\textwidth]{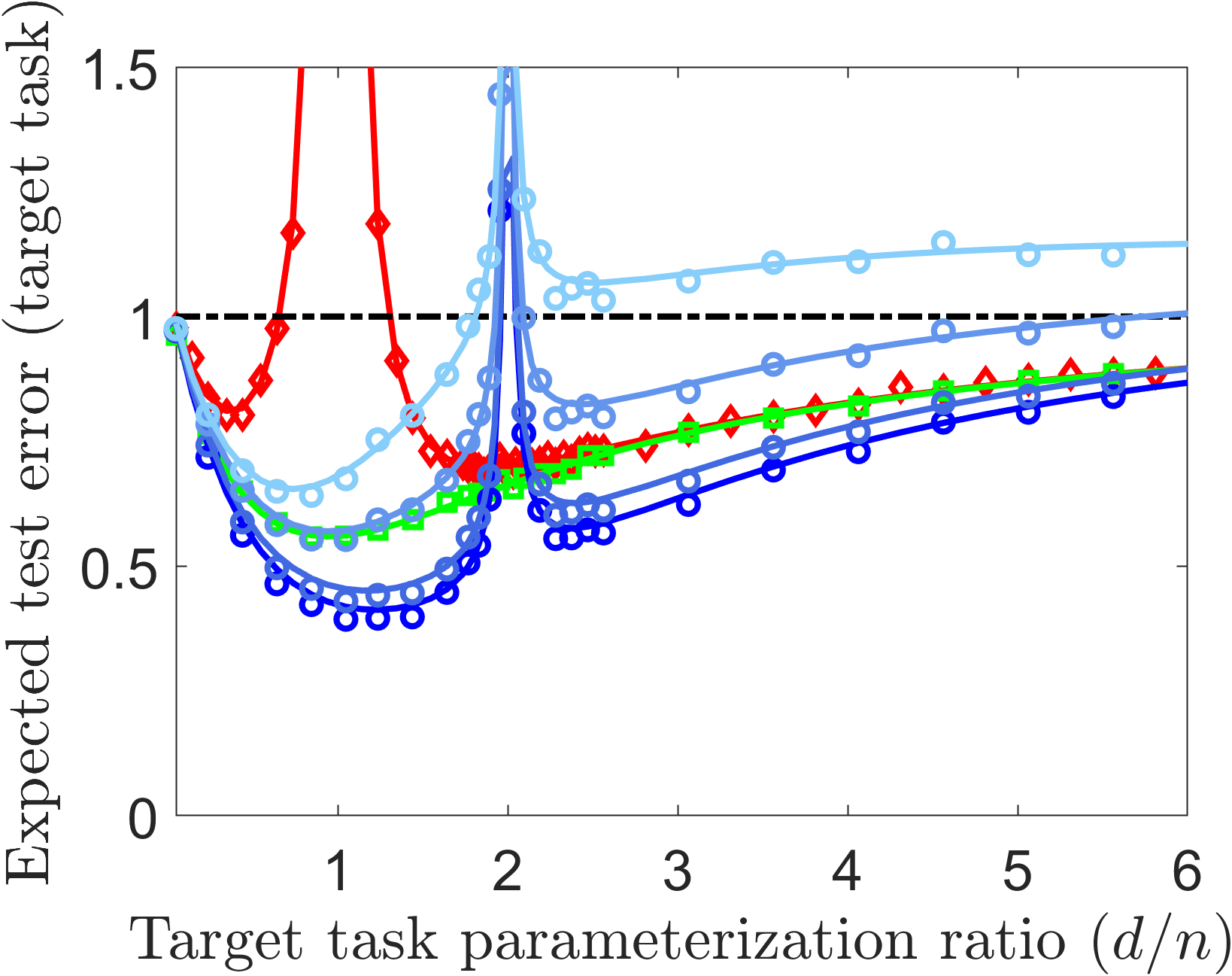}\label{fig:Hgaussian25delta1_HtildeI_error_curves_misspec_Hrho2}}
	\subfloat{\includegraphics[width=0.16\textwidth]{figures/error_curve_legend_without_LMMSE.png}}
	\caption{The test error of the target task under isotropic Gaussian assumption on $\vecgreek{\beta}$ and isotropic target features in a \textbf{misspecified} setting (according to Assumptions \ref{assumption:misspecification}-\ref{assumption:Independent misspecification with isotropic features} and polynomial reduction with $a=2.5$, $q=500$, $\rho=2$). The matrix $\mtx{H}$ is a $d\times d$ circulant matrix corresponding to the discrete version of the continuous-domain convolution kernel ${h_{\rm ker}(\tau)=\delta(\tau)+ e^{-\frac{\lvert\tau-0.5\rvert}{w_{\rm ker}}}}$, here the kernel width is ~${w_{\rm ker}=2/75}$ in (a)-(b) and ${w_{\rm ker}=2/25}$ in (c)-(d). The number of data samples for the target task is $n=64$ and for the source task is $\widetilde{n}=128$.}
	\label{fig:error_curves_diagrams_isotropic_general_H__misspecified}
\end{figure*}

In Figs.~\ref{fig:error_curves_diagrams_isotropic_general_H__well_specified}, \ref{fig:error_curves_diagrams_isotropic_general_H__misspecified} we provide test error evaluations for well-specified and misspecified\footnote{Recall the definition of misspecification in Section \ref{sec:misspecification} and note that one may have a well-specified setting with $\widetilde{\mtx{H}}$ that differs from the true $\mtx{H}$.} settings where $\mtx{H}$ is a circulant matrix that corresponds to circular convolution with the discrete version ($d$ uniformly spaced samples) of the kernel function ${h_{\rm ker}(\tau)=\delta(\tau)+ e^{-\frac{\lvert\tau-0.5\rvert}{w_{\rm ker}}}}$ defined for ${\tau\in[0,1]}$. Here $\delta(\cdot)$ is the Dirac delta.  Note that the discrete convolution kernel is centered to have its peak value at the computed coordinate. Also, the discrete kernel is normalized such that the circulant matrix $\mtx{H}$ satisfies ${\frac{1}{d}\Frobnorm{\mtx{H}}=1}$ for any $d$.  Figs.~\ref{fig:error_curves_diagrams_isotropic_general_H__well_specified}, \ref{fig:error_curves_diagrams_isotropic_general_H__misspecified} show results for different values of the width parameter $w_{\rm ker}$. For a larger $w_{\rm ker}$  the operator $\mtx{H}$ averages a larger neighborhood of coordinates and therefore the source task is less related to the target task, accordingly, Figures \ref{fig:Hgaussian25delta1_HtildeGaussian25delta1_error_curves_wellspec}, \ref{fig:Hgaussian25delta1_HtildeI_error_curves_wellspec}, \ref{fig:Hgaussian25delta1_HtildeGaussian25delta1_error_curves_misspec_Hrho2}, \ref{fig:Hgaussian25delta1_HtildeI_error_curves_misspec_Hrho2} in general show reduced gains from transfer learning compared to Figures \ref{fig:Hgaussian75delta1_HtildeGaussian75delta1_error_curves_wellspec}, \ref{fig:Hgaussian75delta1_HtildeI_error_curves_wellspec}, \ref{fig:Hgaussian75delta1_HtildeGaussian75delta1_error_curves_misspec_Hrho2}, \ref{fig:Hgaussian75delta1_HtildeI_error_curves_misspec_Hrho2}, respectively. 

\begin{figure*}[t]
		\subfloat[Known $\mtx{H}$ (conv.~in DCT domain):\protect\\ $\widetilde{\mtx{H}}=\mtx{H}$]{\includegraphics[width=0.4\textwidth]{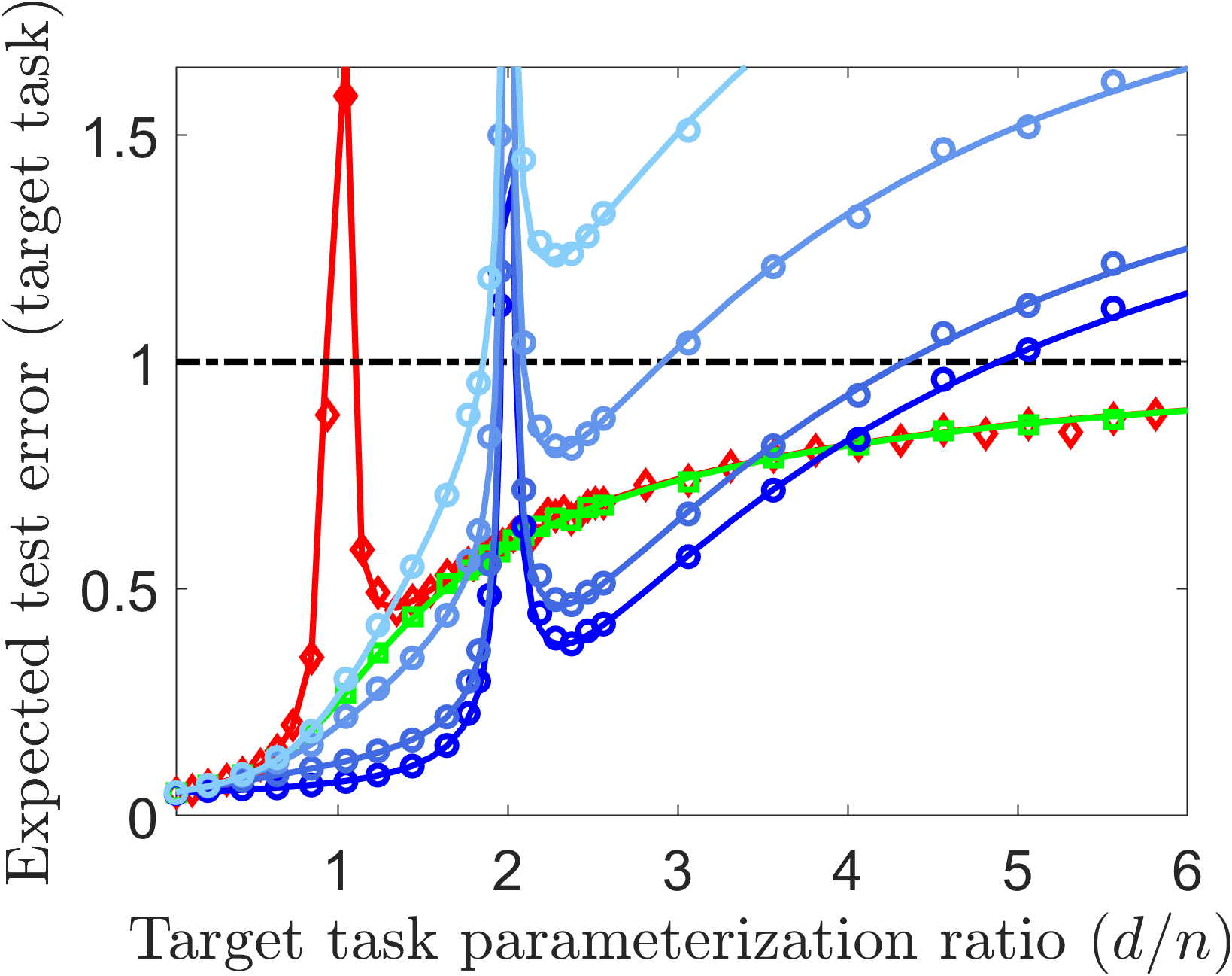}\label{fig:Hgaussian75dct_HtildeGaussian75dct_error_curves_wellspec}}
	\subfloat[Unknown $\mtx{H}$ (conv.~in DCT domain):\protect\\ $\widetilde{\mtx{H}}=\mtx{I}_d$]{\includegraphics[width=0.4\textwidth]{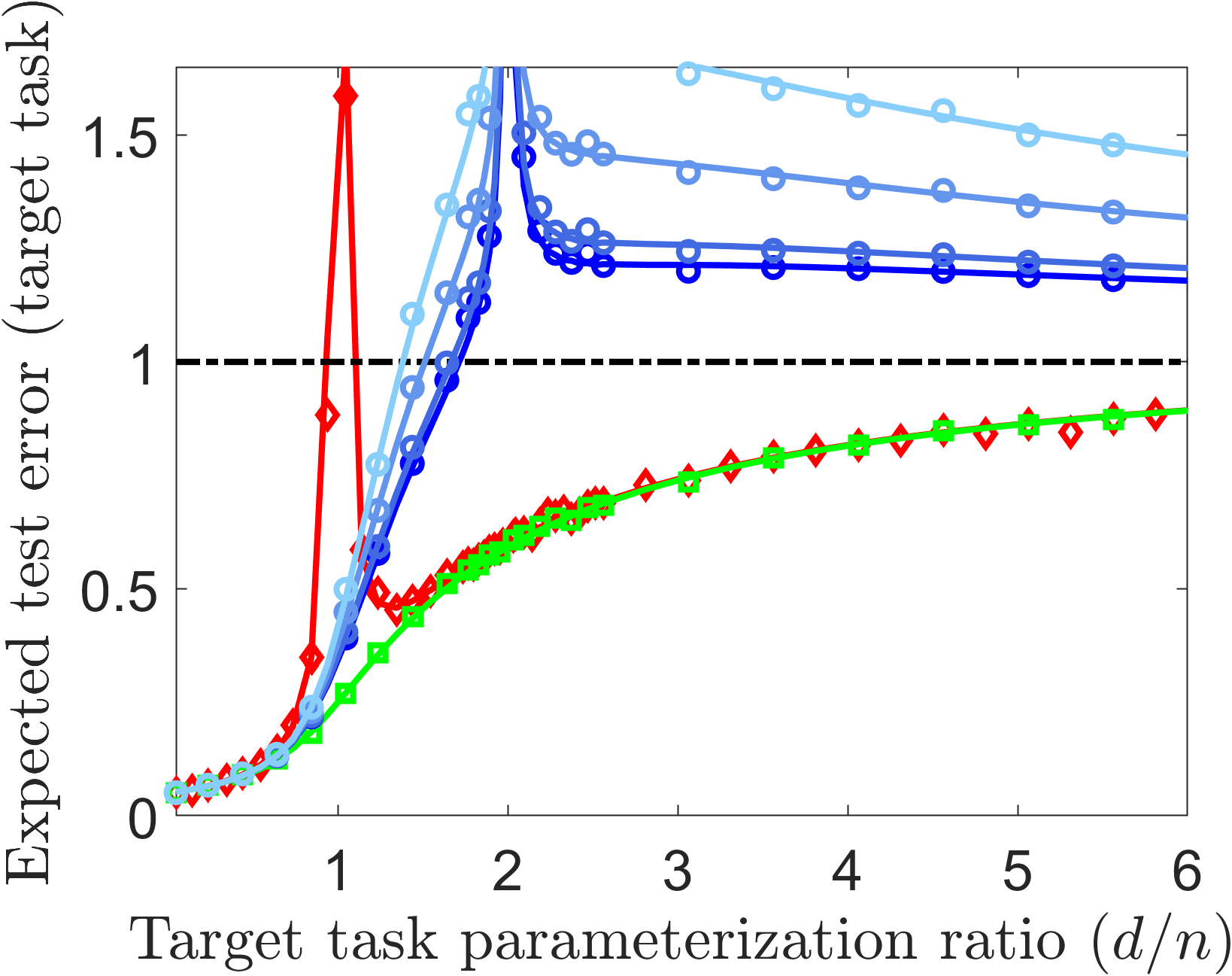}\label{fig:Hgaussian75dct_HtildeI_error_curves_wellspec}}	
 	\subfloat{\includegraphics[width=0.16\textwidth]{figures/error_curve_legend_without_LMMSE.png}}
	\caption{The test error of the target task under isotropic Gaussian assumption on $\vecgreek{\beta}$ and isotropic target features in a \textbf{well specified} setting. The matrix $\mtx{H}$ is a $d\times d$ matrix corresponding to applying the discrete version of the continuous-domain convolution kernel ${h_{\rm ker}(\tau)=\delta(\tau)+ e^{-\frac{\lvert\tau-0.5\rvert}{w_{\rm ker}}}}$ in the \textbf{DCT domain}, here the kernel width is ~${w_{\rm ker}=2/75}$. The number of data samples for the target task is $n=64$ and for the source task is $\widetilde{n}=128$.}
	\label{fig:error_curves_diagrams_isotropic_general_H__wellspec_dct_domain}
\end{figure*}

The results in Fig.~\ref{fig:error_curves_diagrams_isotropic_general_H__wellspec_dct_domain} refer to a setting where $\mtx{H}$ is a convolution with ${h_{\rm ker}}$ but in the DCT domain; namely, $\mtx{H}$ is a composition of the $d\times d$ DCT matrix followed by the $d\times d$ circulant matrix that corresponds to ${h_{\rm ker}}$. 

Figures \ref{fig:error_curves_diagrams_isotropic_general_H__well_specified}, \ref{fig:error_curves_diagrams_isotropic_general_H__misspecified}, \ref{fig:error_curves_diagrams_isotropic_general_H__wellspec_dct_domain} exemplify interesting behaviors that will be discussed in the next sections.

\begin{figure*}[t]
 	\subfloat[${w_{\rm ker}=2/75}$]{\includegraphics[width=0.32\textwidth]{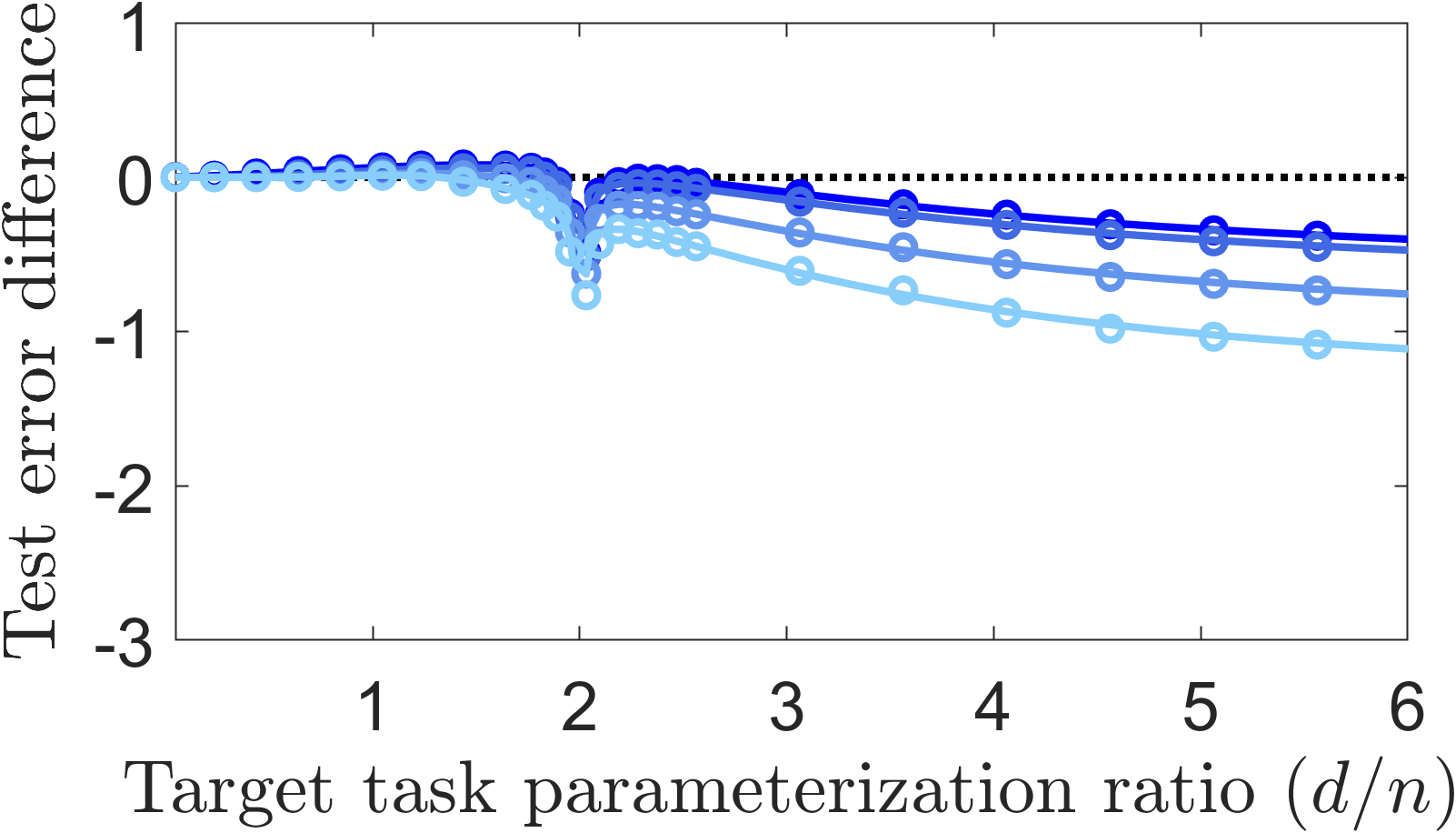}\label{fig:Hgaussian75delta1_error_difference_curves_wellspec}}
	\subfloat[${w_{\rm ker}=2/25}$]{\includegraphics[width=0.32\textwidth]{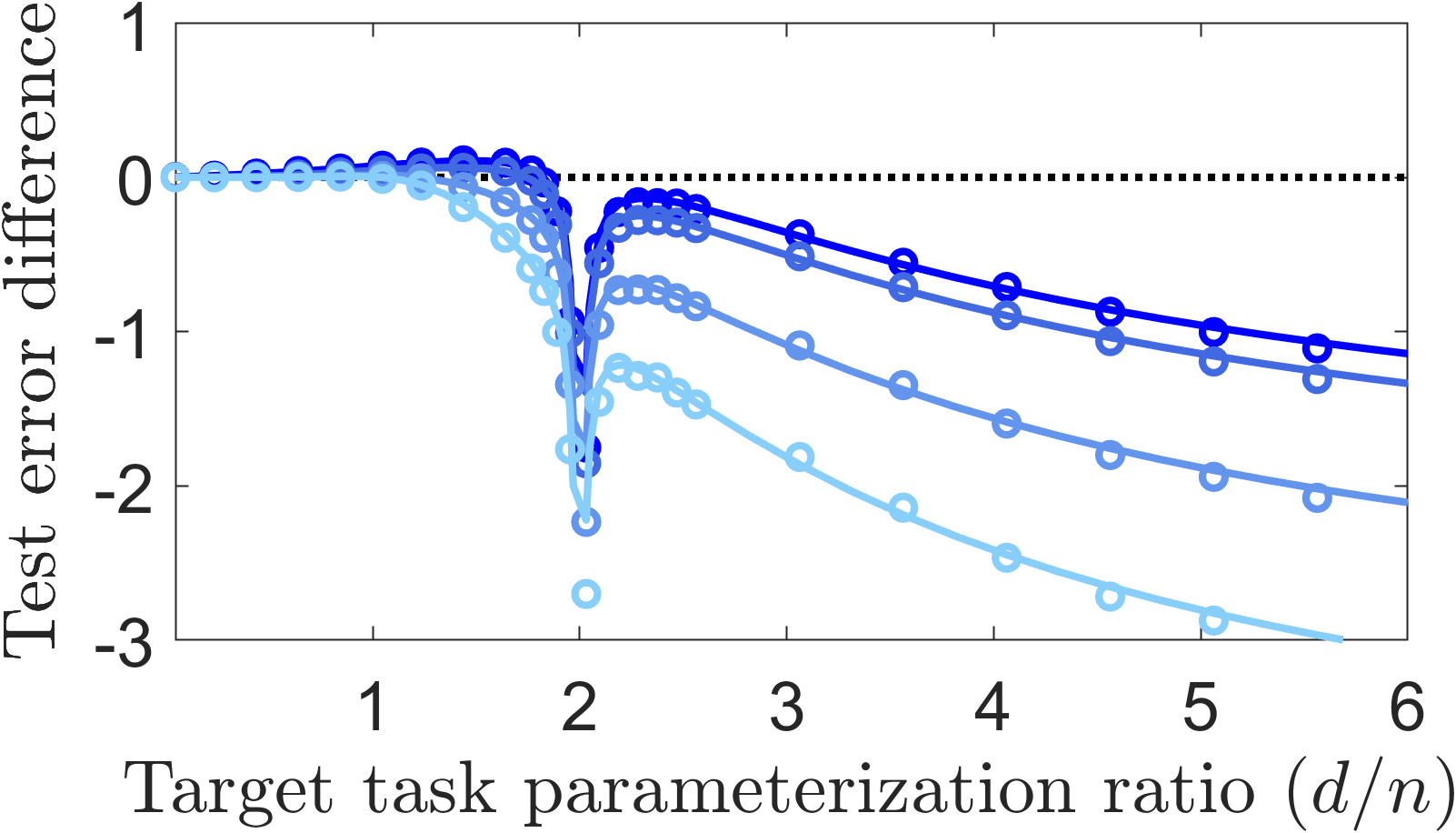}\label{fig:Hgaussian25delta1_error_difference_curves_wellspec}}
 		\subfloat[${w_{\rm ker}=2/75}$ in DCT domain]{\includegraphics[width=0.32\textwidth]{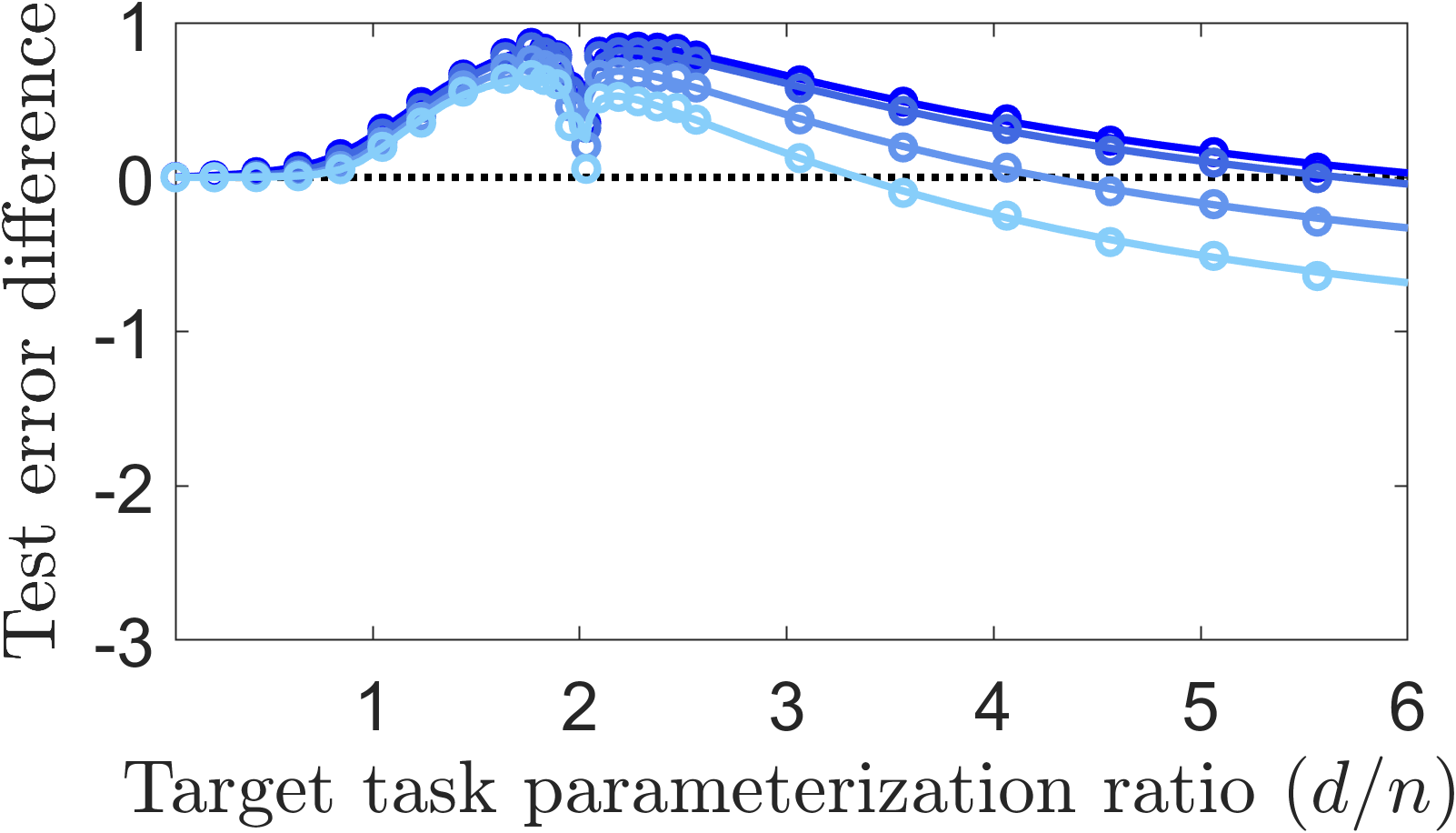}\label{fig:Hgaussian75dct_error_difference_curves_wellspec}}
  \\
  \subfloat[${w_{\rm ker}=2/75}$]{\includegraphics[width=0.32\textwidth]{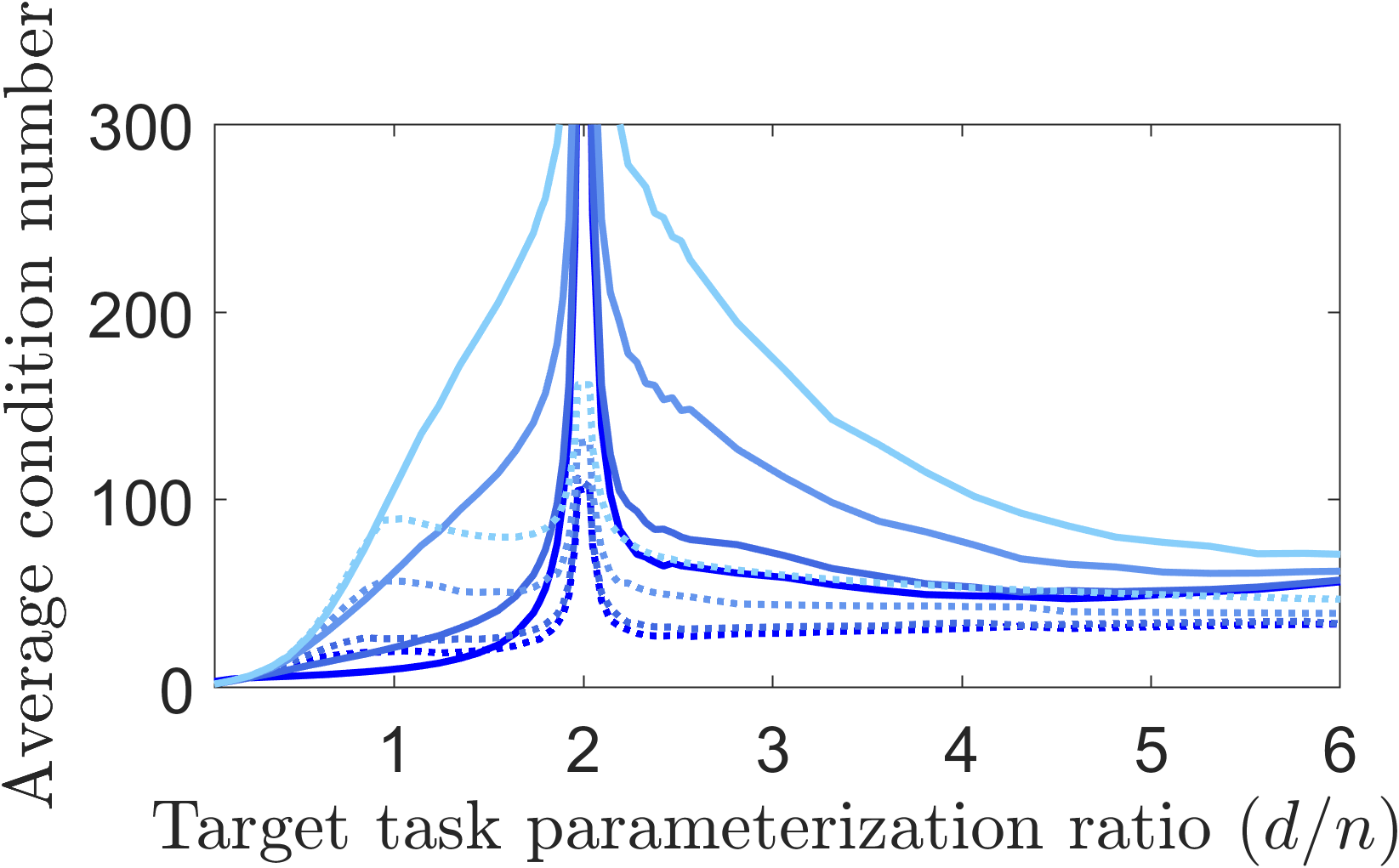}\label{fig:Hgaussian75delta1_condition_number_curves_wellspec}}
	\subfloat[${w_{\rm ker}=2/25}$]{\includegraphics[width=0.32\textwidth]{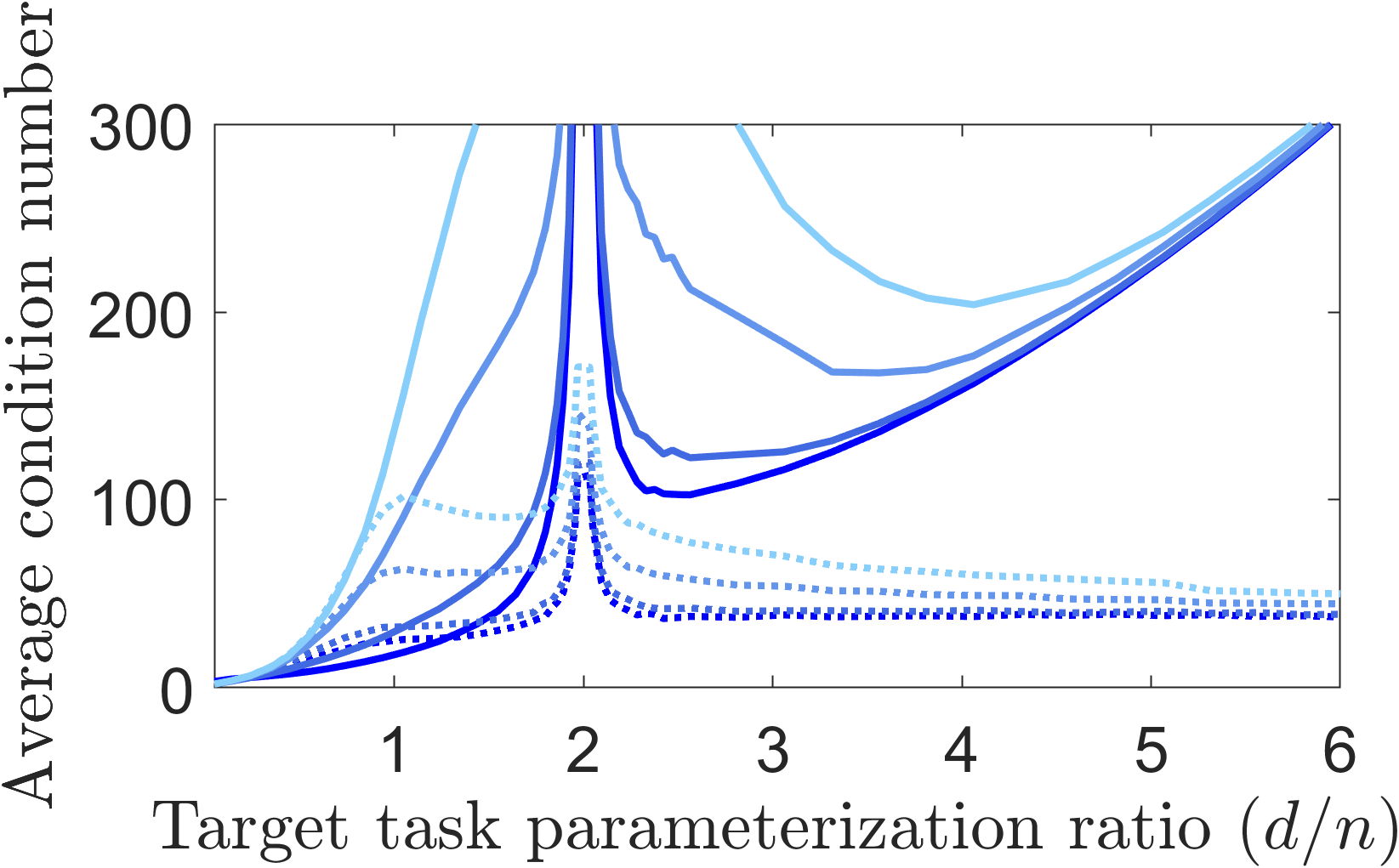}\label{fig:Hgaussian25delta1_condition_number_curves_wellspec}}	 \subfloat[${w_{\rm ker}=2/75}$ in DCT domain]{\includegraphics[width=0.32\textwidth]{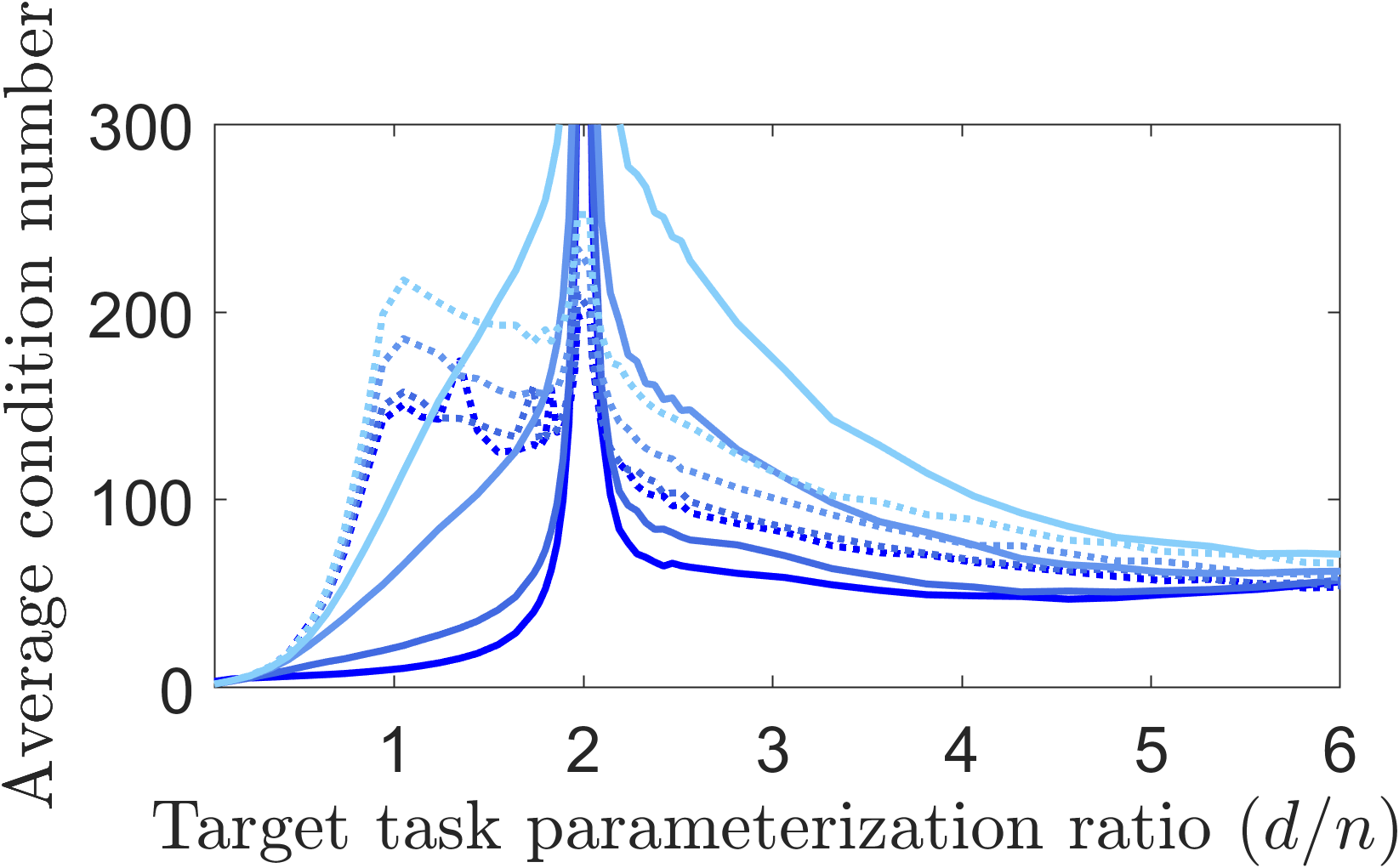}\label{fig:Hgaussian75dct_condition_number_curves_wellspec}}	
	\caption{The test error difference between using $\widetilde{\mtx{H}} = \mtx{I}_d$ and $\widetilde{\mtx{H}} = \mtx{H}$ (subfigures (a)-(c), black dotted lines corresponds to zero difference). Each curve color corresponds to another noise level ($\sigma_{\eta}^2$) in the task relation, the colors refer to the same noise levels as in Fig.~\ref{fig:error_curves_diagrams_isotropic_general_H__well_specified}. The corresponding condition number of the matrix to invert (\ref{eq:matrix to invert in intuitive transfer learning}) are presented in subfigures (d)-(f); solid lines refer to $\widetilde{\mtx{H}} = \mtx{H}$ and dotted lines refer to $\widetilde{\mtx{H}} = \mtx{I}_d$. Note that the peak of the condition number around the interpolation threshold of the source task are due to the optimal transfer learning parameter that is involved in the matrix in (\ref{eq:matrix to invert in intuitive transfer learning}).
 These results are for isotropic Gaussian assumption on $\vecgreek{\beta}$ and isotropic target features in a well specified setting. Each column of subfigures corresponds to another matrix $\mtx{H}$ that is based on applying the discrete version of the continuous-domain convolution kernel ${h_{\rm ker}(\tau)=\delta(\tau)+ e^{-\frac{\lvert\tau-0.5\rvert}{w_{\rm ker}}}}$ for ${w_{\rm ker}=2/75}$ (left column of subfigures), ${w_{\rm ker}=2/25}$ (middle column of subfigures), and ${w_{\rm ker}=2/75}$ in the DCT domain (right column of subfigures). }
	\label{fig:HtildeH_vs_HtildeI_error_difference_and_condition_number_curves}
\end{figure*}

\subsection{The Intuitive Transfer Learning can be Improved by Ignoring the True Task Relation}
\label{subsec:improvements due to Htilde=I}

Let us examine the intuitive transfer learning method for $\widetilde{\mtx{H}}=\mtx{I}_d$ due to an unknown $\mtx{H}$, and compare it to a setting where $\mtx{H}$ is known and $\widetilde{\mtx{H}}=\mtx{H}$. 

Remarkably, our results showcase that, in the overparameterized regime, the intuitive transfer learning with $\widetilde{\mtx{H}}=\mtx{I}_d$ can significantly outperform the intuitive transfer learning with a known $\mtx{H}$ and $\widetilde{\mtx{H}}=\mtx{H}$. This behavior can be observed in several settings where the true $\mtx{H}$ differs from $\mtx{I}_d$; for example, compare the error curves of intuitive transfer learning (in the overparameterized regime) with a known $\mtx{H}$ in Figs.~\ref{fig:Hgaussian75delta1_HtildeGaussian75delta1_error_curves_wellspec}, \ref{fig:Hgaussian25delta1_HtildeGaussian25delta1_error_curves_wellspec}, \ref{fig:Hgaussian75delta1_HtildeGaussian75delta1_error_curves_misspec_Hrho2}, \ref{fig:Hgaussian25delta1_HtildeGaussian25delta1_error_curves_misspec_Hrho2} to their corresponding curves with an unknown $\mtx{H}$ in Figs.~\ref{fig:Hgaussian75delta1_HtildeI_error_curves_wellspec}, \ref{fig:Hgaussian25delta1_HtildeI_error_curves_wellspec}, \ref{fig:Hgaussian75delta1_HtildeI_error_curves_misspec_Hrho2}, \ref{fig:Hgaussian25delta1_HtildeI_error_curves_misspec_Hrho2}, respectively.  Figures~\ref{fig:Hgaussian75delta1_error_difference_curves_wellspec}-\ref{fig:Hgaussian75dct_error_difference_curves_wellspec} show the error differences of the corresponding error curves, such that a negative error difference implies that using $\widetilde{\mtx{H}}=\mtx{I}_d$ has a lower test error than using $\widetilde{\mtx{H}}=\mtx{H}$. 

To explain the potential improvement despite ignoring the true $\mtx{H}$, recall that the closed-form solution of the intuitive transfer learning in (\ref{eq:well specified - constrained linear regression - solution - target task - positive alpha}) includes an inversion of the matrix 
\begin{equation}
\label{eq:matrix to invert in intuitive transfer learning}
\mtx{X}^{T} \mtx{X} + n{\alpha_{\rm TL}}\widetilde{\mtx{H}}^{T} \widetilde{\mtx{H}}. 
\end{equation}
We consider a full rank $\widetilde{\mtx{H}}$ (recall Assumption \ref{assumption: H is full rank}) and, therefore, the matrix in (\ref{eq:matrix to invert in intuitive transfer learning}) is invertible. Yet, there is still a question of the ability of the solution to attenuate noise effectively.

The condition number of (\ref{eq:matrix to invert in intuitive transfer learning}), namely, the ratio between the maximal and minimal eigenvalues of the matrix in (\ref{eq:matrix to invert in intuitive transfer learning}), provides a useful way to characterize a linear system's susceptibility to noise. Note that in the overparameterized regime, $\mtx{X}$ is a $n\times d$ feature matrix where $d>n$ and, thus, the $d\times d$ matrix $\mtx{X}^{T} \mtx{X}$ is rank deficient with at least $d-n$ zero eigenvalues. 
Moreover, the singular values of the full-rank matrix $\widetilde{\mtx{H}}$ are all non-zeros, but they can still be very small. A small singular value of $\widetilde{\mtx{H}}$ yields a small eigenvalue of $\widetilde{\mtx{H}}^{T} \widetilde{\mtx{H}}$. Hence, for $\widetilde{\mtx{H}}^{T} \widetilde{\mtx{H}}$ with a considerable number of small eigenvalues, in highly overparameterized settings (where $\mtx{X}^{T} \mtx{X}$ is significantly rank deficient), the matrix in (\ref{eq:matrix to invert in intuitive transfer learning}) is likely to have many small eigenvalues and a large condition number. To see why this is a problem, if we had $\mtx{X} = \mtx{0}$ (which is morally true on the null space of $\mtx{X}$), then $\widehat{\vecgreek{\beta}}_{\mathrm{TL}} = \widetilde{\mtx{H}}^+ \widehat{\vecgreek{\theta}}$, meaning that any error in $\widehat{\vecgreek{\theta}}$ is amplified by small singular values of $\widetilde{\mtx{H}}$.

Setting $\widetilde{\mtx{H}}=\mtx{I}_d$ eliminates the error amplifying effects of small singuar values of $\widetilde{\mtx{H}}$, which can be even more beneficial than using the true $\mtx{H}$ in the intuitive transfer learning. Put differently, if setting $\widetilde{\mtx{H}}=\mtx{H}$ indeed amplifies error, there is a tradeoff between the errors induced by ignoring the true task relation and by the error amplification due to using the true task relation. For example, (\ref{eq:well specified - out of sample error - target task -  asymptotic - anisotropic target features - general H})-(\ref{eq:well specified - out of sample error - target task -  asymptotic - anisotropic target features - general H - definition of Gamma_TLinf}) show that using $\widetilde{\mtx{H}}$ with a lower condition number can reduce error terms that involve $\widetilde{\mtx{H}}^{-1}$ at the expense of some increase in the error terms that depend on the difference of $\widetilde{\mtx{H}}$  from $\mtx{H}$. 
In many cases, $\widetilde{\mtx{H}}=\mtx{I}_d$ addresses this tradeoff well and importantly shows that \textbf{one does not have to use nor know $\mtx{H}$!}

Figures \ref{fig:error_curves_diagrams_isotropic_general_H__well_specified}, \ref{fig:error_curves_diagrams_isotropic_general_H__misspecified}, \ref{fig:error_curves_diagrams_isotropic_general_H__wellspec_dct_domain} and the error difference curves in Fig.~\ref{fig:HtildeH_vs_HtildeI_error_difference_and_condition_number_curves} demonstrate when using $\widetilde{\mtx{H}}=\mtx{I}_d$ can outperform $\widetilde{\mtx{H}}=\mtx{H}$, and when it cannot. 
Whereas Figs.~\ref{fig:Hgaussian75delta1_error_difference_curves_wellspec}, \ref{fig:Hgaussian25delta1_error_difference_curves_wellspec} clearly show the significant benefits of using $\widetilde{\mtx{H}}=\mtx{I}_d$ over using the true $\mtx{H}$, Fig.~\ref{fig:Hgaussian75dct_error_difference_curves_wellspec} does not show such benefits except for the ultra high overparameterization levels. 
In Fig.~\ref{fig:Hgaussian75dct_error_difference_curves_wellspec} using $\widetilde{\mtx{H}}=\mtx{I}_d$ performs worse than $\widetilde{\mtx{H}}=\mtx{H}$ because the gap between the condition numbers of the two cases is moderate (Fig.~\ref{fig:Hgaussian75dct_condition_number_curves_wellspec}) and $\mtx{H}$  is too far from $\mtx{I}_d$ (due to the transformation to the DCT domain before the convolution). 
In contrast, the significant benefits of using $\widetilde{\mtx{H}}=\mtx{I}_d$ in Fig.~\ref{fig:Hgaussian25delta1_error_difference_curves_wellspec} are reflected also by its ability to resolve the vast increase in the condition number of (\ref{eq:matrix to invert in intuitive transfer learning}) for $\widetilde{\mtx{H}}=\mtx{H}$ in the highly overparameterized regime. 
Also, Figs.~\ref{fig:Hgaussian75delta1_error_difference_curves_wellspec}, \ref{fig:Hgaussian25delta1_error_difference_curves_wellspec} show greater gains from using $\widetilde{\mtx{H}}=\mtx{I}_d$ when $\mtx{H}$ is farther from $\mtx{I}_d$ (i.e., note the greater gains for $\mtx{H}$ with ${w_{\rm ker}=2/25}$ compared to ${w_{\rm ker}=2/75}$, although the latter is closer to $\mtx{I}_d$).

More generally, we observed that using the true $\mtx{H}$ in our intuitive transfer learning approach can be outperformed by other linear models that do not use the true $\mtx{H}$. This raises the question of what is the optimal linear model for transfer learning with a known $\mtx{H}$ --- we will answer this question in the next section.

\begin{figure*}[h]
		\subfloat[Well specified, ${w_{\rm ker}=2/75}$]{\includegraphics[width=0.46\textwidth]{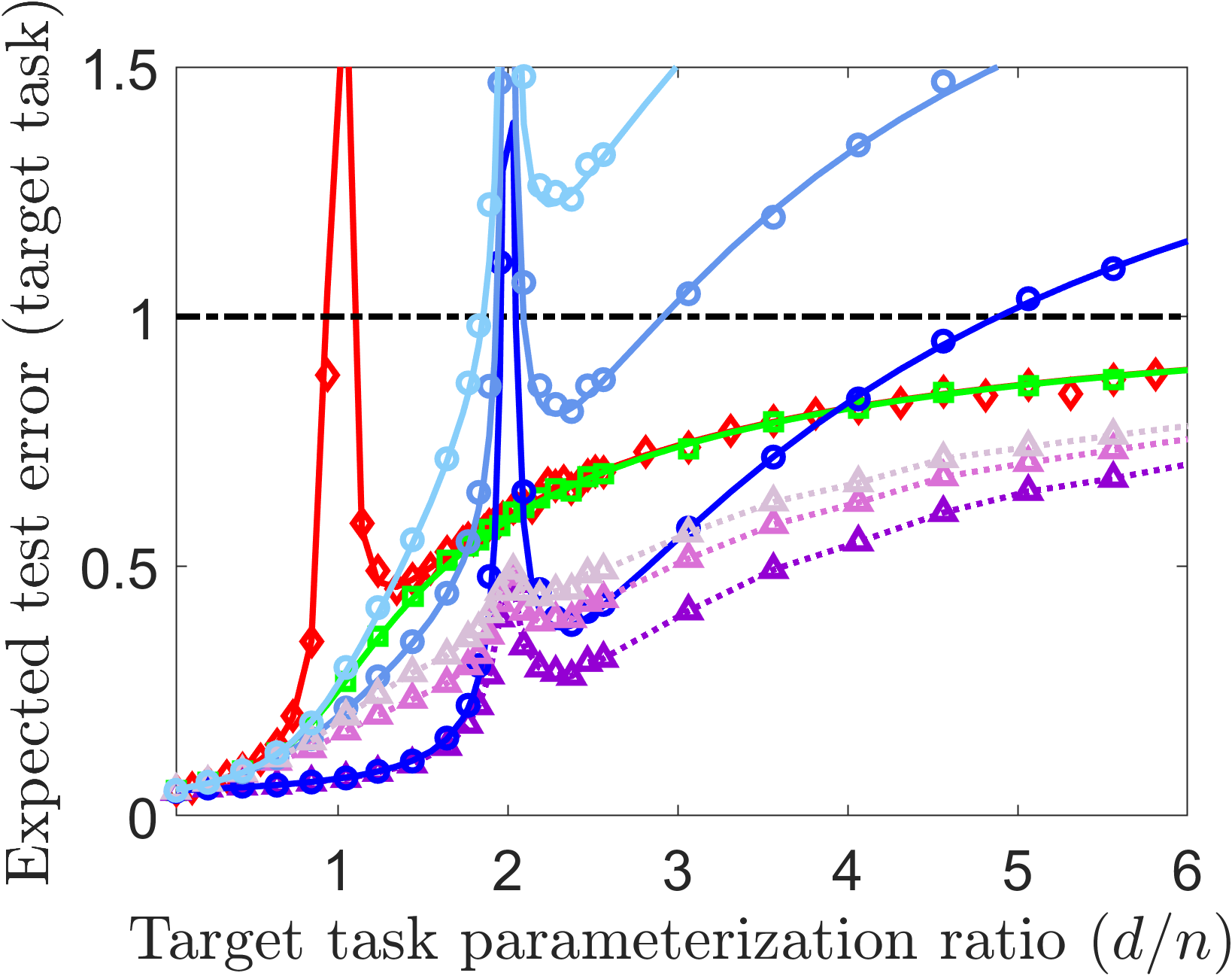}\label{fig:Hgaussian75delta1_HtildeGaussian75delta1_error_curves_wellspec_woEta0p1}}
	~~\subfloat[Misspecified, ${w_{\rm ker}=2/75}$]{\includegraphics[width=0.46\textwidth]{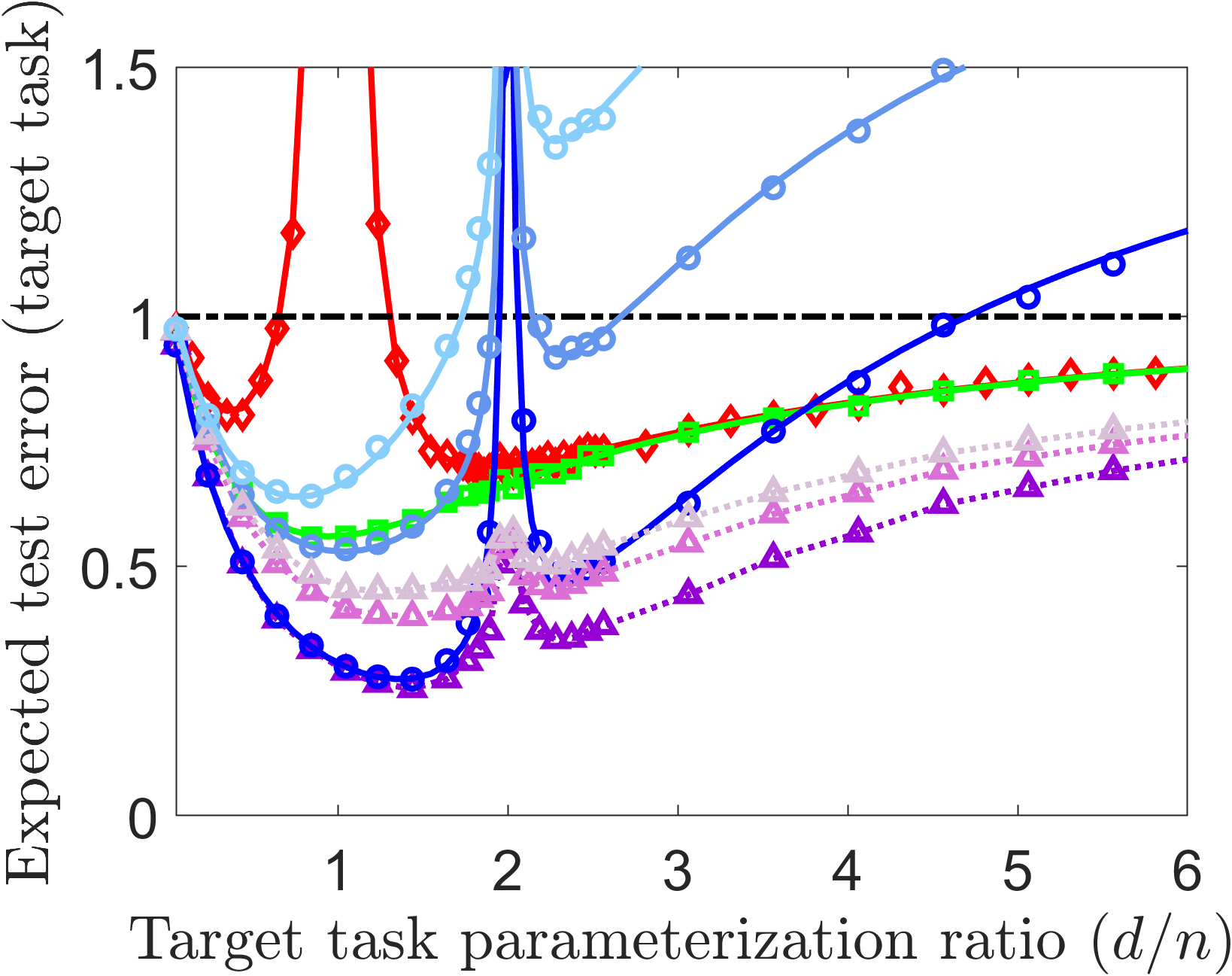}\label{fig:Hgaussian75delta1_HtildeGaussian75delta1_error_curves_misspec_Hrho2_woEta0p1}}	
 \\
		\subfloat[Well specified, ${w_{\rm ker}=2/25}$]{\includegraphics[width=0.46\textwidth]{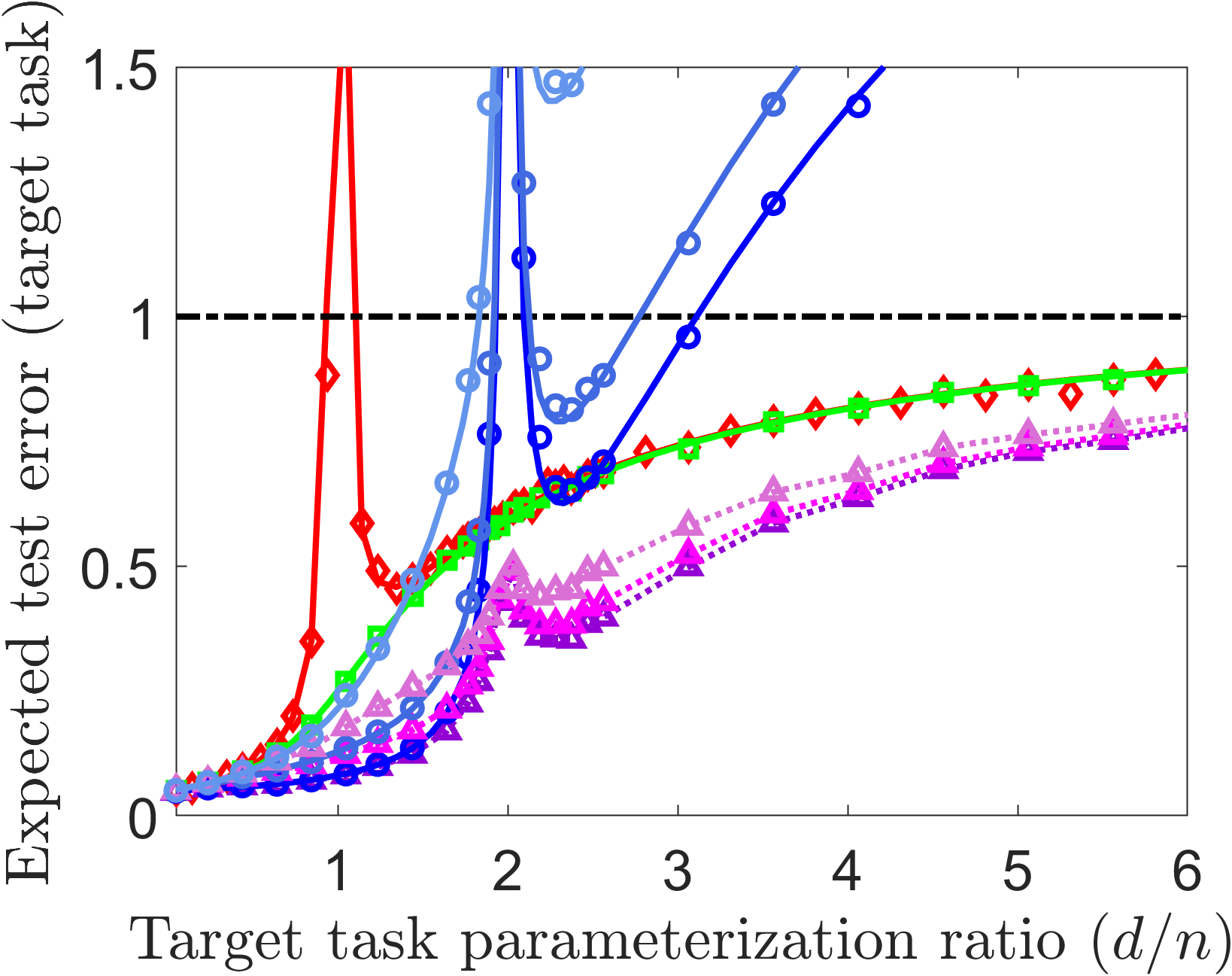}\label{fig:Hgaussian25delta1_HtildeGaussian25delta1_error_curves_wellspec_woEta1}}
	~~\subfloat[Misspecified, ${w_{\rm ker}=2/25}$]{\includegraphics[width=0.46\textwidth]{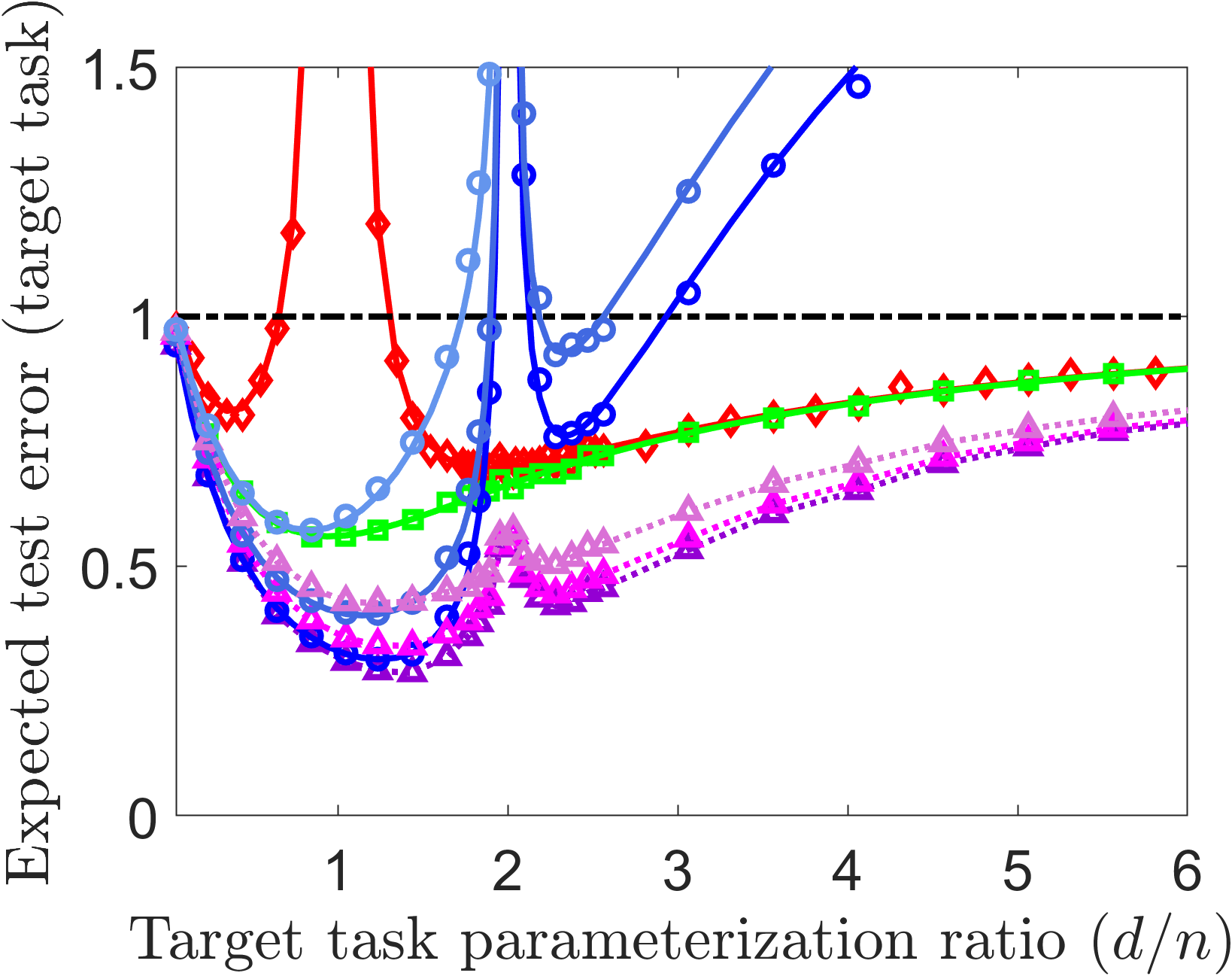}\label{fig:Hgaussian25delta1_HtildeGaussian25delta1_error_curves_misspec_Hrho2_woEta1}}	
 \\
 		\subfloat[Well specified, ${w_{\rm ker}=2/75}$ (in DCT domain)]{\includegraphics[width=0.46\textwidth]{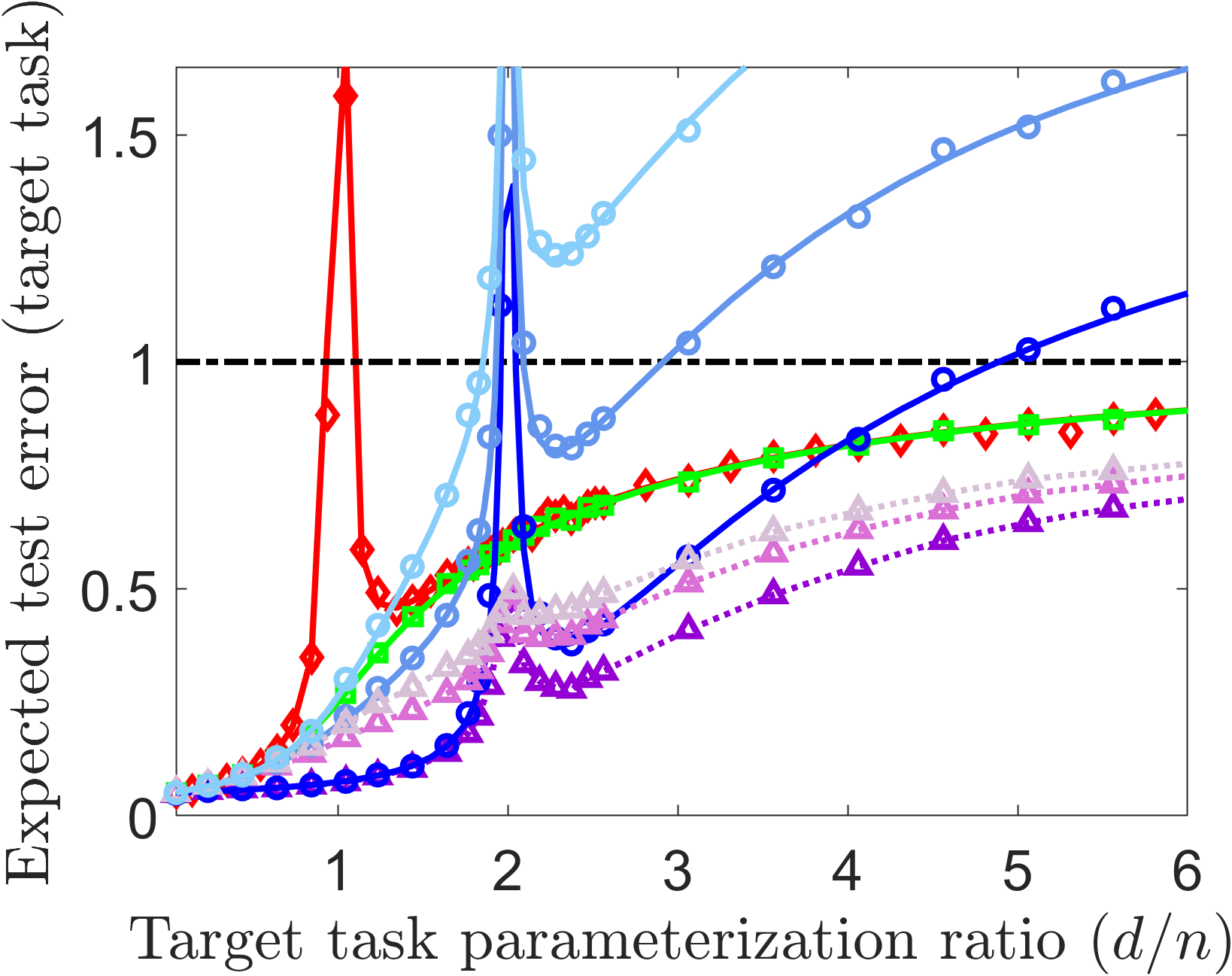}\label{fig:Hgaussian75dct_HtildeGaussian75dct_error_curves_wellspec_woEta0p1}}
	~~~~~~~~~~~~~~~~~~~\subfloat{\includegraphics[width=0.2\textwidth]{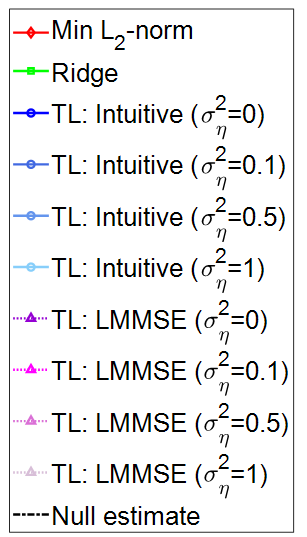}}
	\caption{The test error of the target task. Comparison of the intuitive transfer learning method to the LMMSE transfer learning method. Subfigures (a), (b), (c), (d), (e) extend Subfigures \ref{fig:Hgaussian75delta1_HtildeGaussian75delta1_error_curves_wellspec}, \ref{fig:Hgaussian75delta1_HtildeGaussian75delta1_error_curves_misspec_Hrho2}, \ref{fig:Hgaussian25delta1_HtildeGaussian25delta1_error_curves_wellspec}, \ref{fig:Hgaussian25delta1_HtildeGaussian25delta1_error_curves_misspec_Hrho2}, \ref{fig:Hgaussian75dct_HtildeGaussian75dct_error_curves_wellspec}, respectively, by adding the LMMSE error curves. Here, in each of the subfigures, one of the noise level $\sigma_{\eta}^2$ curves is absent for better visibility. 
 All the results in this figure are for $\widetilde{\mtx{H}}=\mtx{H}$.}
	\label{fig:error_curves_diagrams_isotropic_general_H__with_LMMSE_TL}
\end{figure*}

\begin{figure*}[h]
		\subfloat[Well specified, ${w_{\rm ker}=2/75}$]{\includegraphics[width=0.46\textwidth]{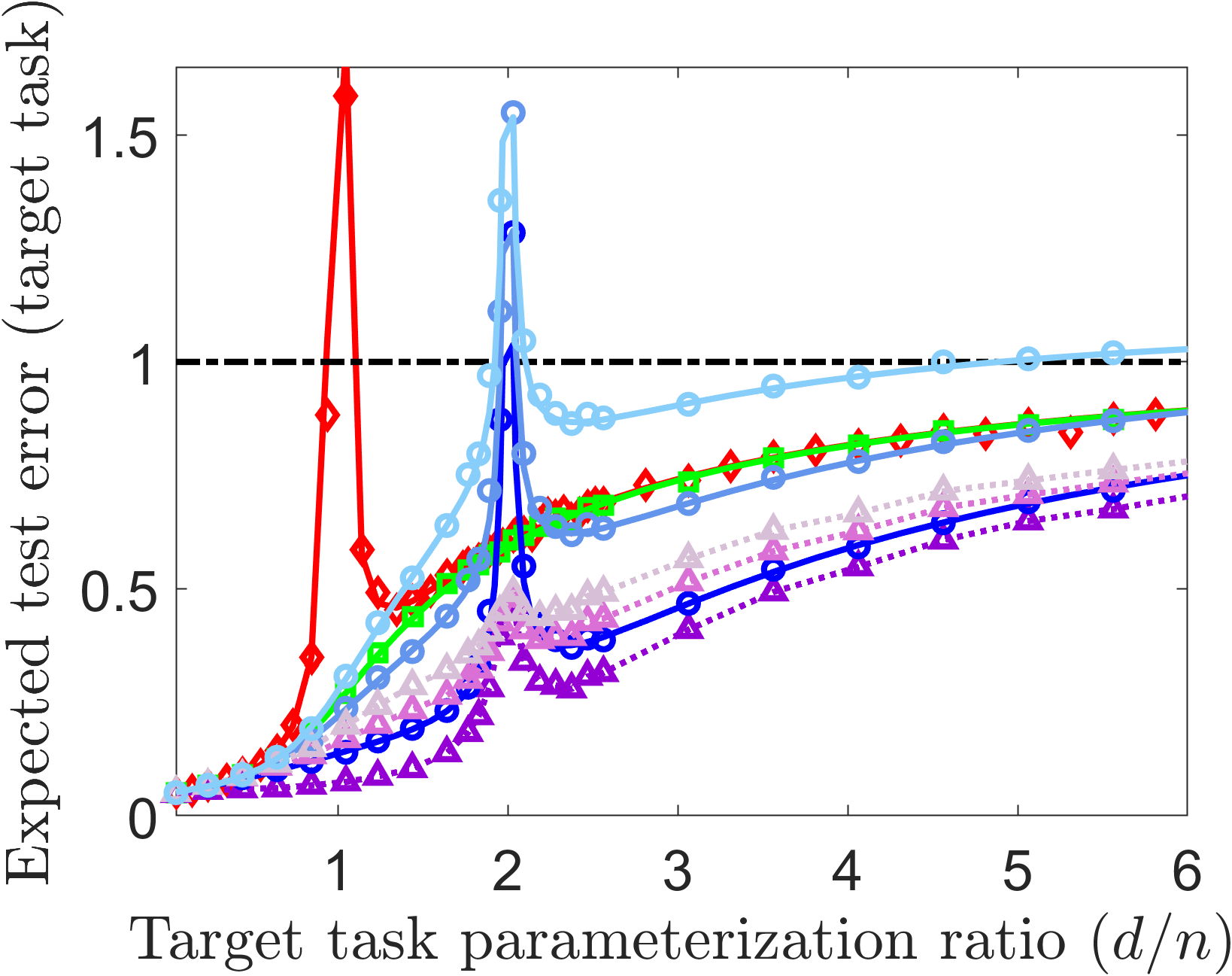}\label{fig:Hgaussian75delta1_HtildeI_error_curves_wellspec_woEta0p1}}
	~~\subfloat[Misspecified, ${w_{\rm ker}=2/75}$]{\includegraphics[width=0.46\textwidth]{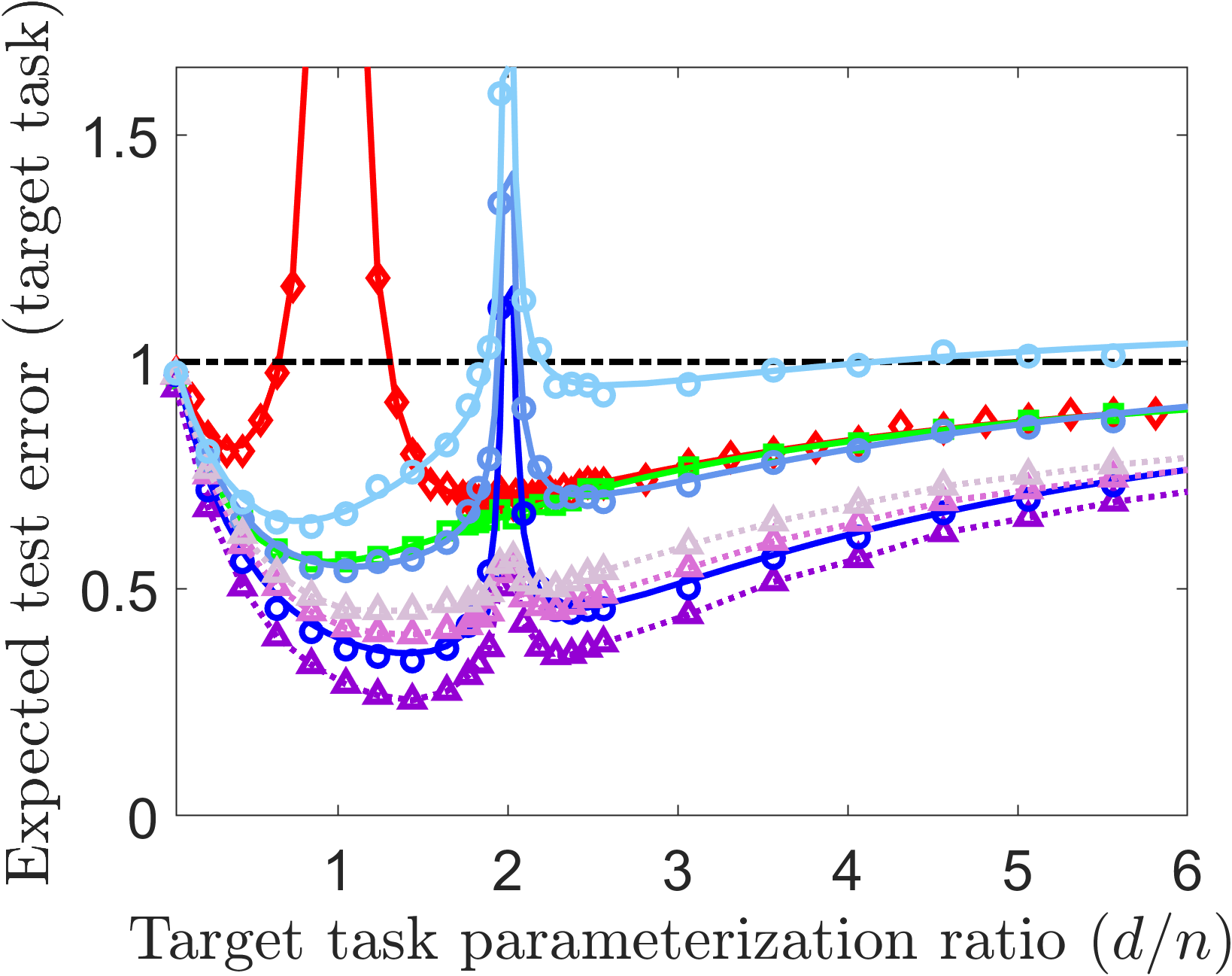}\label{fig:Hgaussian75delta1_HtildeI_error_curves_misspec_Hrho2_woEta0p1}}	
 \\
		\subfloat[Well specified, ${w_{\rm ker}=2/25}$]{\includegraphics[width=0.46\textwidth]{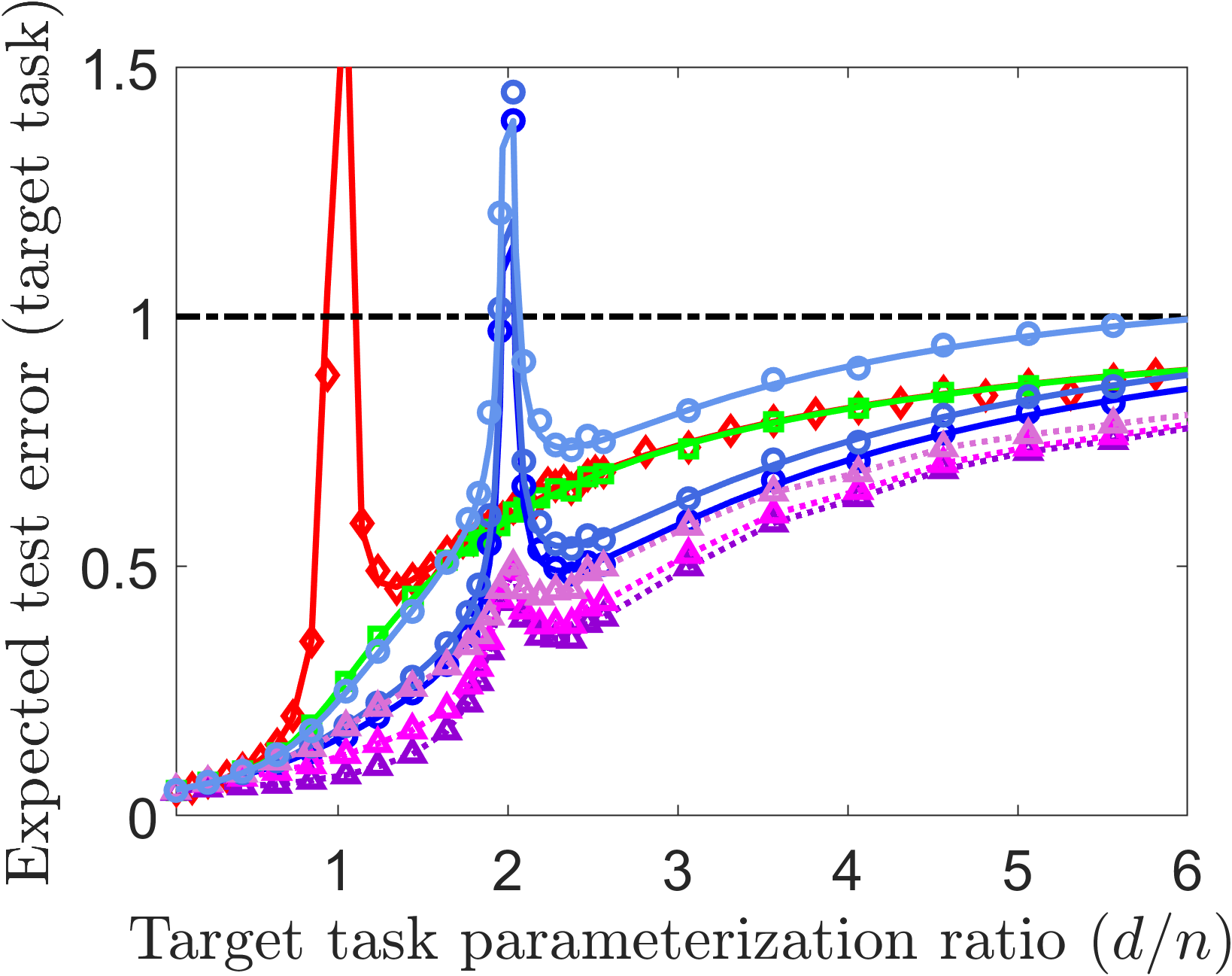}\label{fig:Hgaussian25delta1_HtildeI_error_curves_wellspec_woEta1}}
	~~\subfloat[Misspecified, ${w_{\rm ker}=2/25}$]{\includegraphics[width=0.46\textwidth]{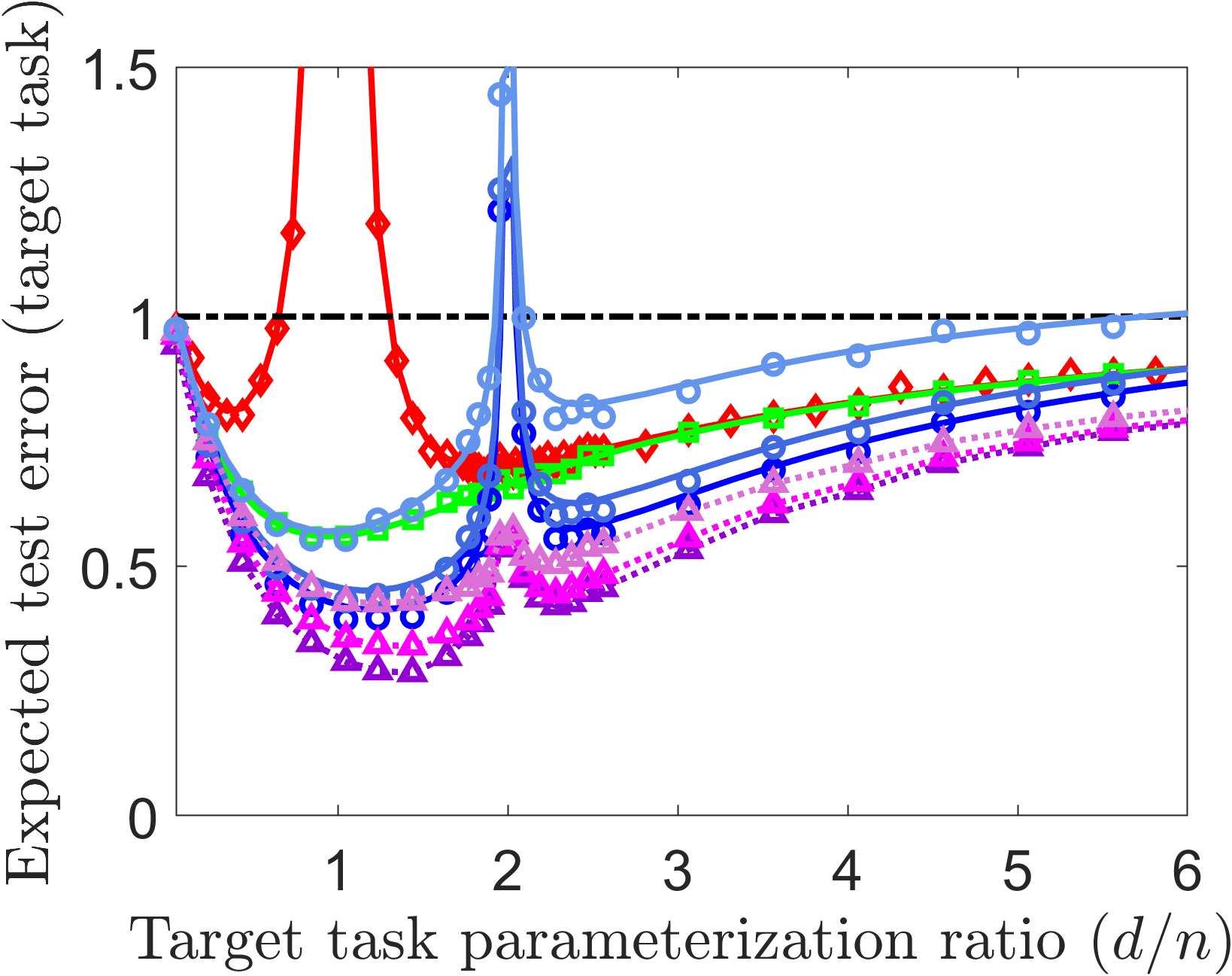}\label{fig:Hgaussian25delta1_HtildeI_error_curves_misspec_Hrho2_woEta1}}	
 \\
 		\subfloat[Well specified, ${w_{\rm ker}=2/75}$ (in DCT domain)]{\includegraphics[width=0.46\textwidth]{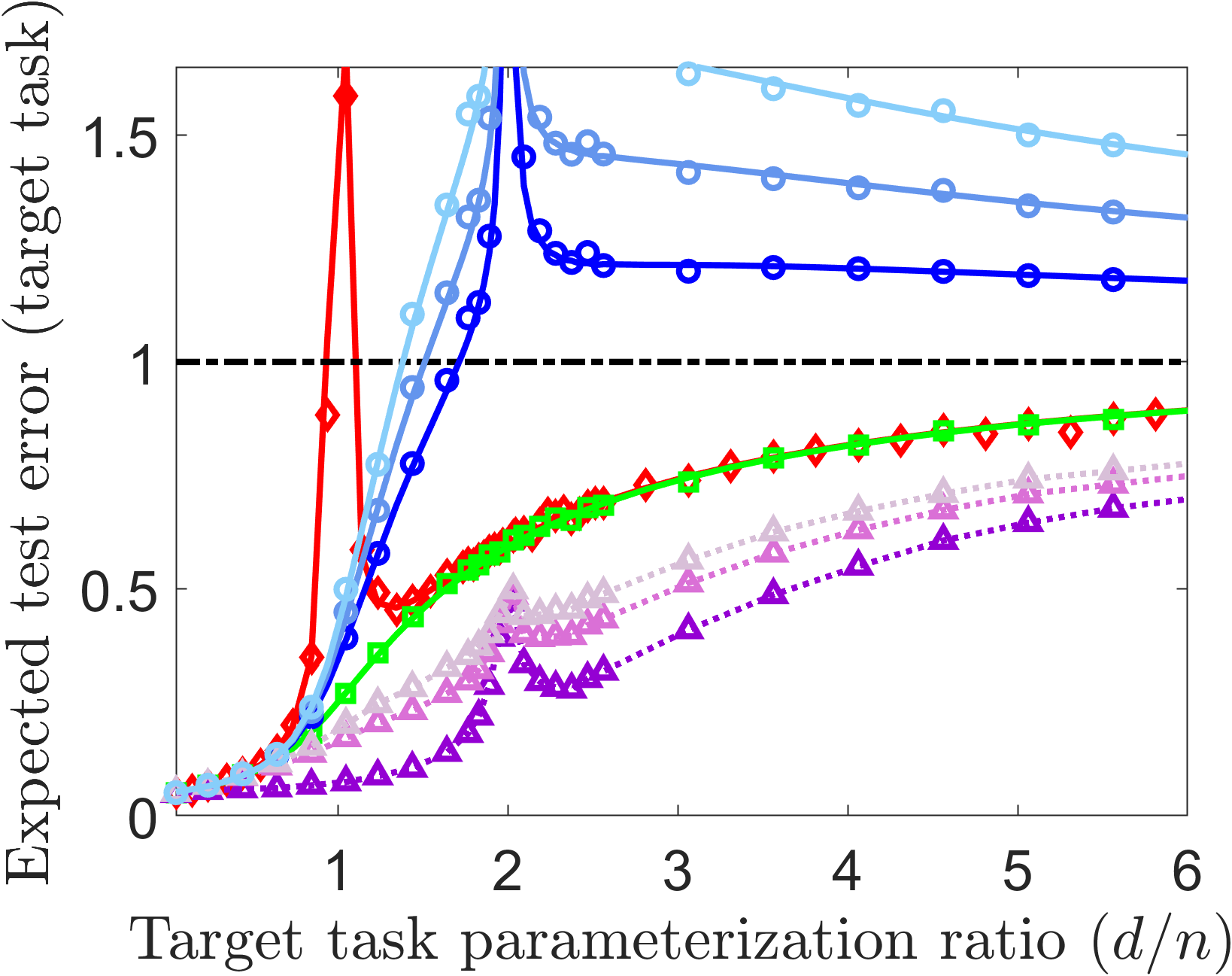}\label{fig:Hgaussian75dct_HtildeI_error_curves_wellspec_woEta0p1}}
		~~~~~~~~~~~~~~~~~~~\subfloat{\includegraphics[width=0.2\textwidth]{figures/error_curve_legend.png}}
	\caption{The test error of the target task. Comparison of the intuitive transfer learning method to the LMMSE transfer learning method. Subfigures (a), (b), (c), (d), (e) extend Subfigures \ref{fig:Hgaussian75delta1_HtildeI_error_curves_wellspec}, \ref{fig:Hgaussian75delta1_HtildeI_error_curves_misspec_Hrho2}, \ref{fig:Hgaussian25delta1_HtildeI_error_curves_wellspec}, \ref{fig:Hgaussian25delta1_HtildeI_error_curves_misspec_Hrho2}, \ref{fig:Hgaussian75dct_HtildeI_error_curves_wellspec}, respectively, by adding the LMMSE error curves. Here, in each of the subfigures, one of the noise level $\sigma_{\eta}^2$ curves is absent for better visibility. 
 All the results in this figure are for the intuitive TL with $\widetilde{\mtx{H}}=\mtx{I}_d$, but recall that the LMMSE TL always uses $\mtx{H}$.}
	\label{fig:error_curves_diagrams_isotropic_general_H__with_LMMSE_TL__HtildeI}
\end{figure*}

\subsection{The Linear MMSE Solution to Transfer Learning}
\label{subsec:The Linear MMSE Solution to Transfer Learning}

The intuitive transfer learning design (\ref{eq:well specified - constrained linear regression - solution - target task - positive alpha}) can perform excellently in a particular range of overparameterized settings; however, this performance can significantly degrade as the overparameterization further increases.  This degradation is particularly evident for  $\widetilde{\mtx{H}}=\mtx{H}$ where $\mtx{H}$ is non-orthonormal, see Figs.~\ref{fig:Hgaussian75delta1_HtildeGaussian75delta1_error_curves_wellspec}, \ref{fig:Hgaussian25delta1_HtildeGaussian25delta1_error_curves_wellspec}, \ref{fig:Hgaussian75delta1_HtildeGaussian75delta1_error_curves_misspec_Hrho2}, \ref{fig:Hgaussian25delta1_HtildeGaussian25delta1_error_curves_misspec_Hrho2}, \ref{fig:Hgaussian75dct_HtildeGaussian75dct_error_curves_wellspec} (in contrast, an orthonormal $\mtx{H}$ has a condition number 1 and therefore the related matrix inversion does not lead to significant degradation, compared to ridge regression, in the overparameterized regime, see Fig.~\ref{fig:error_curves_diagrams_isotropic_H_is_orthonormal}).  Motivated by this degradation behavior, and since (\ref{eq:well specified - constrained linear regression - solution - target task - positive alpha}) is a linear estimator, we now turn to explore the optimal linear solution to our transfer learning problem.

\begin{theorem}
\label{theorem:LMMSE Transfer Learning}
Consider $\vecgreek{\beta}$ as a zero mean random vector with a known covariance matrix $\mtx{B}_d$. 
Then, the linear MMSE (LMMSE) estimate of $\vecgreek{\beta}$ given the target dataset $\mtx{X},\vec{y}$ and the precomputed source task solution $\widehat{\vecgreek{\theta}}$: 
\begin{equation}
\label{eq:LMMSE Transfer Learning - estimator formula}
\widehat{\vecgreek{\beta}}_{\rm LMMSE} = \begin{bmatrix}
\mtx{B}_d \mtx{X}^T & \expectation{\vecgreek{\beta}\widehat{\vecgreek{\theta}}^T}
\end{bmatrix} 
{\begin{bmatrix}
	\mtx{X}\mtx{B}_d \mtx{X}^T + \sigma_{\epsilon}^2 \mtx{I}_d & \mtx{X}\expectation{\vecgreek{\beta}\widehat{\vecgreek{\theta}}^T} \\
	\expectation{\widehat{\vecgreek{\theta}}\vecgreek{\beta}^T}\mtx{X}^T & \expectation{\widehat{\vecgreek{\theta}}\widehat{\vecgreek{\theta}}^T}
	\end{bmatrix}}^{-1}
\begin{bmatrix}
\vec{y} \\ \widehat{\vecgreek{\theta}}
\end{bmatrix}
\end{equation}
where $\expectation{\vecgreek{\beta}\widehat{\vecgreek{\theta}}^T} = \left({\begin{cases}
	1 & \text{for}~~ d\le\widetilde{n} \\
	\frac{\widetilde{n}}{d} & \text{for}~~ d>\widetilde{n}
	\end{cases}}\right)\times \mtx{B}_d \mtx{H}^T$
 and 
 \begin{equation}
\label{eq:LMMSE Transfer Learning - theta hat covariance term}
\expectation{\widehat{\vecgreek{\theta}}\widehat{\vecgreek{\theta}}^T} = \begin{cases}
	\mathmakebox[23em][l]{\mtx{K} + \left({\frac{\sigma_{\eta}^2}{d} + \frac{\sigma_{\xi}^2}{\widetilde{n}-d-1} }\right)\mtx{I}_d} \text{for } d\le\widetilde{n}-2 \\
	\mathmakebox[23em][l]{\infty} \text{for } \widetilde{n}-1\le d\le\widetilde{n}+1 \\
	\mathmakebox[23em][l]{\frac{\widetilde{n}}{d}\left({ \frac{\widetilde{n}+1}{d+1}\mtx{K} + 
	\frac{d-\widetilde{n}}{d^2 -1} {\rm diag}\left({ \{ \mtxtrace{\mtx{K}} - k_{jj} \}_{j=1,\dots,d} }\right) +		
		  \left({\frac{\sigma_{\eta}^2}{d} + \frac{\sigma_{\xi}^2}{d-\widetilde{n}-1} }\right)\mtx{I}_d}\right)} 
	  \\
	  \mathmakebox[23em][l]{~}\text{for } d\ge\widetilde{n}+2
	\end{cases}
	\nonumber
\end{equation}
where $\mtx{K} \triangleq \mtx{H}\mtx{B}_d \mtx{H}^T$ and $k_{jj}$ is its $(j,j)^{\rm th}$ component.
\end{theorem}
 The proof of Theorem \ref{theorem:LMMSE Transfer Learning} is provided in Appendix \ref{appendix:sec:proof of LMMSE theorem}. 
 According to the formulation in the theorem, the LMMSE naturally includes additional isotropic regularization (addition of a scaled $\mtx{I}_d$) in the matrix inversion for the $\widehat{\vecgreek{\theta}}$ component, which addresses the problem of the intuitive transfer learning with $\widetilde{\mtx{H}} = \mtx{H}$ when $\mtx{H}$ is poorly conditioned. In this way, we can think of intuitive transfer learning with $\widetilde{\mtx{H}} = \mtx{I}_d$ as a coarse approximation to the LMMSE, except it does not use $\mtx{H}$. Of course, the LMMSE will always perform better given complete knowledge.

The empirically-evaluated generalization errors of the LMMSE transfer learning solution are denoted by the purple markers in Figs.~\ref{fig:error_curves_diagrams_isotropic_general_H__with_LMMSE_TL}, \ref{fig:error_curves_diagrams_isotropic_general_H__with_LMMSE_TL__HtildeI}. 
As expected, the errors of the LMMSE solution (for a known $\mtx{H}$) lower bound the errors of the intuitive (linear) design to transfer learning from (\ref{eq:well specified - constrained linear regression - solution - target task - positive alpha}) with $\widetilde{\mtx{H}}=\mtx{H}$ (Fig.~\ref{fig:error_curves_diagrams_isotropic_general_H__with_LMMSE_TL}) or $\widetilde{\mtx{H}}=\mtx{I}_d$ (Fig.~\ref{fig:error_curves_diagrams_isotropic_general_H__with_LMMSE_TL__HtildeI}).


The results for general $\mtx{H}$ and $\mtx{\Sigma}_{\vec{x}}$ (Fig.~\ref{fig:error_curves_diagrams_isotropic_general_H__with_LMMSE_TL}) show that the LMMSE significantly improves the intuitive TL design at the highly overparameterized settings. 
Note that the LMMSE solution form in (\ref{eq:LMMSE Transfer Learning - estimator formula}) is obtained by a linear processing of the ``effective measurements" vector $\begin{bmatrix}\vec{y}\\\widehat{\vecgreek{\theta}}\end{bmatrix}$ that  concatenates the source task solution to training data of the target task. Accordingly, the significant performance gains due to the LMMSE solution suggest that the common intuition to transfer learning implementation can sometimes be far from achieving the potential benefits of using the source task.

\section{Conclusions}
\label{sec:Conclusions}
We have established a new perspective on transfer learning as a regularizer of overparameterized learning. We defined a transfer learning process between two linear regression tasks such that the target task is optimized with regularization on the distance of its learned parameters from parameters transferred from an already computed source task. We showed that the examined transfer learning method resolves the peak in the generalization errors of the minimum $\ell_2$-norm solution to the target task. We demonstrated that if the source task is sufficiently related to the target task and solved in sufficient accuracy,  then optimally tuned transfer learning can significantly outperform optimally tuned ridge regression over a wide range of parameterization levels. Remarkably, we show that our transfer learning can perform well also without knowing the true task relation, and in various cases to outperform utilization of the true task relation. 
The generalization performance of our transfer learning can degrade at very high overparameterization levels. Hence, we show that this issue can be resolved by implementing the linear MMSE solution to transfer learning, whose form poses interesting questions on the common intuition to transfer learning designs. 
Future extensions may study other optimization formulations such as hybrid regularizers that merge the ideas of ridge regression and parameter transfer, and additional task relation models and their utilization in the transfer learning process.
Moreover, future work may use other analysis tools and optimality definitions (e.g., minimax optimality) to further understand the performance of the transfer learning methods that we proposed in this paper.

\appendix

\section{Additional Details for Section \ref{sec:Problem Settings: Transfer Learning between Linear Regression Tasks}}
\label{appendix:sec:Additional Details for Section of The Test Error of the Source Task}

\subsection{The Test Error of the Source Task}
\label{appendix:subsec:The Test Error of the Source Task}

The test input-response pair $\left( { \vec{z}^{(\rm test)}, v^{(\rm test)} } \right)$  is independently drawn from the $\left(\vec{z},v\right)$ distribution defined above. 
Given the input $\vec{z}^{(\rm test)}$, the source task goal is to estimate the response value $v^{(\rm test)}$ by the value ${\widehat{v} \triangleq \vec{z}^{({\rm test}),T}\widehat{\vecgreek{\theta}}}$, where $\widehat{\vecgreek{\theta}}$ is learned using  $\widetilde{\mathcal{D}}$.  We can assess the generalization performance of the source task using the test squared error ${\mathcal{E}_{\rm src}} \triangleq \expectation{ \left( \widehat{v} - v^{(\rm test)} \right)^2  } =   \sigma_{\xi}^2 + \expectation{ \left \Vert { \widehat{\vecgreek{\theta}} - \vecgreek{\theta} } \right \Vert _2^2 }$,
where the expectation in the definition of ${\mathcal{E}_{\rm src}}$ is with respect to the test data $\left( { \vec{z}^{(\rm test)}, v^{(\rm test)} } \right)$ and the training data ${\widetilde{\mathcal{D}}}$.  Note that $\widehat{\vecgreek{\theta}}$ is a function of the training data. A lower value of ${\mathcal{E}_{\rm src}}$ reflects better generalization performance of the source task.

The source test error of the minimum $\ell_2$-norm solution (\ref{eq:linear regression - source data class}) can be formulated in non-asymptotic settings as 
\begin{equation}
	\label{eq:out of sample error - source task}
	{\mathcal{E}_{\rm src}} = 
	\begin{cases}
		\mathmakebox[15em][l]{\left({ 1 + \frac{d}{\widetilde{n}-d-1} }\right) \sigma_{\xi}^2}        \text{for } d \le \widetilde{n}-2,  
		\\
		\mathmakebox[15em][l]{\infty} \text{for } \widetilde{n}-1 \le d \le \widetilde{n}+1,
		\\
		\mathmakebox[15em][l]{\left({ 1 + \frac{\widetilde{n}}{d-\widetilde{n}-1} }\right)  \sigma_{\xi}^2  + \left({ 1 - \frac{\widetilde{n}}{d} }\right) \Ltwonorm{\vecgreek{\theta}} } \text{for } d \ge \widetilde{n}+2, 
	\end{cases}
\end{equation}		
which is a particular case of the results in \cite{belkin2020two,dar2020double}.

\subsection{The Well-Specified Problem Setting at Different Resolutions}
\label{appendix:subsec:The Operator H at Different Resolutions}

The target task solution is based on estimating the parameter vector $\vecgreek{\beta}\in\mathbb{R}^d$. 
Let us assume that $\vecgreek{\beta}$ is a random vector that follows a Gaussian \textit{prior distribution} $\mathcal{N}(\vec{0},\mtx{B}_d)$ where $\mtx{B}_d$ is the $d\times d$ covariance matrix of $\vecgreek{\beta}$. For considering the \textit{same problem} at \textit{different resolutions} we will assume that ${\expectation{\Ltwonorm{\vecgreek{\beta}}}=\mtxtrace{\mtx{B}_d}=\omega_{\vecgreek{\beta}}}$ where $\omega_{\vecgreek{\beta}}$ is the same positive real constant for all $d,n$. 
For example, the last assumption is satisfied by $\mtx{B}_d=\frac{1}{d}\mtx{I}_d$ that corresponds to an isotropic Gaussian prior distribution for $\vecgreek{\beta}$ and a constant ${\omega_{\vecgreek{\beta}}=1}$. 

Next we characterize the way that the parameter vector of the \textit{source} task and the relation model to the target task behave at the resolution induced by the dimension $d$. 
The relation between $\vecgreek{\theta}$ and $\vecgreek{\beta}$ as presented in (\ref{eq:theta-beta relation}) implies that $\vecgreek{\theta}$ is a random vector that is defined by a noisy linear transformation of the random vector $\vecgreek{\beta}$. Specifically, the distribution of $\vecgreek{\theta}$ is multivariate Gaussian with zero mean and covariance matrix ${\mtx{H}\mtx{B}_d\mtx{H}^{T} + \frac{\sigma_{\eta}^2}{d}\mtx{I}_d}$.
Similar to the above case of the target task parameters, for examining the same problem at different resolutions we assume that ${\expectation{\Ltwonorm{\vecgreek{\theta}}}=\mtxtrace{\mtx{H}\mtx{B}_d\mtx{H}^{T} + \frac{\sigma_{\eta}^2}{d}\mtx{I}_d}=\omega_{\vecgreek{\theta}}}$ where ${\omega_{\vecgreek{\theta}}}$ is the same positive real constant for all $d,n,\widetilde{n}$. In the case of ${\mtx{B}_d=\frac{1}{d}\mtx{I}_d}$, the last assumption is satisfied for ${\frac{1}{d}\Frobnorm{\mtx{H}}=\omega_{\mtx{H}}}$ where $\omega_{\mtx{H}}$ is the same positive real constant for all $d,n,\widetilde{n}$. 
Note the following examples that satisfy this rule: 
\begin{enumerate}
	\item \textit{Transformation to another basis}: $\mtx{H}=\mtx{\Psi}_{d}^{T}$ where $\mtx{\Psi}_{d}$ is a $d\times d$ real orthonormal matrix, i.e., $\mtx{\Psi}_{d}^T \mtx{\Psi}_{d}=\mtx{I}_{d}$. Examples for such $\mtx{\Psi}_{d}$ are the $d\times d$ forms of the identity matrix, discrete cosine transform (DCT) matrix, Hadamard matrix (the case of Hadamard is defined only for $d$ values that satisfy its recursive construction). In this setting, we study the generalization performance versus the dimension $d$ that is coupled with a $d\times d$ orthonormal matrix $\mtx{\Psi}$ of the same type (e.g., DCT). 
	
	\item \textit{Circular convolution operation}: In this case, $\mtx{H}$ is a $d\times d$ circulant matrix that can be interpreted as a discrete version of the circular convolution kernel ${h_{\rm ker}:[0,1]\rightarrow \mathbb{R}}$, which is defined over the continuous interval ${[0,1]}$. The function ${h_{\rm ker}}$ is assumed to be smooth.
	Again, $\mtx{H}$ should be properly scaled to ensure ${\frac{1}{d}\Frobnorm{\mtx{H}}=\omega_{\mtx{H}}}$ where $\omega_{\mtx{H}}$ is a constant independent of $d,n,\widetilde{n}$.	
	
\end{enumerate}

~

\section{Details and Proofs for Section \ref{sec:misspecification}}
\label{appendix:sec: misspecification details and proofs}

\subsection{Independent Misspecification with Isotropic Features}
\label{appendix:subsec:Independent Misspecification with Isotropic Features}
Under Assumptions \ref{assumption:misspecification}-\ref{assumption:Independent misspecification with isotropic features} we can develop the test error of the target task as follows. We learn a $d$-dimensional $\widehat{\vecgreek{\beta}}$ and apply it on a test feature vector $\vec{x}^{(\rm test)}\in\mathbb{R}^d$ to obtain the response estimate $\widehat{y}=(\vec{x}^{(\rm test)})^T \widehat{\vecgreek{\beta}}$. Meanwhile, the true test response has the form $y^{({\rm test})} = (\vec{x}^{({\rm test})})^T \vecgreek{\beta} + (\vec{x}_{\rm ms}^{({\rm test})})^T \vecgreek{\beta}_{\rm ms} + \epsilon^{({\rm test})}$.  Then, the corresponding test error is 
\begin{equation}
	\label{appendix:eq:misspecification - test error of target task}
	\bar{\mathcal{E}} \triangleq \expectation{ \left( \widehat{y} - y^{(\rm test)} \right)^2 } = \sigma_{\epsilon}^2 + \expectation{ \left \Vert { \widehat{\vecgreek{\beta}} - \vecgreek{\beta} } \right \Vert _{\mtx{\Sigma}_{\vec{x}}}^2 } + \expectation{ \left \Vert { \vecgreek{\beta}_{\rm ms} } \right \Vert _{2}^2 }.
\end{equation}
Now, consider a well specified model that corresponds to $y = \vec{x}^T \vecgreek{\beta} + \widetilde{\epsilon}$,
where ${\vec{x}\sim \mathcal{N}\left(\vec{0},\mtx{\Sigma}_{\vec{x}}\right)}$ is $d$-dimensional and ${\widetilde{\epsilon}\sim\mathcal{N}\left(0,\widetilde{\sigma_{\epsilon}}^2\right)}$ with $\widetilde{\sigma_{\epsilon}}^2=\sigma_{\epsilon}^2 + \expectation{ \left \Vert { \vecgreek{\beta}_{\rm ms} } \right \Vert _{2}^2}$. Then, the test error of this well-specified model is the same as the test error of the misspecified model from (\ref{appendix:eq:misspecification - test error of target task}).

Next, under Assumptions \ref{assumption:misspecification}-\ref{assumption:Independent misspecification with isotropic features} (specifically, recall that $\vecgreek{\beta}$ and $\vecgreek{\beta}_{\rm ms}$ are independent), the misspecified task relation model $\vecgreek{\theta} = \mtx{H}\vecgreek{\beta} + {\mtx{H}_{\rm ms}} \vecgreek{\beta}_{\rm ms} + \vecgreek{\eta}$ implies that $\vecgreek{\theta}$ has zero mean and covariance 
\begin{align}
	\label{appendix:covariance of misspecified theta}
	\expectation{\vecgreek{\theta}\vecgreek{\theta}^T}&= \mtx{H}\expectation{\vecgreek{\beta}\vecgreek{\beta}^T}\mtx{H}^T + {\mtx{H}_{\rm ms}}\expectation{ \vecgreek{\beta}_{\rm ms}\vecgreek{\beta}_{\rm ms}^T}{\mtx{H}_{\rm ms}^T} + \expectation{\vecgreek{\eta}\vecgreek{\eta}^T}
	\nonumber\\
	&= \mtx{H}\mtx{B}_d\mtx{H}^T + b_{\rm ms}{\mtx{H}_{\rm ms}}{\mtx{H}_{\rm ms}^T} + \sigma_{\eta}^2\mtx{I}_d
	\nonumber\\
	&= \mtx{H}\mtx{B}_d\mtx{H}^T + b_{\rm ms}\rho\mtx{I}_d + \sigma_{\eta}^2\mtx{I}_d
	\nonumber\\
	&= \mtx{H}\mtx{B}_d\mtx{H}^T + (b_{\rm ms}\rho + \sigma_{\eta}^2)\mtx{I}_d
\end{align}
where we denote $\expectation{ \vecgreek{\beta}_{\rm ms}\vecgreek{\beta}_{\rm ms}^T}=b_{\rm ms}\mtx{I}_q$ for some constant $b_{\rm ms}\ge0$.
Accordingly, we define $\widetilde{\vecgreek{\eta}} \triangleq {\mtx{H}_{\rm ms}} \vecgreek{\beta}_{\rm ms} + \vecgreek{\eta}$ and note that $\widetilde{\vecgreek{\eta}}\sim\mathcal{N}(\vec{0},(b_{\rm ms}\rho + \sigma_{\eta}^2)\mtx{I}_d)$ is independent of $\vecgreek{\beta}$. 
Hence, the well-specified task relation $\vecgreek{\theta} = \mtx{H}\vecgreek{\beta} + \widetilde{\vecgreek{\eta}}$ is equivalent to the above misspecified model.

Recall that in Section \ref{sec:misspecification} we assume that $\expectation{\Ltwonorm{\vecgreek{\beta}}+\Ltwonorm{\vecgreek{\beta}_{\rm ms}}}=\omega_{\vecgreek{\beta}_{\rm all}}$ for the same constant $\omega_{\vecgreek{\beta}_{\rm all}}$ for all $d,q$, and that $\expectation{\Ltwonorm{\vecgreek{\beta}_{\rm ms}}}/\omega_{\vecgreek{\beta}_{\rm all}}=\left(1+\frac{d}{n}\right)^{-a}$ for $a>0$. This implies that the variance of a misspecified parameter is $b_{\rm ms}=\frac{\omega_{\vecgreek{\beta}_{\rm all}}}{q}\left(1+\frac{d}{n}\right)^{-a}$, which reflects the reduction in the misspecification level as the number of utilized features $d$ increases.

\subsection{Additional Examples for Generalization Performance in Misspecified Settings}
\label{appendix:subsec:Additional Examples for Generalization Performance in Misspecified Settings}

\begin{figure*}[t]
	\centering
	\subfloat[$\widetilde{\mtx{H}}=\mtx{H}$]{\includegraphics[width=0.4\textwidth]{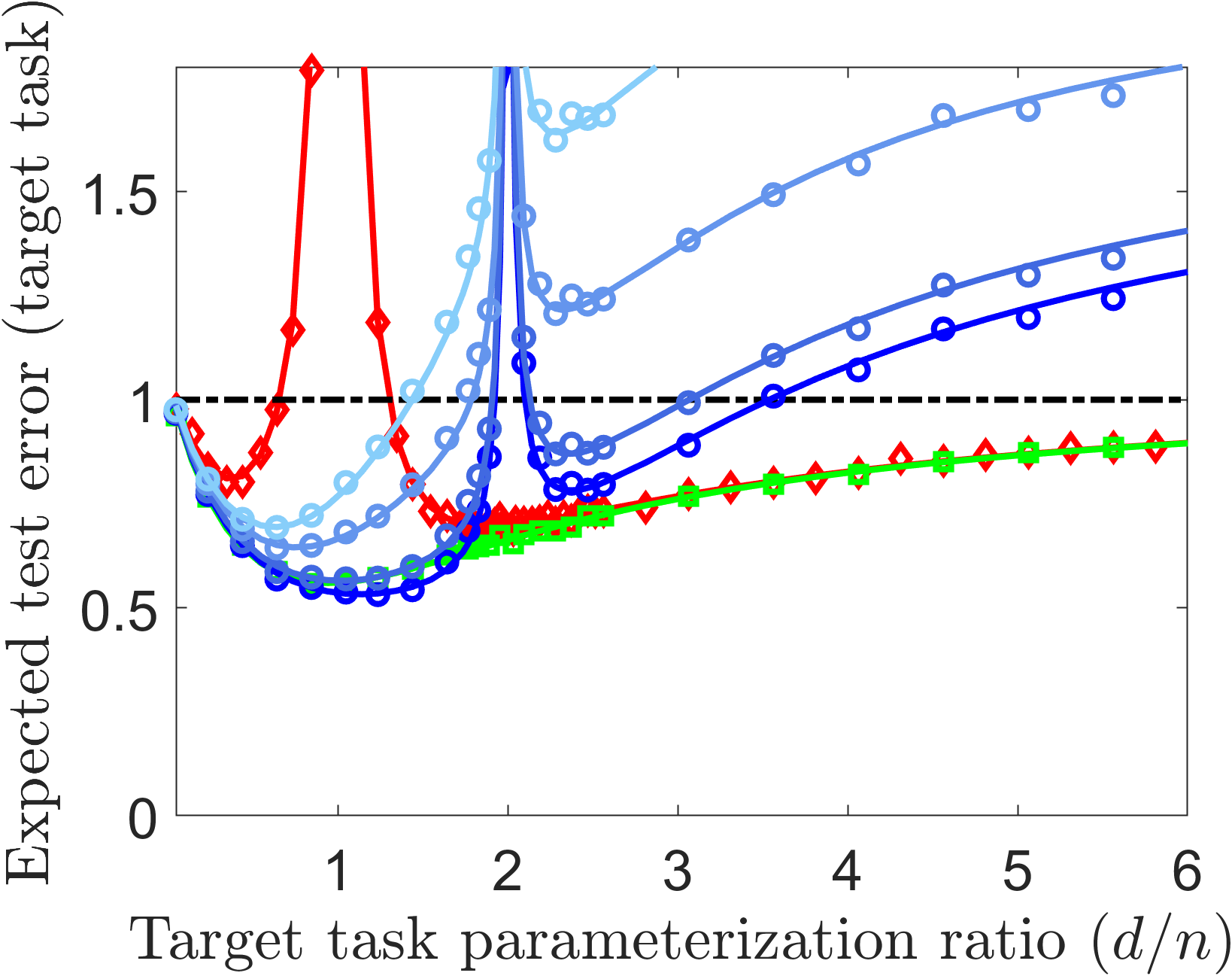}\label{fig:Hgaussian75delta1_HtildeGaussian75delta1_error_curves_misspec_Hrho25}}
	\subfloat[$\widetilde{\mtx{H}}=\mtx{I}_d$]{\includegraphics[width=0.4\textwidth]{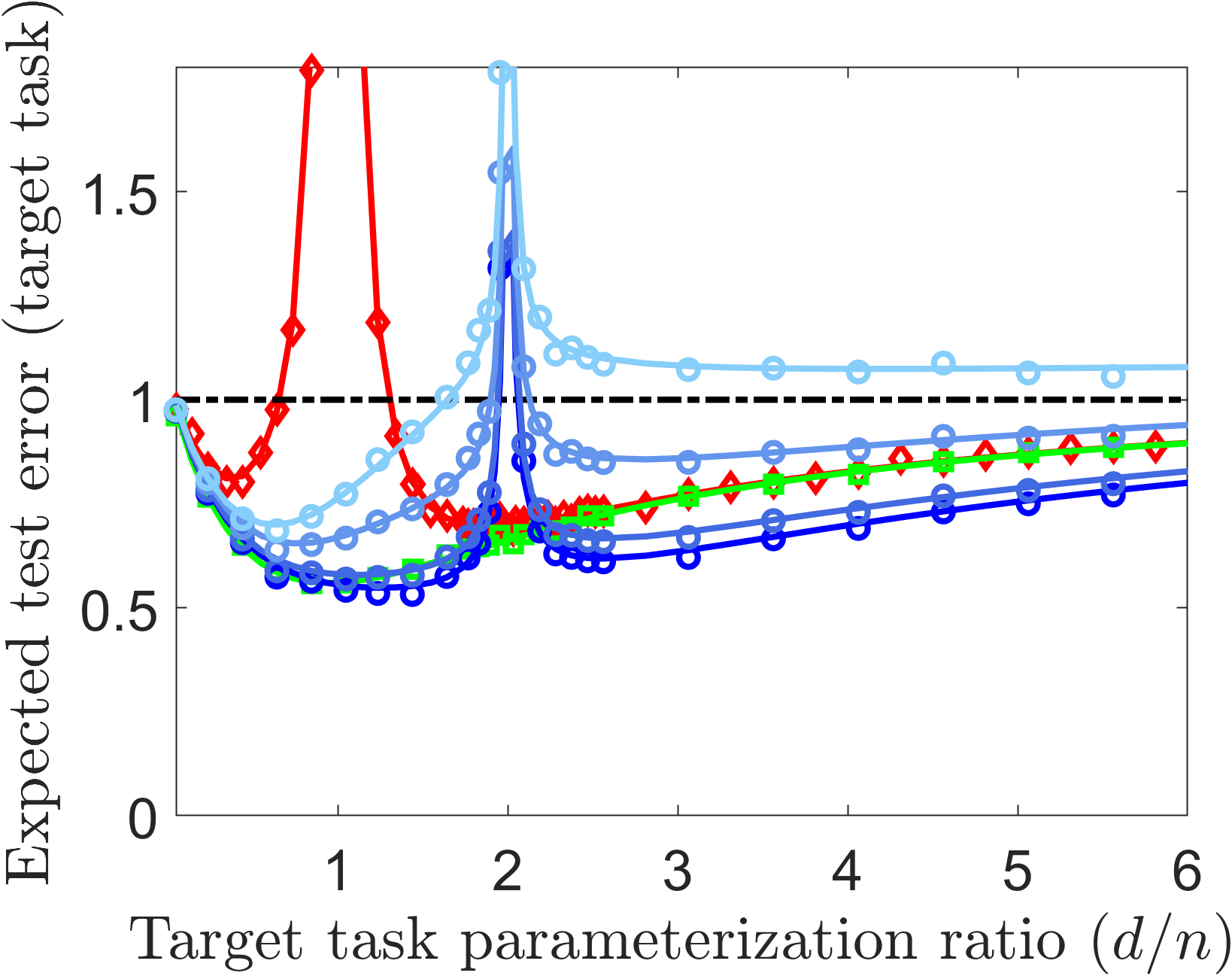}\label{fig:Hgaussian75delta1_HtildeI_error_curves_misspec_Hrho25}}
	\subfloat{\includegraphics[width=0.16\textwidth]{figures/error_curve_legend_without_LMMSE.png}}
	\caption{The test error of the target task under isotropic Gaussian assumption on $\vecgreek{\beta}$ and isotropic target features. The matrix $\mtx{H}$ is a $d\times d$ circulant matrix corresponding to the discrete version of the continuous-domain convolution kernel ${h_{\rm ker}(\tau)=\delta(\tau)+ e^{-\frac{\lvert\tau-0.5\rvert}{w_{\rm ker}}}}$, here the kernel width is ~${w_{\rm ker}=2/75}$. 
		Both subfigures correspond to misspecified models according to Assumptions \ref{assumption:misspecification}-\ref{assumption:Independent misspecification with isotropic features} and polynomial reduction with $a=2.5$, $q=500$, $\rho=25$.	The number of data samples for the target task is $n=64$ and for the source task is $\widetilde{n}=128$. }
	\label{fig:error_curves_misspec_rho_25}
\end{figure*}

In Fig.~\ref{fig:error_curves_misspec_rho_25} we present the test error evaluations for the same setting of the convolutional $\mtx{H}$ as in Figs.~\ref{fig:Hgaussian75delta1_HtildeGaussian75delta1_error_curves_misspec_Hrho2}, \ref{fig:Hgaussian75delta1_HtildeI_error_curves_misspec_Hrho2} but with a stronger misspecification level of $\rho=25$ in the task relation.


\section{Proofs for Section \ref{subsec:Analysis for H of an Orthonormal Matrix Form}}
\label{appendix:sec:proofs for section 3}

Recall that in Section \ref{sec:Analysis for H of an Orthonormal Matrix Form} we consider $\widetilde{\mtx{H}}=\mtx{H}$ and, therefore, the corresponding proofs directly consider $\mtx{H}$ instead of $\widetilde{\mtx{H}}$. 
\subsection{Proof of Lemma \ref{lemma:well specified - out of sample error - target task -  nonzero alpha - expectation of eigenvalues - orthonormal H}}
\label{appendix:subsec:proof of lemma 3.1}
The expected test error of the transfer learning solution to the target task is developed as follows. 

\begin{align}
\label{appendix:eq:proof of lemma 3.1 - error development 1}
&\bar{\mathcal{E}}_{\rm TL} \triangleq \expectationwrt{\mathcal{E}_{\rm TL}}{\vecgreek{\beta}} = \sigma_{\epsilon}^2 + \expectation{ \left \Vert { \widehat{\vecgreek{\beta}}_{\rm TL} - \vecgreek{\beta} } \right \Vert _2^2 }
\nonumber\\
&= \sigma_{\epsilon}^2 + \expectation{ \left \Vert { \left({ \mtx{X}^{T} \mtx{X} + n{\alpha_{\rm TL}}\mtx{H}^{T}\mtx{H} }\right)^{-1} \left( { \mtx{X}^{T} \vec{y} + n{\alpha_{\rm TL}}\mtx{H}^{T} \widehat{\vecgreek{\theta}} }\right) - \vecgreek{\beta} } \right \Vert _2^2 }
\nonumber\\
&= \sigma_{\epsilon}^2 + \expectation{ \left \Vert { \left({ \mtx{X}^{T} \mtx{X} + n{\alpha_{\rm TL}}\mtx{H}^{T}\mtx{H} }\right)^{-1} \left( { \mtx{X}^{T} \vecgreek{\epsilon} + n{\alpha_{\rm TL}}\mtx{H}^{T} \left({\vecgreek{\eta}+\left(\widehat{\vecgreek{\theta}}-\vecgreek{\theta}\right)}\right) }\right) } \right \Vert _2^2 }
\end{align}
where we used the relation ${\widehat{\vecgreek{\theta}}=\vecgreek{\theta}+\left(\widehat{\vecgreek{\theta}}-\vecgreek{\theta}\right)=\mtx{H}\vecgreek{\beta}+\vecgreek{\eta}+\left(\widehat{\vecgreek{\theta}}-\vecgreek{\theta}\right)}$ and that\newline $\left({ \mtx{X}^{T} \mtx{X} + n{\alpha_{\rm TL}}\mtx{H}^{T}\mtx{H} }\right)$ is always invertible under the full-rank assumption on $\mtx{H}$ (i.e., Assumption \ref{assumption: H is full rank}). 

Lemma \ref{lemma:well specified - out of sample error - target task -  nonzero alpha - expectation of eigenvalues - orthonormal H} considers the case where $\mtx{H}=\mtx{\Psi}^T$ where $\mtx{\Psi}$ is a $d\times d$ orthonormal matrix. Hence, ${\mtx{\Psi}^{T}\mtx{\Psi}=\mtx{\Psi}\mtx{\Psi}^{T}=\mtx{I}_{d}}$. We denote ${\mtx{X}_{\mtx{\Psi}}\triangleq \mtx{X}\mtx{\Psi}}$. Because the rows of $\mtx{X}$ have isotropic Gaussian distributions then their transformations by $\mtx{\Psi}^{T}$ do not change their distribution. Then, we can develop the error expression for $\bar{\mathcal{E}}_{\rm TL}$ from (\ref{appendix:eq:proof of lemma 3.1 - error development 1}) into 
\begin{equation}
\label{appendix:eq:proof of lemma 3.1 - error development 2}
\bar{\mathcal{E}}_{\rm TL} = \sigma_{\epsilon}^2 + \mtxtrace{ \expectation{ \left({ \mtx{X}_{\mtx{\Psi}}^{T} \mtx{X}_{\mtx{\Psi}} + n{\alpha_{\rm TL}}\mtx{I}_{d} }\right)^{-2}  \left( { \sigma_{\epsilon}^{2} \mtx{X}_{\mtx{\Psi}}^{T} \mtx{X}_{\mtx{\Psi}}  + n^2{\alpha_{\rm TL}^2} \mtx{\Gamma}_{\rm TL}   }\right) } }
\end{equation}
where 
\begin{equation}
\label{appendix:eq:proof of lemma 3.1 - Gamma TL matrix}
\mtx{\Gamma}_{\rm TL}\triangleq \expectation{\vecgreek{\eta}\vecgreek{\eta}^T} + \expectation{\left(\widehat{\vecgreek{\theta}}-\vecgreek{\theta}\right)\left(\widehat{\vecgreek{\theta}}-\vecgreek{\theta}\right)^{T}} + \expectation{\left(\widehat{\vecgreek{\theta}}-\vecgreek{\theta}\right)\vecgreek{\eta}^{T}} + \expectation{\vecgreek{\eta}\left(\widehat{\vecgreek{\theta}}-\vecgreek{\theta}\right)^{T}}
\end{equation}
Now we provide two fundamental results that are useful for the following developments. 
The first result is about the $\widetilde{n}\times d$ matrix $\mtx{Z}$ that has i.i.d.~standard Gaussian components, therefore the expectation of the $d\times d$ projection matrix $\mtx{Z}^{+} \mtx{Z}$ is formulated (almost surely) as 
\begin{equation}
\label{appendix:proof of proposition 3.4 - random projection of Gaussian iid matrix}
\expectation{ \mtx{Z}^{+} \mtx{Z}} = \mtx{I}_{d}\times \begin{cases}
\mathmakebox[3em][l] { 1 }   \text{for } d \le \widetilde{n},  
\\
\mathmakebox[3em][l] {\frac{\widetilde{n}}{d} }   \text{for } d > \widetilde{n}. 
\end{cases} 
\end{equation} 
The second fundamental result is on the expectation of the pseudoinverse of the $d\times d$ Wishart matrix $\mtx{Z}^{T} \mtx{Z}$ that almost surely satisfies 
\begin{equation}
\label{appendix:proof of proposition 3.4 - pinvZ pinvZtransposed}
\expectation{  \left(\mtx{Z}^{T} \mtx{Z}\right)^{+} } = \expectation{ \mtx{Z}^{+} \left(\mtx{Z}^{+} \right)^{T} } = \mtx{I}_{d}\times \begin{cases}
\mathmakebox[6em][l] { \frac{1}{\widetilde{n}-d-1} }   \text{for } d \le \widetilde{n}-2,  
\\
\mathmakebox[6em][l] { \infty }   \text{for } \widetilde{n}-1 \le d \le \widetilde{n}+1,  
\\
\mathmakebox[6em][l] {\frac{\widetilde{n}}{d} \cdot \frac{1}{d-\widetilde{n}-1}}   \text{for } d \ge \widetilde{n}+2. 
\end{cases} 
\end{equation}
The last result can be proved using the tools given in Theorem 1.3 of \cite{breiman1983many}.

Using the auxiliary results (\ref{appendix:proof of proposition 3.4 - random projection of Gaussian iid matrix})-(\ref{appendix:proof of proposition 3.4 - pinvZ pinvZtransposed}) we get that 
\begin{align}
\label{appendix:eq:proof of lemma 3.1 - source error covariance for orthonormal H}
&\expectation{\left(\widehat{\vecgreek{\theta}}-\vecgreek{\theta}\right)\left(\widehat{\vecgreek{\theta}}-\vecgreek{\theta}\right)^{T}} = 
\expectation{\left(\mtx{Z}^{+} \mtx{Z}\vecgreek{\theta} + \mtx{Z}^{+}\vecgreek{\xi} -\vecgreek{\theta}\right)\left(\mtx{Z}^{+} \mtx{Z}\vecgreek{\theta} + \mtx{Z}^{+}\vecgreek{\xi}-\vecgreek{\theta}\right)^{T}} 
\nonumber \\
& = \mtx{I}_{d} \times  \begin{cases}
\mathmakebox[15em][l]{ \frac{\sigma_{\xi}^2}{\widetilde{n}-d-1} }        \text{for } d \le \widetilde{n}-2,  
\\
\mathmakebox[15em][l]{\infty} \text{for } \widetilde{n}-1 \le d \le \widetilde{n}+1,
\\
\mathmakebox[15em][l]{ \frac{\widetilde{n}}{d}\cdot\frac{\sigma_{\xi}^2}{d-\widetilde{n}-1}     + \left({ 1 - \frac{\widetilde{n}}{d} }\right) \frac{b+\sigma_{\eta}^2}{d} } \text{for } d \ge \widetilde{n}+2
\end{cases}
\end{align}
and 
\begin{equation}
\label{appendix:eq:proof of lemma 3.1 - source error cross-covariance for orthonormal H}
\expectation{\left(\widehat{\vecgreek{\theta}}-\vecgreek{\theta}\right)\vecgreek{\eta}^{T}} = \expectation{\vecgreek{\eta}\left(\widehat{\vecgreek{\theta}}-\vecgreek{\theta}\right)^{T}} = 
\begin{cases}
\mathmakebox[6em][l]{ \mtx{0} }        \text{for } d \le \widetilde{n},  
\\
\mathmakebox[6em][l]{ \left({ \frac{\widetilde{n}}{d} - 1 }\right)\frac{\sigma_{\eta}^2}{d}\mtx{I}_{d} } \text{for } d > \widetilde{n}
\end{cases}
\end{equation}
where we also used the result $\expectationwrt{\vecgreek{\theta}\vecgreek{\theta}^{T}}{\vecgreek{\beta},\vecgreek{\eta}}=\frac{b+\sigma_{\eta}^2}{d}\mtx{I}_{d}$, which is due to the task relation model (\ref{eq:theta-beta relation}), Assumption \ref{assumption:isotropic prior on beta - well specified}, and because $\mtx{H}=\mtx{\Psi}^T$ where $\mtx{\Psi}$ is a $d\times d$ orthonormal matrix.
Hence, the matrix form in (\ref{appendix:eq:proof of lemma 3.1 - source error covariance for orthonormal H}), which is a scaled identity matrix, lets us to express the error formula from (\ref{appendix:eq:proof of lemma 3.1 - error development 2}) as 
\begin{equation}
\label{appendix:proof of lemma 3.1 - expression to prove - error}
\bar{\mathcal{E}}_{\rm TL} = 
\sigma_{\epsilon}^2  + \mathbb{E}\Biggl\{{ \sum_{k=1}^{d} {\frac{ n^2 {{\alpha_{\rm TL}^2}} C_{\rm TL} + \sigma_{\epsilon}^2 \cdot\eigenvalue{\mtx{X}_{\mtx{\Psi}}^{T} \mtx{X}_{\mtx{\Psi}}}{k} }{\left({\eigenvalue{\mtx{X}_{\mtx{\Psi}}^{T} \mtx{X}_{\mtx{\Psi}}}{k} + n{\alpha_{\rm TL}}}\right)^{2} } } }\Biggr\}
\end{equation}
where $\eigenvalue{{\mtx{X}}_{\mtx{\Psi}}^{T} \mtx{X}_{\mtx{\Psi}}}{k}$ is the $k^{\rm th}$ eigenvalue of the $d\times d$ matrix $\mtx{X}_{\mtx{\Psi}}^{T} \mtx{X}_{\mtx{\Psi}}$, and 
\begin{equation}
\label{appendix:proof of lemma 3.1 - expression to prove - C TL}
C_{\rm TL} \triangleq
\begin{cases}
\mathmakebox[16em][l] {\frac{\sigma_{\eta}^2}{d} + \frac{\sigma_{\xi}^2}{\widetilde{n}-d-1} }   \text{for } d \le \widetilde{n}-2,  
\\
\mathmakebox[16em][l]{\infty} \text{for } \widetilde{n}-1 \le d \le \widetilde{n}+1,
\\
\mathmakebox[16em][l] { \left({1-\frac{\widetilde{n}}{d}}\right) \frac{b}{d} + \frac{\widetilde{n}}{d}\left({\frac{\sigma_{\eta}^2}{d} + \frac{\sigma_{\xi}^2}{d-\widetilde{n}-1}}\right) } \text{for } d \ge \widetilde{n}+2.
\end{cases} 
\end{equation}
This concludes the proof of Lemma \ref{lemma:well specified - out of sample error - target task -  nonzero alpha - expectation of eigenvalues - orthonormal H}.

\subsection{Proof of Theorem \ref{theorem:well specified - out of sample error - target task -  nonzero alpha - expectation of eigenvalues - optimal - orthonormal H}}
\label{appendix:subsec:proof of theorem 3.2}

The derivative of the error expression for $\bar{\mathcal{E}}_{\rm TL}$ as given in Lemma \ref{lemma:well specified - out of sample error - target task -  nonzero alpha - expectation of eigenvalues - orthonormal H} with respect to $\alpha_{\rm TL}$ is 
\begin{equation}
\label{appendix:proof of theorem 3.2 - derivative of error}
\frac{\partial\bar{\mathcal{E}}_{\rm TL}}{\partial \alpha_{\rm TL}} = 
2n \left({\alpha_{\rm TL} n C_{\rm TL} - \sigma_{\epsilon}^2 }\right) \cdot \mathbb{E}\Biggl\{{ \sum_{k=1}^{d} {\frac{ \eigenvalue{\mtx{X}_{\mtx{\Psi}}^{T} \mtx{X}_{\mtx{\Psi}}}{k}  }{\left({\eigenvalue{\mtx{X}_{\mtx{\Psi}}^{T} \mtx{X}_{\mtx{\Psi}}}{k} + n{\alpha_{\rm TL}}}\right)^{3} } } }\Biggr\}.
\end{equation}
Since we consider $\alpha_{\rm TL}>0$ then the necessary condition for optimality, ${\frac{\partial\bar{\mathcal{E}}_{\rm TL}}{\partial \alpha_{\rm TL}}=0}$, yields 
\begin{equation}
\label{appendix:proof of theorem 3.2 - optimal alpha_TL}
\alpha_{\rm TL}^{\rm opt} = \frac{\sigma_{\epsilon}^2}{n C_{\rm TL}}, 
\end{equation}
which is the optimal value of $ \alpha_{\rm TL}>0$ for our transfer learning process when $\mtx{H}=\mtx{\Psi}^T$ is an  orthonormal matrix and ${d\notin\{\widetilde{n}-1,\widetilde{n},\widetilde{n}+1\}}$. 
Next, we set the expression for $\alpha_{\rm TL}^{\rm opt}$ in the error expression for $\bar{\mathcal{E}}_{\rm TL}$ from Lemma \ref{lemma:well specified - out of sample error - target task -  nonzero alpha - expectation of eigenvalues - orthonormal H} and using some algebra gives, for $d\notin\{\widetilde{n}-1,\widetilde{n},\widetilde{n}+1\}$,
\begin{align}
\label{appendix:eq:proof of theorem 3.2 - minimal error expression}
&\bar{\mathcal{E}}_{\rm TL}^{\rm opt}= \sigma_{\epsilon}^2 \left({ 1 + \mathbb{E}\Biggl\{{ \sum_{k=1}^{d} {\frac{ 1 }{ \eigenvalue{\mtx{X}_{\mtx{\Psi}}^{T} \mtx{X}_{\mtx{\Psi}}}{k} + n{\alpha_{\rm TL}^{\rm opt}} } } }\Biggr\} }\right)  
\nonumber\\
&=\sigma_{\epsilon}^2
\left( { 1 + \expectationwrt{  \mtxtrace{ \left( {\mtx{X}_{\mtx{\Psi}}^{T} \mtx{X}_{\mtx{\Psi}} + n\alpha_{\rm TL}^{\rm opt}\mtx{I}_{d}  }\right)^{-1} } }{\mtx{X}_{\mtx{\Psi}} } } \right).
\end{align}
Note that for $d\in\{\widetilde{n}-1,\widetilde{n},\widetilde{n}+1\}$, $C_{\rm TL}=\infty$ and based on the error expression in (\ref{appendix:proof of lemma 3.1 - expression to prove - error}) we get that $\bar{\mathcal{E}}_{\rm TL}=\infty$ for any ${\alpha_{\rm TL}}> 0$. 
This concludes the proof outline for Theorem \ref{theorem:well specified - out of sample error - target task -  nonzero alpha - expectation of eigenvalues - optimal - orthonormal H}.

\subsection{Proof of Theorem \ref{theorem:well specified - out of sample error - target task -  nonzero alpha - expectation of eigenvalues - optimal - asymptotic isotropic - orthonormal H}}
\label{appendix:subsec:proof of theorem 3.3}
In the asymptotic setting (i.e., under Assumption \ref{assumption:Asymptotic settings - well specified}), the optimal parameter $\alpha_{\rm TL}^{\rm opt}$ from (\ref{appendix:proof of theorem 3.2 - optimal alpha_TL}) goes to its limiting value 
\begin{equation}
\label{appendix:eq:proof of theorem 3.3 - limiting value of alpha TL}
\alpha_{{\rm TL}, \infty}^{\rm opt} = 
\sigma_{\epsilon}^2 \gamma_{\rm tgt} \times 
\begin{cases}
\mathmakebox[16em][l]{ \left({\sigma_{\eta}^2 + \frac{\gamma_{\rm src}\cdot \sigma_{\xi}^2 }{1-\gamma_{\rm src}}}\right)^{-1} }   \text{for } d \le \widetilde{n}-2,  
\\
\mathmakebox[16em][l]{ \left({ \frac{\gamma_{\rm src}-1}{\gamma_{\rm src}} b + \frac{1}{\gamma_{\rm src}}\left({ \sigma_{\eta}^2 + \frac{\gamma_{\rm src}\cdot \sigma_{\xi}^2 }{\gamma_{\rm src}-1}}\right) }\right)^{-1} } \text{for } d \ge \widetilde{n}+2 
\end{cases} 
\end{equation}
Moreover, note that the error expression of optimally tuned transfer learning in (\ref{appendix:eq:proof of theorem 3.2 - minimal error expression}) includes the form of ${\expectationwrt{  \mtxtrace{ \left( {\mtx{X}_{\mtx{\Psi}}^{T} \mtx{X}_{\mtx{\Psi}} + n\alpha_{\rm TL}^{\rm opt}\mtx{I}_{d}  }\right)^{-1} } }{\mtx{X}_{\mtx{\Psi}} }}$  where $\mtx{X}_{\mtx{\Psi}}$ is a $n\times d$ random matrix of i.i.d.~Gaussian variables $\mathcal{N}\left(0,1\right)$. This form, however with a different parameter than $\alpha_{\rm TL}^{\rm opt}$, appears also in the analysis of optimally tuned ridge regression by Dobriban and Wager \cite{appendixDobriban2018high}. Accordingly, we can readily use the results from \cite{appendixDobriban2018high} in conjunction with the limiting value of our parameter $\alpha_{{\rm TL}, \infty}^{\rm opt}$ from (\ref{appendix:eq:proof of theorem 3.3 - limiting value of alpha TL}) and get that 
\begin{equation}
\label{appendix:eq:well specified - out of sample error - target task - nonzero alpha - expectation of eigenvalues - optimal - asymptotic isotropic - orthonormal H}
{\bar{\mathcal{E}}_{\rm TL} ^{\rm opt}} \rightarrow 
\sigma_{\epsilon}^2 
\left( { 1 + \gamma_{\rm tgt}\cdot m\left( -\alpha_{{\rm TL}, \infty}^{\rm opt}; \gamma_{\rm tgt} \right) }\right)
\end{equation}
where 
\begin{align}
\label{appendix:eq:well specified - out of sample error - target task - nonzero alpha - expectation of eigenvalues - optimal - asymptotic isotropic - definition of m function - orthonormal H}
m\left( -\alpha_{{\rm TL}, \infty}^{\rm opt}; \gamma_{\rm tgt} \right) = \frac{ -\left({ 1 - \gamma_{\rm tgt} + \alpha_{{\rm TL}, \infty}^{\rm opt} }\right) + \sqrt{ \left({ 1 - \gamma_{\rm tgt} + \alpha_{{\rm TL}, \infty}^{\rm opt} }\right)^2 + 4\gamma_{\rm tgt} \alpha_{{\rm TL}, \infty}^{\rm opt} } }{2\gamma_{\rm tgt} \alpha_{{\rm TL}, \infty}^{\rm opt} }
\end{align}
is the Stieltjes transform of the Marchenko-Pastur distribution, which is the limiting spectral distribution of the sample covariance associated with $n$ samples that are drawn from a Gaussian distribution $\mathcal{N}\left(\vec{0},\mtx{I}_{d}\right)$. This completes the proof outline for Theorem \ref{theorem:well specified - out of sample error - target task -  nonzero alpha - expectation of eigenvalues - optimal - asymptotic isotropic - orthonormal H}.

\subsection{Generalization Error of ML2N Regression Under Assumption \ref{assumption:isotropic prior on beta - well specified}}
\label{appendix:subsec:Generalization Error of OLS Regression Under Isotropic Prior Assumption}
The test error of the ML2N regression solution of the individual target task was provided in (\ref{eq:well specified - out of sample error - target task - no transfer learning - alpha is zero - OLS}) for a given parameter vector $\vecgreek{\beta}$. Then, the expectation of $\mathcal{E}_{\rm OLS}$ from (\ref{eq:well specified - out of sample error - target task - no transfer learning - alpha is zero - OLS}) with respect to the isotropic Gaussian prior on $\vecgreek{\beta}$ (i.e., under Assumption \ref{assumption:isotropic prior on beta - well specified}) is 
\begin{equation}
\label{appendix:eq:Generalization Error of OLS Regression Under Assumption}
\expectationwrt{\mathcal{E}_{\rm ML2N}}{\vecgreek{\beta}} = 
\begin{cases}
\mathmakebox[15em][l]{\left({ 1 + \frac{d}{n-d-1} }\right)  \sigma_{\epsilon}^2 }     \text{for } d \le n-2,  
\\
\mathmakebox[15em][l]{\infty} \text{for } n-1 \le d \le n+1,
\\
\mathmakebox[15em][l]{\left({ 1 + \frac{n}{d-n-1} }\right) \sigma_{\epsilon}^2  + \left({ 1 - \frac{n}{d} }\right) b }  \text{for } d \ge n+2. 
\end{cases}
\end{equation}

\section{Proofs and Details for Section \ref{sec:Transfer Learning versus Ridge Regression}}
\label{appendix:sec:proofs for section 4}

\subsection{Generalization Error of Ridge Regression in Non-Asymptotic Settings}
\label{appendix:subsec:Generalization Error of Ridge Regression at Non-Asymptotic Settings}
The expected test error of the ridge regression solution of the (individual) target task can be developed as outlined next. 
\begin{align}
\label{appendix:eq:ridge details - error development 1}
&\bar{\mathcal{E}}_{\rm ridge} \triangleq \expectationwrt{\mathcal{E}_{\rm ridge}}{\vecgreek{\beta}} = \sigma_{\epsilon}^2 + \expectation{ \left \Vert { \widehat{\vecgreek{\beta}}_{\rm ridge} - \vecgreek{\beta} } \right \Vert _2^2 }
\nonumber\\
&= \sigma_{\epsilon}^2 + \expectation{ \left \Vert { \left({ \mtx{X}^{T} \mtx{X} + n{\alpha_{\rm ridge}}\mtx{I}_{d} }\right)^{-1} \mtx{X}^{T} \vec{y} - \vecgreek{\beta} } \right \Vert _2^2 }
\nonumber\\ 
&= \sigma_{\epsilon}^2 + \expectation{ \left \Vert { \left({ \mtx{X}^{T} \mtx{X} + n{\alpha_{\rm ridge}}\mtx{I}_{d} }\right)^{-1} \mtx{X}^{T} \vecgreek{\epsilon} } \right \Vert _2^2 } \nonumber\\ 
&\quad+ \expectation{ \left \Vert { \left( { \left({ \mtx{X}^{T} \mtx{X} + n{\alpha_{\rm ridge}}\mtx{I}_{d} }\right)^{-1}\mtx{X}^{T} \mtx{X} - \mtx{I}_{d} }\right) \vecgreek{\beta} } \right \Vert _2^2 }
\nonumber\\
\end{align}
where we used the fact that $\vecgreek{\epsilon}$ is independent of $\mtx{X}$. 
Consider the eigendecomposition
\begin{equation}
\label{appendix:eq:eigendecomposition of XX}
{\mtx{X}^{T} \mtx{X} = \mtx{\Phi}_{\mtx{X}}\mtx{\Lambda}_{\mtx{X}}\mtx{\Phi_{\mtx{X}}}^{T}}
\end{equation}
where $\mtx{\Phi}_{\mtx{X}}$ is a $d\times d$ orthonormal matrix with columns being eigenvectors of ${{\mtx{X}}^{T} \mtx{X}}$ and the corresponding eigenvalues  $\eigenvalue{\mtx{X}^{T} \mtx{X}}{k}$, $k=1,\dots,d$, are on the main diagonal of the $d\times d$ diagonal matrix $\mtx{\Lambda}_{\mtx{X}}$.
Then, we can continue to develop (\ref{appendix:eq:ridge details - error development 1}) as follows. 
\begin{align}
\label{appendix:eq:ridge details - error development 2}
&\bar{\mathcal{E}}_{\rm ridge} = \sigma_{\epsilon}^2 + \sigma_{\epsilon}^2 \expectation{ \mtxtrace{ { \left({ \mtx{\Lambda}_{\mtx{X}} + n{\alpha_{\rm ridge}}\mtx{I}_{d} }\right)^{-2} \mtx{\Lambda}_{\mtx{X}} } } } \nonumber\\ 
&\quad+ \frac{b}{d}\expectation{ \mtxtrace{  { \left( { \left({ \mtx{\Lambda}_{\mtx{X}} + n{\alpha_{\rm ridge}}\mtx{I}_{d} }\right)^{-1} \mtx{\Lambda}_{\mtx{X}} - \mtx{I}_{d} }\right)^2 } } }
\nonumber\\
&= \sigma_{\epsilon}^2  + \mathbb{E}\Biggl\{{ \sum_{k=1}^{d} {\frac{ \sigma_{\epsilon}^2\eigenvalue{{\mtx{X}}^{T} \mtx{X}}{k} + \frac{b}{d} n^2 \alpha_{\rm ridge}^2 }{\left({\eigenvalue{{\mtx{X}}^{T} \mtx{X}}{k} + n{\alpha_{\rm ridge}}}\right)^{2} } }}\Biggr\}. 
\end{align}

By equating the derivative (w.r.t.~$\alpha_{\rm ridge}$) of the expression in (\ref{appendix:eq:ridge details - error development 2}) to zero, one can obtain the $\alpha_{\rm ridge}>0$ that minimizes the error $\bar{\mathcal{E}}_{\rm ridge}$. The suggested calculations show that $\alpha_{\rm ridge}^{\rm opt}=\frac{d\sigma_{\epsilon}^2}{n b}$. By setting $\alpha_{\rm ridge}^{\rm opt}$ back in (\ref{appendix:eq:ridge details - error development 2}) one can show that the minimal expected test error for ridge regression is 
\begin{equation}
\label{appendix:eq:well specified - standard ridge regression - optimal - test error}
\bar{\mathcal{E}}_{\rm ridge}^{\rm opt}= 
\sigma_{\epsilon}^2
\left( { 1 + \expectationwrt{  \mtxtrace{ \left( {\mtx{X}^{T} \mtx{X} + n\alpha_{\rm ridge}^{\rm opt}\mtx{I}_{d}  }\right)^{-1} } }{\mtx{X} } } \right).
\end{equation}

\subsection{Proof Outline for Corollary \ref{corollary:well specified - transfer learning is better than ridge - isotropic beta assumption}}
\label{appendix:subsec:proof of corollary 4.1}
Consider $\mtx{H}=\mtx{\Psi}^T$ and $\mtx{\Psi}$ is an orthonormal matrix. 
The main case to be proved is for $d \notin \{{\widetilde{n}-1,\widetilde{n},\widetilde{n}+1}\}$. 
Then, according to Theorem \ref{theorem:well specified - out of sample error - target task -  nonzero alpha - expectation of eigenvalues - optimal - orthonormal H} and (\ref{appendix:eq:proof of theorem 3.2 - minimal error expression}), the test error of optimally tuned transfer learning can be written as 
\begin{equation}
\label{appendix:eq:proof of corollary 4.1 - transfer learning summation form}
\bar{\mathcal{E}}_{\rm TL}^{\rm opt}= \sigma_{\epsilon}^2 \left({ 1 +  \sum_{k=1}^{d} \mathbb{E}\Biggl\{{ { \frac{ 1 }{ \eigenvalue{\mtx{X}_{\mtx{\Psi}}^{T} \mtx{X}_{\mtx{\Psi}}}{k} + n{\alpha_{\rm TL}^{\rm opt}} } } }\Biggr\} }\right)  
\end{equation}
where $\mtx{X}_{\mtx{\Psi}}$ is a $n\times d$ matrix of i.i.d.~standard Gaussian variables. Note that the eigenvalues $\eigenvalue{\mtx{X}_{\mtx{\Psi}}^{T} \mtx{X}_{\mtx{\Psi}}}{k}$ are i.i.d. random variables. 

The optimally tuned ridge regression solution has the test error (\ref{appendix:eq:well specified - standard ridge regression - optimal - test error}) that can be also expressed as 
\begin{equation}
\label{appendix:eq:proof of corollary 4.1 - ridge summation form}
\bar{\mathcal{E}}_{\rm ridge}^{\rm opt}= \sigma_{\epsilon}^2 \left({ 1 +  \sum_{k=1}^{d} \mathbb{E}\Biggl\{{ { \frac{ 1 }{ \eigenvalue{\mtx{X}^{T} \mtx{X}}{k} + n{\alpha_{\rm ridge}^{\rm opt}} } } }\Biggr\} }\right)  
\end{equation}
where $\mtx{X}$ is a $n\times d$ matrix of i.i.d.~standard Gaussian variables. Note that the eigenvalues $\eigenvalue{\mtx{X}^{T} \mtx{X}}{k}$ are i.i.d. random variables. 

$\mtx{X}$ and $\mtx{X}_{\mtx{\Psi}}$ have the same distribution, hence, their eigenvalues $\eigenvalue{\mtx{X}^{T} \mtx{X}}{k}$, $\eigenvalue{\mtx{X}_{\mtx{\Psi}}^{T} \mtx{X}_{\mtx{\Psi}}}{k}$ are also identically distributed. Therefore, the only difference between (\ref{appendix:eq:proof of corollary 4.1 - transfer learning summation form}) and (\ref{appendix:eq:proof of corollary 4.1 - ridge summation form}) is the respective values of $\alpha_{\rm TL}^{\rm opt}$ and $\alpha_{\rm ridge}^{\rm opt}$. 
Then, according to the forms in (\ref{appendix:eq:proof of corollary 4.1 - transfer learning summation form})-(\ref{appendix:eq:proof of corollary 4.1 - ridge summation form}), ${\bar{\mathcal{E}}_{\rm TL}^{\rm opt} < \bar{\mathcal{E}}_{\rm ridge}^{\rm opt}}$ when ${\alpha_{\rm TL}^{\rm opt} > \alpha_{\rm ridge}^{\rm opt}}$. According to Theorem \ref{theorem:well specified - out of sample error - target task -  nonzero alpha - expectation of eigenvalues - optimal - orthonormal H}, the condition ${\alpha_{\rm TL}^{\rm opt} > \alpha_{\rm ridge}^{\rm opt}}$ is satisfied when 
${\frac{\sigma_{\epsilon}^2}{n C_{\rm TL}} > \frac{d\sigma_{\epsilon}^2}{n b} }$ where $C_{\rm TL}$ is defined in Lemma \ref{lemma:well specified - out of sample error - target task -  nonzero alpha - expectation of eigenvalues - orthonormal H} for the case of $\mtx{H}=\mtx{\Psi}^T$. This leads to the condition 
\begin{equation}
\label{appendix:eq:proof of corollary 4.1 - condition}
{\sigma_{\eta}^2 + \frac{d\cdot\sigma_{\xi}^2}{|d-\widetilde{n}|-1} <  b }
\end{equation}
for $d \notin \{{\widetilde{n}-1,\widetilde{n},\widetilde{n}+1}\}$.

Theorem \ref{theorem:well specified - out of sample error - target task -  nonzero alpha - expectation of eigenvalues - optimal - orthonormal H} states that the transfer learning error is infinite for ${d \in \{{\widetilde{n}-1,\widetilde{n},\widetilde{n}+1}\}}$. Hence,  ${\bar{\mathcal{E}}_{\rm TL}^{\rm opt} < \bar{\mathcal{E}}_{\rm ridge}^{\rm opt}}$ is never satisfied for ${d \in \{{\widetilde{n}-1,\widetilde{n},\widetilde{n}+1}\}}$.

\subsection{Generalization Error of Ridge Regression in Asymptotic Settings}
\label{appendix:subsec:Generalization Error of Ridge Regression at Asymptotic Settings}
Previous studies \cite{appendixDobriban2018high,appendixHastie2019surprises} already provided the analytical formula for the expected test error of ridge regression when the true parameter vector ($\vecgreek{\beta}$ in our case) originates at isotropic Gaussian distribution and the sample data matrix ($\mtx{X}$ in our case) has i.i.d.~Gaussian $\mathcal{N}(0,1)$ components. 
Then, translating the results from \cite{appendixDobriban2018high,appendixHastie2019surprises} to our notations shows that 
\begin{equation}
\label{appendix:eq:ridge asymptotic error}
{\bar{\mathcal{E}}_{\rm ridge} ^{\rm opt}} \rightarrow 
\sigma_{\epsilon}^2 
\left( { 1 + \gamma_{\rm tgt}\cdot m\left( -\alpha_{{\rm ridge}, \infty}^{\rm opt}; \gamma_{\rm tgt} \right) }\right)
\end{equation}
where $\alpha_{{\rm ridge}, \infty}^{\rm opt}=\frac{ \gamma_{\rm tgt} \sigma_{\epsilon}^2}{b}$ is the limiting value of $\alpha_{\rm ridge}^{\rm opt}$, and 
\begin{align}
\label{appendix:ridge asymptotic m function}
\resizebox{.9\hsize}{!}{$m\left( -\alpha_{{\rm ridge}, \infty}^{\rm opt}; \gamma_{\rm tgt} \right) = \frac{ -\left({ 1 - \gamma_{\rm tgt} + \alpha_{{\rm ridge}, \infty}^{\rm opt} }\right) + \sqrt{ \left({ 1 - \gamma_{\rm tgt} + \alpha_{{\rm ridge}, \infty}^{\rm opt} }\right)^2 + 4\gamma_{\rm tgt} \alpha_{{\rm ridge}, \infty}^{\rm opt} } }{2\gamma_{\rm tgt} \alpha_{{\rm ridge}, \infty}^{\rm opt} }$}
\end{align}
is the Stieltjes transform of the Marchenko-Pastur distribution. For more details see \cite{appendixDobriban2018high,appendixHastie2019surprises}.

\section{Details and Proofs for Section \ref{sec:Analysis for H of a General Form}}
\label{appendix:sec:Details and Proofs for Theorem 5}

\subsection{Proof of Proposition \ref{proposition:moments of theta given beta}}
\label{appendix:subsec:proof of proposition 3.4}
\begin{align}
\label{appendix:proof of proposition 3.4 - mean of theta hat given beta}
&\expectation{\widehat{\vecgreek{\theta}} \big| \vecgreek{\beta}} = \expectation{ \mtx{Z}^{+} \vec{v} \big| \vecgreek{\beta}}
\nonumber \\ 
&= \expectation{ \mtx{Z}^{+} \left({\mtx{Z}\vecgreek{\theta}+\vecgreek{\xi}}\right) \big| \vecgreek{\beta}}
= \expectation{ \mtx{Z}^{+} \mtx{Z}\left({\mtx{H}\vecgreek{\beta}+\vecgreek{\eta}}\right) \big| \vecgreek{\beta}}
\nonumber \\ 
& = \expectation{ \mtx{Z}^{+} \mtx{Z}} \mtx{H}\vecgreek{\beta}
\nonumber \\ 
&=\begin{cases}
\mathmakebox[6em][l] {\mtx{H}\vecgreek{\beta} }   \text{for } d \le \widetilde{n},  
\\
\mathmakebox[6em][l] {\frac{\widetilde{n}}{d}\mtx{H}\vecgreek{\beta} }   \text{for } d > \widetilde{n}
\end{cases} 
\end{align}
The last development relies on the fundamental result from (\ref{appendix:proof of proposition 3.4 - random projection of Gaussian iid matrix}).

Now, we continue to the covariance matrix of $\widehat{\vecgreek{\theta}}$ given $\vecgreek{\beta}$, namely,   
\begin{equation}
\label{appendix:eq:proof of proposition 3.4 - covariance of theta given beta - general formula}
\expectation{ \left( \widehat{\vecgreek{\theta}} - \expectation{\widehat{\vecgreek{\theta}} \big| \vecgreek{\beta}}\right)\left( \widehat{\vecgreek{\theta}} - \expectation{\widehat{\vecgreek{\theta}} \big| \vecgreek{\beta}}\right)^{T} \Big| \vecgreek{\beta}} =
\expectation{ \widehat{\vecgreek{\theta}} \widehat{\vecgreek{\theta}}^{T} \Big| \vecgreek{\beta}} - \expectation{\widehat{\vecgreek{\theta}} \big| \vecgreek{\beta}} \left({\expectation{\widehat{\vecgreek{\theta}} \big| \vecgreek{\beta}}}\right)^T.
\end{equation} 
Using (\ref{appendix:proof of proposition 3.4 - mean of theta hat given beta}) we can easily get that 
\begin{equation}
\label{appendix:eq:proof of proposition 3.4 - covariance of theta given beta - mean outer product}
\expectation{\widehat{\vecgreek{\theta}} \big| \vecgreek{\beta}} \left({\expectation{\widehat{\vecgreek{\theta}} \big| \vecgreek{\beta}}}\right)^T = \mtx{H}\vecgreek{\beta}\vecgreek{\beta}^T \mtx{H}^T \times \begin{cases}
\mathmakebox[3em][l] { 1 }   \text{for } d \le \widetilde{n},  
\\
\mathmakebox[3em][l] { \left({\frac{\widetilde{n}}{d}}\right)^2 }   \text{for } d > \widetilde{n}.
\end{cases} 
\end{equation} 
We also need analytical formulation for $\expectation{ \widehat{\vecgreek{\theta}} \widehat{\vecgreek{\theta}}^{T} \Big| \vecgreek{\beta}}$, as explained next. 
\begin{align}
\label{appendix:eq:proof of proposition 3.4 - covariance of theta given beta - theta hat outer product}
&\expectation{ \widehat{\vecgreek{\theta}} \widehat{\vecgreek{\theta}}^{T} \Big| \vecgreek{\beta}} = \expectation{ \mtx{Z}^{+} \left({\mtx{Z}\left({\mtx{H}\vecgreek{\beta}+\vecgreek{\eta}}\right)+\vecgreek{\xi}}\right)\left({\mtx{Z}\left({\mtx{H}\vecgreek{\beta}+\vecgreek{\eta}}\right)+\vecgreek{\xi}}\right)^{T} \left(\mtx{Z}^{+} \right)^{T} \Big| \vecgreek{\beta}}
\nonumber\\
& = \expectation{ \mtx{Z}^{+} \mtx{Z}\mtx{H}\vecgreek{\beta}\vecgreek{\beta}^{T}\mtx{H}^{T} \left(\mtx{Z}^{+} \mtx{Z}\right)^{T} \Big| \vecgreek{\beta}} + \expectation{ \mtx{Z}^{+} \mtx{Z}\vecgreek{\eta}\vecgreek{\eta}^{T} \left(\mtx{Z}^{+} \mtx{Z}\right)^{T}} + \expectation{ \mtx{Z}^{+} \vecgreek{\xi}\vecgreek{\xi}^{T} \left(\mtx{Z}^{+} \right)^{T}}
\nonumber\\
& = \expectation{ \mtx{Z}^{+} \mtx{Z}\mtx{H}\vecgreek{\beta}\vecgreek{\beta}^{T}\mtx{H}^{T} \left(\mtx{Z}^{+} \mtx{Z}\right)^{T} \Big| \vecgreek{\beta}} 
+ \frac{\sigma_{\eta}^2}{d} \expectation{ \mtx{Z}^{+} \mtx{Z}} + \sigma_{\xi}^2 \expectation{ \mtx{Z}^{+} \left(\mtx{Z}^{+} \right)^{T}}
\end{align} 
where the second term in the last expression can be explicitly formulated using (\ref{appendix:proof of proposition 3.4 - random projection of Gaussian iid matrix}). 
The third term in (\ref{appendix:eq:proof of proposition 3.4 - covariance of theta given beta - theta hat outer product}) requires the fundamental result from (\ref{appendix:proof of proposition 3.4 - pinvZ pinvZtransposed}).
The first term in (\ref{appendix:eq:proof of proposition 3.4 - covariance of theta given beta - theta hat outer product}) is an instance of the more general form  $\expectation{ \mtx{Z}^{+} \mtx{Z}\vec{a}\vec{a}^{T}\left(\mtx{Z}^{+} \mtx{Z}\right)^{T}}$, where $\vec{a}\in\mathbb{R}^{d}$ is a non-random vector. For $d\le \widetilde{n}$, we almost surely have that $\mtx{Z}^{+} \mtx{Z}=\mtx{I}_{d}$ and therefore ${\expectation{ \mtx{Z}^{+} \mtx{Z}\vec{a}\vec{a}^{T}\left(\mtx{Z}^{+} \mtx{Z}\right)^{T}}=\vec{a}\vec{a}^{T}}$. For $d > \widetilde{n}$, consider the decomposition $\mtx{Z}^{+} \mtx{Z}=\mtx{R}\mtx{R}^{T}$ where $\mtx{R}$ is a $d\times \widetilde{n}$ matrix with $\widetilde{n}$ orthonormal columns that are taken from a random orthonormal matrix that is uniformly distributed over the set of $d\times d$ orthonormal matrices (i.e., the Haar distribution of matrices). Then, using the non-asymptotic properties of Haar-distributed matrices (see, e.g., Lemma 2.5 in \cite{tulino2004random} and Proposition 1.2 in \cite{hiai2000asymptotic}) and some algebra, one can prove that, for $d > \widetilde{n}$, 
\begin{equation}
\label{appendix:eq:proof of propisition 3.4 - auxiliary result}
\expectation{ \mtx{Z}^{+} \mtx{Z}\vec{a}\vec{a}^{T}\left(\mtx{Z}^{+} \mtx{Z}\right)^{T}} = \frac{\widetilde{n}}{d}\left({ \frac{\widetilde{n}+1}{d+1}\vec{a}\vec{a}^{T} +  \frac{d-\widetilde{n}}{d^2 -1} {\rm diag}\left({ \{ \Ltwonorm{\vec{a}} - \left(a_{j}\right)^2 \}_{j=1,\dots,d} }\right)}\right)
\end{equation} 
where $a_{j}$ is the $j^{\rm th}$ component of the vector $\vec{a}$. 
Based on the described proof outline, one can use (\ref{appendix:eq:proof of proposition 3.4 - covariance of theta given beta - theta hat outer product}) to develop (\ref{appendix:eq:proof of proposition 3.4 - covariance of theta given beta - general formula}) into the form 
\begin{equation}
\label{appendix:eq:proof of proposition 3.4 - covariance of theta given beta - underparameterized source task}
\expectation{ \left( \widehat{\vecgreek{\theta}} - \expectation{\widehat{\vecgreek{\theta}} \big| \vecgreek{\beta}}\right)\left( \widehat{\vecgreek{\theta}} - \expectation{\widehat{\vecgreek{\theta}} \big| \vecgreek{\beta}}\right)^{T} \Big| \vecgreek{\beta}} = \left(\frac{\sigma_{\eta}^2}{d} + \frac{\sigma_{\xi}^{2}}{\widetilde{n}-d-1} \right)\mtx{I}_{d}
\end{equation} 
for $d\le \widetilde{n}-2$, and 
\begin{align}
\label{appendix:eq:proof of proposition 3.4 - covariance of theta give beta - overparameterized source task}
&\resizebox{.45\hsize}{!}{$\expectation{ \left( \widehat{\vecgreek{\theta}} - \expectation{\widehat{\vecgreek{\theta}} \big| \vecgreek{\beta}}\right)\left( \widehat{\vecgreek{\theta}} - \expectation{\widehat{\vecgreek{\theta}} \big| \vecgreek{\beta}}\right)^{T} \Big| \vecgreek{\beta}} =$}
\nonumber\\
& \resizebox{.9\hsize}{!}{$\frac{\widetilde{n}}{d}\left({ \frac{d-\widetilde{n}}{d(d+1)}\mtx{H}\vecgreek{\beta}\vecgreek{\beta}^{T}\mtx{H}^{T} +  \frac{d-\widetilde{n}}{d^2 -1} {\rm diag}\left({ \{ \Ltwonorm{\mtx{H}\vecgreek{\beta}} - \left(\{\mtx{H}\vecgreek{\beta}\}_{j}\right)^2 \}_{j=1,\dots,d} }\right) + \left(\frac{\sigma_{\eta}^2}{d}  + \frac{\sigma_{\xi}^{2}}{d-\widetilde{n}-1} \right)\mtx{I}_{d}}\right)$}
\end{align} 
for $d \ge \widetilde{n}+2$. In (\ref{appendix:eq:proof of proposition 3.4 - covariance of theta give beta - overparameterized source task}), ${\{\mtx{H}\vecgreek{\beta}\}_{j}}$ is the $j^{\rm th}$ component of the vector $\mtx{H}\vecgreek{\beta}$. 
For $d\in\{\widetilde{n}-1,\widetilde{n},\widetilde{n}+1\}$ the covariance matrix is infinite valued as a result of the infinite valued $\expectation{  \left(\mtx{Z}^{T} \mtx{Z}\right)^{+} }$, see (\ref{appendix:proof of proposition 3.4 - pinvZ pinvZtransposed}).

\subsection{Lemma~\ref{lemma:random-matrix-theory}}
To consider the general covariance case in the asymptotic setting, we prove a general result on linear and quadratic functionals of resolvents of the random data sample covariance matrix.

\begin{lemma}
    \label{lemma:random-matrix-theory}
    If $\mtx{X} = [\vec{x}^{(1)}, \ldots, \vec{x}^{(n)}]^T$ for i.i.d.\ $\vec{x}^{(i)} \sim \mathcal{N}(\vec{0}, \mtx{\Sigma})$ for $\mtx{\Sigma} \in \mathbb{R}^{d \times d}$ having bounded spectral norm, and $\mtx{\Theta} \in \mathbb{R}^{d \times d}$ such that $\mtxtrace{\left( \mtx{\Theta}^T \mtx{\Theta} \right)^{1/2}}$ is uniformly bounded in $p$, and $\mtx{\Xi} \in \mathbb{R}^{d \times d}$ is a positive semi-definite matrix, then with probability one, for each $\alpha > 0$, as $n, d \to \infty$ such that $d/n \to \gamma_{\rm tgt}$,
    \begin{align}
    \label{appendix:eq:random matrix theory lemma - result 1}
        \mtxtrace{\mtx{\Theta} \left( \left(\tfrac{1}{n} \mtx{X} \mtx{X}^T + \alpha \mtx{I}_d \right)^{-1} - \left(c(\alpha) \mtx{\Sigma} + \alpha \mtx{I}_d \right)^{-1} \right)} \to 0
    \end{align}
    and
    \begin{align}
    \label{appendix:eq:random matrix theory lemma - result 2}
        \Tr \Big \lbrace \mtx{\Theta} \Big ( &\left(\tfrac{1}{n} \mtx{X} \mtx{X}^T + \alpha \mtx{I}_d \right)^{-1} \mtx{\Xi} \left(\tfrac{1}{n} \mtx{X} \mtx{X}^T + \alpha \mtx{I}_d \right)^{-1} \nonumber \\
        &- \left(c(\alpha) \mtx{\Sigma} + \alpha \mtx{I}_d \right)^{-1} \left(c'(\alpha) \mtx{\Sigma} + \mtx{\Xi}\right) \left(c(\alpha) \mtx{\Sigma} + \alpha \mtx{I}_d \right)^{-1} \Big ) \Big \rbrace \to 0,
    \end{align}
    where $c(\alpha)$ is the unique solution $c$ of $\frac{1}{c} - 1 = \frac{\gamma_{\rm tgt}}{d} \mtxtrace{\mtx{\Sigma} (c \mtx{\Sigma} + \alpha \mtx{I}_d)^{-1}}$, and 
    \begin{align}
        c'(\alpha) = \frac{\frac{\gamma_{\rm tgt}}{d} \mtxtrace{\mtx{\Sigma} (c(\alpha) \mtx{\Sigma} + \alpha \mtx{I}_d)^{-1} \mtx{\Xi} (c(\alpha) \mtx{\Sigma} + \alpha \mtx{I}_d)^{-1}}}{c(\alpha)^{-2} - \frac{\gamma_{\rm tgt}}{d} \mtxtrace{\mtx{\Sigma} (c(\alpha) \mtx{\Sigma} + \alpha \mtx{I}_d)^{-1} \mtx{\Sigma} (c(\alpha) \mtx{\Sigma} + \alpha \mtx{I}_d)^{-1}}}.
    \end{align}
\end{lemma}

Proof of Lemma~\ref{lemma:random-matrix-theory}: by Theorem~1 of~\cite{rubio2011spectral}, for any $t \geq 0$, we have that with probability one, for any $\alpha > 0$,
\begin{align}
    \label{eq:first-order-equivalence}
    \mtxtrace{\mtx{\Theta} \left( \left(t \mtx{\Xi} + \tfrac{1}{n} \mtx{X} \mtx{X}^T + \alpha \mtx{I}_d \right)^{-1} - \left(t \mtx{\Xi} + c(\alpha, t) \mtx{\Sigma} + \alpha \mtx{I}_d \right)^{-1} \right)} \to 0,
\end{align}
where $c(\alpha, t)$ is the unique solution $c$ of
\begin{align}
    \frac{1}{c} - 1 = \frac{\gamma_{\rm tgt}}{d} \mtxtrace{\mtx{\Sigma} (t \mtx{\Xi} + c \mtx{\Sigma} + \alpha \mtx{I}_d)^{-1}}.
\end{align}
By choosing $t = 0$, we obtain the first result of Lemma~\ref{lemma:random-matrix-theory} immediately. Then by Theorem~11 of \cite{dobriban2020wonder}, we know that the derivative with respect to $t$ of the left-hand side of \eqref{eq:first-order-equivalence} also goes to zero. That is,
\begin{align}
    \Tr \Big \lbrace \mtx{\Theta} \Big ( &\left(t \mtx{\Xi} + \tfrac{1}{n} \mtx{X} \mtx{X}^T + \alpha \mtx{I}_d \right)^{-1} \mtx{\Xi} \left(t \mtx{\Xi} + \tfrac{1}{n} \mtx{X} \mtx{X}^T + \alpha \mtx{I}_d \right)^{-1} \nonumber \\
    &- \left(t \mtx{\Xi} + c(\alpha, t) \mtx{\Sigma} + \alpha \mtx{I}_d \right)^{-1} \left(c'(\alpha, t) \mtx{\Sigma} + \mtx{\Xi}\right) \left(t \mtx{\Xi} + c(\alpha, t) \mtx{\Sigma} + \alpha \mtx{I}_d \right)^{-1} \Big ) \Big \rbrace \to 0,
\end{align}
where
\begin{align}
    c'(\alpha, t) & = \frac{\partial c(\alpha, t)}{\partial t} 
    \\ \nonumber
    &= \frac{\frac{\gamma_{\rm tgt}}{d} \mtxtrace{\mtx{\Sigma} (t \mtx{\Xi} + c(\alpha, t) \mtx{\Sigma} + \alpha \mtx{I}_d)^{-1} \mtx{\Xi} (t \mtx{\Xi} + c(\alpha, t) \mtx{\Sigma} + \alpha \mtx{I}_d)^{-1}}}{c(\alpha, t)^{-2} - \frac{\gamma_{\rm tgt}}{d} \mtxtrace{\mtx{\Sigma} (t \mtx{\Xi} + c(\alpha, t) \mtx{\Sigma} + \alpha \mtx{I}_d)^{-1} \mtx{\Sigma} (t \mtx{\Xi} + c(\alpha, t) \mtx{\Sigma} + \alpha \mtx{I}_d)^{-1}}}.
\end{align}
By again choosing $t = 0$, we obtain the second result of Lemma~\ref{lemma:random-matrix-theory}.

\subsection{Proof of Theorem \ref{theorem:well specified - out of sample error - target task -  asymptotic - anisotropic target features - general H}}
\label{appendix:subsec:proof of theorem 5}

Similarly to (\ref{appendix:eq:proof of lemma 3.1 - error development 1})-(\ref{appendix:eq:proof of lemma 3.1 - error development 2}) that were given above for the case of orthonormal $\mtx{H}$, $\widetilde{\mtx{H}}=\mtx{H}$ and $\mtx{\Sigma}_{\vec{x}}=\mtx{I}_{d}$, one can express the expected error for the case of general forms of $\widetilde{\mtx{H}}$, $\mtx{H}$ and $\mtx{\Sigma}_{\vec{x}}$ (specifically note that, here, $\widetilde{\mtx{H}}$ can differ from $\mtx{H}$) as
\begin{equation}
\label{appendix:eq:proof of lemma 3.5 - error development 1}
\bar{\mathcal{E}}_{\rm TL} = \sigma_{\epsilon}^2 + \expectation{ \left \Vert { \left({ \mtx{X}^{T} \mtx{X} + n{\alpha_{\rm TL}}\widetilde{\mtx{H}}^{T}\widetilde{\mtx{H}} }\right)^{-1} \left( { \mtx{X}^{T} \vecgreek{\epsilon} + n{\alpha_{\rm TL}}\widetilde{\mtx{H}}^{T} \left({\widehat{\vecgreek{\theta}}-\widetilde{\mtx{H}}\vecgreek{\beta}}\right) }\right) } \right \Vert _{\mtx{\Sigma}_{\vec{x}}}^2 }
\end{equation}
Then, we define \[\mtx{\Gamma}_{\rm TL}\triangleq \expectation{\left({\widehat{\vecgreek{\theta}}-\widetilde{\mtx{H}}\vecgreek{\beta}}\right)\left({\widehat{\vecgreek{\theta}}-\widetilde{\mtx{H}}\vecgreek{\beta}}\right)^T}\] and develop its expression using (\ref{appendix:proof of proposition 3.4 - random projection of Gaussian iid matrix}), (\ref{appendix:proof of proposition 3.4 - pinvZ pinvZtransposed}), Proposition \ref{proposition:moments of theta given beta}, and the isotropic assumption on $\vecgreek{\beta}$. 
Also, we define $\mtx{W}\triangleq (\widetilde{\mtx{H}}^{-1})^{T}\mtx{\Sigma}_{\vec{x}}\widetilde{\mtx{H}}^{-1}$ and $\mtx{X}_{\widetilde{\mtx{H}}^{-1}}\triangleq \mtx{X}\widetilde{\mtx{H}}^{-1} $. Then, we bring (\ref{appendix:eq:proof of lemma 3.5 - error development 1}) into the form of 
\begin{align}
\label{appendix:eq:proof of theorem 5 - error development 1 - row 1}
& \bar{\mathcal{E}}_{\rm TL} = \sigma_{\epsilon}^2 + \sigma_{\epsilon}^2 \frac{d}{n}\expectation{ \mtxtrace{\frac{1}{d}\mtx{W}\left({ \frac{1}{n}\mtx{X}_{\widetilde{\mtx{H}}^{-1}}^{T} \mtx{X}_{\widetilde{\mtx{H}}^{-1}} + {\alpha_{\rm TL}}\mtx{I}_d }\right)^{-1}}} 
\\
\label{appendix:eq:proof of theorem 5 - error development 1 - row 2}
& + \sigma_{\epsilon}^2 \frac{d}{n}\expectation{ \mtxtrace{ \mtx{A} \left({ \frac{1}{n}\mtx{X}_{\widetilde{\mtx{H}}^{-1}}^{T} \mtx{X}_{\widetilde{\mtx{H}}^{-1}} + {\alpha_{\rm TL}}\mtx{I}_d }\right)^{-1}\mtx{W} \left({ \frac{1}{n}\mtx{X}_{\widetilde{\mtx{H}}^{-1}}^{T} \mtx{X}_{\widetilde{\mtx{H}}^{-1}} + {\alpha_{\rm TL}}\mtx{I}_d }\right)^{-1} }}
\end{align}
where $\mtx{A}\triangleq \frac{n \alpha_{\rm TL}^2}{d\sigma_{\epsilon}^2}\mtx{\Gamma}_{\rm TL} - \frac{\alpha_{\rm TL}}{d}\mtx{I}_d$.
Note that the rows of $\mtx{X}_{\widetilde{\mtx{H}}^{-1}}$ are i.i.d.~from $\mathcal{N}\left({\vec{0},\mtx{W}}\right)$ and that by choosing   $\mtx{\Theta}=\frac{1}{d}\mtx{W}$ we can apply (\ref{appendix:eq:random matrix theory lemma - result 1}) from Lemma \ref{lemma:random-matrix-theory}
on the trace term in (\ref{appendix:eq:proof of theorem 5 - error development 1 - row 1}).
Moreover, by choosing $\mtx{\Theta}=\frac{n \alpha_{\rm TL}^2}{d\sigma_{\epsilon}^2}\mtx{\Gamma}_{\rm TL} - \frac{\alpha_{\rm TL}}{d}\mtx{I}_d$ and $\mtx{\Xi}=\mtx{W}$ we can apply (\ref{appendix:eq:random matrix theory lemma - result 2}) from Lemma \ref{lemma:random-matrix-theory}
on the trace term in (\ref{appendix:eq:proof of theorem 5 - error development 1 - row 2}). 
Consequently, the limiting value of $\bar{\mathcal{E}}_{\rm TL}$ can be formulated as in Theorem \ref{theorem:well specified - out of sample error - target task -  asymptotic - anisotropic target features - general H}.

\subsection{Proof of Theorem \ref{theorem:LMMSE Transfer Learning}}
\label{appendix:sec:proof of LMMSE theorem}
We seek to find the estimator of $\vecgreek{\beta}$ that is linear in $\vec{u} = \begin{bmatrix}
\vec{y} \\
\widehat{\vecgreek{\vec{\theta}}}
\end{bmatrix}$ and minimizes the mean squared error. That is, we seek $\widehat{\vecgreek{\beta}}_{\rm LMMSE} = \mtx{M} \vec{u}$ for some $\mtx{M} \in \mathbb{R}^{d \times (n + d)}$ that minimizes
\begin{align}
    \expectation{\left\Vert \widehat{\vecgreek{\beta}}_{\rm LMMSE} - \vecgreek{\beta} \right\Vert_2^2 \Big| \mtx{X}}.
\end{align}
This problem has a solution given by the orthogonality principle:
\begin{align}
    \expectation{(\mtx{M} \vec{u} - \vecgreek{\beta}) \vec{u}^T | \mtx{X}} = \vec{0} 
    \quad \implies \quad
    \mtx{M} = \expectation{\vecgreek{\beta} \vec{u}^T | \mtx{X}} \left( \expectation{\vec{u} \vec{u}^T | \mtx{X}} \right)^{-1}.
\end{align}
These expectations are simple to evaluate:
\begin{align}
    \expectation{\vecgreek{\beta} \vec{u}^T | \mtx{X}} = 
    \begin{bmatrix}
    \mtx{B}_d \mtx{X}^T & \expectation{\vecgreek{\beta}\widehat{\vecgreek{\theta}}^T}
    \end{bmatrix},
    \quad
    \expectation{\vec{u} \vec{u}^T | \mtx{X}} = 
    \begin{bmatrix}
    	\mtx{X}\mtx{B}_d \mtx{X}^T + \sigma_{\epsilon}^2 \mtx{I}_d & \mtx{X}\expectation{\vecgreek{\beta}\widehat{\vecgreek{\theta}}^T} \\
    	\expectation{\widehat{\vecgreek{\theta}}\vecgreek{\beta}^T}\mtx{X}^T & \expectation{\widehat{\vecgreek{\theta}}\widehat{\vecgreek{\theta}}^T}
	\end{bmatrix},
\end{align}
and we can use Proposition~\ref{proposition:moments of theta given beta} to formulate $\expectation{\vecgreek{\beta}\widehat{\vecgreek{\theta}}^T}$ and $\expectation{\widehat{\vecgreek{\theta}}\widehat{\vecgreek{\theta}}^T}$ in the forms that are provided in Theorem \ref{theorem:LMMSE Transfer Learning}.

\bibliographystyle{siamplain}
\bibliography{fine_tuning_references}

\end{document}